\documentclass{article} 
\usepackage[usenames,dvipsnames]{color}
\usepackage{amsmath} 
\usepackage{amssymb} 
\usepackage{bm}
\usepackage{amsthm}
\usepackage{CJK}
\usepackage{indentfirst}
\usepackage{tikz}
\usepackage{amsmath,amsfonts,amsthm,mathrsfs}
\usepackage{hyperref}
\numberwithin{equation}{section}
\usepackage{geometry}
\usepackage{extarrows}
\usepackage{setspace}

\numberwithin{equation}{section}
\newtheorem{theorem}{Theorem}[section]
\newtheorem{lemma}{Lemma}[section]

\newtheorem{definition}{Definition}[section]
\newtheorem{remark}{Remark}[section]

\usepackage{algorithm,algpseudocode}
\usepackage{amsthm}
\usepackage{esint} 
\usepackage{subfigure}
\usepackage{natbib}
\usepackage{lineno}

\def\NN{\mathbb{N}}
\def\PP{\mathbb{P}}
\def\RR{\mathbb{R}}
\def\BB{\mathbb{B}}
\def\mL{\mathcal{L}^2}
\def\mB{\mathcal{B}_{2,\infty}}
\def\mH{\mathcal{H}}

\def\SS{\mathbb{S}}

\def\mal{\max\limits}
\def\mil{\min\limits}
\def\sul{\sum\limits}

\def\pc{\mathcal{P}^c}
\def\mq{\mathfrak{q}}

\makeatletter
\renewcommand\@fnsymbol[1]{\arabic{footnote}}
\makeatother

\title{Geometry-Dependent Approximation and Stability Properties\\
of Shallow ReLU$^k$ Networks on Bounded Euclidean Domains}

\author{Yuqing Liu\thanks{RPTU University Kaiserslautern-Landau, Kaiserslautern and Landau in der Pfalz, Rhineland-Palatinate, Germany}\and Tong Mao\thanks{Shenzhen Loop Area Institute, Shenzhen 518000, P. R. China}\and Jinchao Xu\thanks{Shenzhen Loop Area Institute, Shenzhen 518000, P. R. China}}
\date{}

\begin{document}
	
\maketitle

\begin{abstract}
On bounded Euclidean domains, we study how hidden-parameter geometry affects shallow ReLU$^k$ networks through results on approximation rates, saturation-type lower bounds, Gram and stiffness matrix spectra, Bernstein and Marcinkiewicz--Zygmund inequalities, discrete least-squares recovery, and inverse approximation.

Our analysis uses smooth localization and dyadic Fourier estimates. We quantify parameter geometry by the fill distance $h$ and antipodal separation distance $\underline h$. For the prescribed network space $L_n^k$, representative approximation estimates take the form
\begin{equation*}
\begin{aligned}
\inf_{g_n\in L_n^k}\|f-g_n\|_{\mathcal L^2(\Omega')}&\lesssim h^r\|f\|_{\mathcal H^r(\Omega)},\\
\inf_{g_n\in L_n^k}\|f-g_n\|_{\mathcal L^2(\Omega)}&\gtrsim \underline h^{\frac{d+2k+1}{2}}\|f\|_{\mathcal L^2(\Omega')}.
\end{aligned}
\end{equation*}
The upper bound exploits target regularity up to the critical order $\frac{d+2k+1}{2}$, while the lower bound applies asymptotically to fixed targets smoother than this order. The upper estimate holds on an interior subdomain $\Omega'$, whereas the lower estimate concerns the larger domain $\Omega$.

We also obtain geometry-dependent spectral estimates that quantify conditioning and coefficient stability. Further results establish norm discretization and regularized recovery from function values, and infer target regularity from approximation rates under coefficient or separation conditions. Altogether, these results describe how parameter geometry governs approximation accuracy, the stability of coefficient representations, discrete recovery, and the regularity information contained in convergence rates.
\end{abstract}
\section{Introduction}
\label{sec:introduction}

From a mathematical perspective, central questions in machine learning concern how target regularity and structure, network size, and available samples jointly determine achievable accuracy \cite{cucker2007learning,devore2021neural,liu2025functional}. Addressing these questions requires quantitative approximation estimates and an analysis of stable recovery from finitely many function values.

Shallow neural networks provide a starting point for these questions and a reference for studying deeper architectures \cite{pinkus1999approximation,barron1993universal,devore2021neural}. For an activation function $\sigma$, consider
\begin{equation}\label{eq:intro-nonlinear-network}
    \Sigma_n^\sigma
    :=\left\{
        x\mapsto\sum_{j=1}^n a_j\sigma(w_j\cdot x+b_j):
        w_j\in\mathbb R^d,\quad a_j,b_j\in\mathbb R
    \right\}.
\end{equation}
To study how network size affects achievable accuracy, approximation theory provides density results under suitable activation assumptions \cite{cybenko1989approximation,hornik1989multilayer,leshno1993multilayer} and quantitative width-dependent error bounds in Barron, variation, spectral Barron, and Sobolev spaces \cite{barron1993universal,mhaskar1993approximation,petrushev1998approximation,devore2021neural,liu2024korobov}.

To study how available samples affect achievable accuracy, statistical learning theory combines approximation and estimation error analyses through empirical risk minimization, regularization, and capacity bounds \cite{anthony1999neural,bartlett2002rademacher,cucker2007learning}. For neural networks, this interplay yields risk bounds and near-minimax estimation guarantees \cite{barron1994approximation,parhi2022near,liu2024korobov,
liu2025classification}; related analyses of random-design regression and random-feature learning quantify the roles of sample size, model complexity, and regularization \cite{hsu2014random,rudi2017generalization}.

Alongside network size and available samples, the distribution of hidden parameters influences both approximation accuracy and recovery from discrete data. We study these relationships for shallow networks with the ReLU activation and its integral-order variants, defined by
\begin{equation*}
\sigma_k(t)=
\begin{cases}
t^k, & t>0,\\
0, & t\leq0,
\end{cases}
\quad k=0,1,2,\ldots.
\end{equation*}
Approximation rates for shallow ReLU$^k$ networks have been studied in \cite{siegel2022optimal,mao2023rates,yang2024optimal,mao2024approximation}. In particular, \cite{mao2024approximation} establishes the $\mathcal O(n^{-\frac{r}{d}})$ rate in the $\mL$ norm for targets in $\mathcal H^r$ with $0<r\leq\frac{d+2k+1}{2}$, showing that these networks can exploit target regularity well beyond the degree of individual neurons.

To examine this dependence, we normalize each nonzero hidden-parameter vector by positive homogeneity as $\displaystyle\theta=\binom{w}{b}\in\mathbb S^d$. For a configuration $\Theta_n=\{\theta_j^*\}_{j=1}^n$, with $\displaystyle\theta_j^*=\binom{w_j^*}{b_j^*}$, the associated networks form the space
\begin{equation}\label{shallowk}
    L_n^k=L_n^k(\Theta_n)
    :=\operatorname{span}\left\{
        \sigma_k(w_j^*\cdot x+b_j^*):j=1,\ldots,n
    \right\}=\operatorname{span}\left\{
        \sigma_k(\theta_j^*\cdot\widetilde x):j=1,\ldots,n
    \right\}.
\end{equation}
Here we write
\begin{equation*}
\widetilde x:=\binom{x}{1},\qquad x\in\RR^d.
\end{equation*}

Prescribed parameter distributions yield geometry-dependent approximation upper bounds \cite{petrushev1998approximation,liu2025integral}. We ask how this geometry also affects coefficient stability, recovery from discrete data, and the limits of approximation.

For prescribed hidden parameters, computing the output coefficients by least squares reduces to solving a linear system whose coefficient matrix is the associated Gram matrix. Its spectrum therefore connects parameter geometry to the conditioning of this system and the stability of the computed coefficients \cite{liu2025condition,cohen2013stability}.

For prescribed parameter configurations, these issues lead to the following questions:
\begin{enumerate}
    \item How does the distribution of hidden parameters affect the achievable approximation accuracy?
    \item How does the parameter geometry determine the eigenvalue spectrum and condition number of the associated Gram matrices, and hence the stability of the coefficient representation?
    \item How many function values are needed to recover the continuous approximation rate through discrete least squares?
    \item What regularity of the target function can be inferred from a given approximation rate?
    \item Does additional target smoothness always improve the approximation rate?
\end{enumerate}

Classical polynomial, spline, and kernel approximation connects direct estimates with norm control, sampling stability, inverse regularity, and saturation \cite{devore1993constructive,schumaker2007spline,wendland2005scattered}. This framework motivates a corresponding analysis of shallow networks.

For shallow neural networks, several parts of this framework have recently been established on the sphere: geometry-dependent approximation upper bounds \cite{liu2025integral}, Sobolev Gram matrix spectra and condition numbers \cite{liu2025condition}, and approximation lower bounds and saturation \cite{mao2025configuration}. Bernstein inequalities, sampling estimates, discrete least-squares recovery, and inverse results further relate parameter geometry to discrete data and target regularity \cite{liu2026optimal}. The approximation and spectral bounds are sharp under antipodal quasi-uniformity. Section~\ref{sec_related} gives further details and comparisons.

Extending these results to bounded Euclidean domains raises additional difficulties. On the full sphere, harmonic expansions provide a common framework for approximation, spectral estimates, and sampling. A bounded Euclidean domain lifts to only part of $\mathbb S^d$, and restriction may disrupt the spectral structure and norm comparisons; see \cite[Section~7]{liu2025condition}. Spherical condition-number bounds, Bernstein inequalities, approximation lower bounds, and least-squares guarantees may therefore fail on general domains. Corresponding results require formulations that account for both parameter geometry and domain geometry.

In this paper, we study how hidden-parameter distributions affect the approximation properties of shallow ReLU$^k$ networks on bounded Euclidean domains. We use smooth localization to work in a whole-space Fourier framework. Let $\Omega\subset\mathbb R^d$ be a bounded connected domain and $\Omega'\Subset\Omega$ an interior subdomain, and choose a cutoff satisfying
\begin{equation}\label{eq:cutoff}
\chi\in\mathcal{C}_c^\infty(\Omega),\quad 0\leq\chi\leq1,\quad \chi=1\ \text{on }\Omega'.
\end{equation}
The corresponding localized neurons are
\begin{equation}\label{eq:localized-neuron}
    \varphi_\theta(x):=\chi(x)\sigma_k(w\cdot x+b),\quad \theta=\binom{w}{b}\in\mathbb S^d,\quad x\in\mathbb R^d,
\end{equation}
and, for a parameter configuration $\Theta_n$, they span the localized network space
\begin{equation}\label{eq:localized-network-space}
    \chi L_n^k=\chi L_n^k(\Theta_n):=\{\chi g:g\in L_n^k(\Theta_n)\}=\operatorname{span}\{\varphi_{\theta_j^*}:j=1,\ldots,n\}.
\end{equation}

The localized functions are supported away from $\partial\Omega$, so Fourier analysis on $\mathbb R^d$ introduces no boundary singularities. Since $\chi=1$ on $\Omega'$, estimates for the localized error also control the original approximation error on $\Omega'$.

To ensure that localization retains the singularity of each neuron, we choose a connected closed set $U\subset\Omega$ with $\Omega'\Subset U^\circ$ and $\chi\geq c_1>0$ on $U$, and consider parameters in the active set
\begin{equation}\label{eq:active-set}
\mathcal A_\chi:=\left\{\binom{w}{b}\in\mathbb S^d:\{x:w\cdot x+b=0\}\cap U\neq\varnothing\right\}.
\end{equation}
For the localized estimates, we take $\Theta_n\subset\mathcal A_\chi$, so every singular hyperplane intersects a region where $\chi$ is uniformly positive.

Let $\rho$ denote the geodesic distance on $\mathbb S^d$, and write $\rho_\pm(\theta,\eta):=\min\{\rho(\theta,\eta),\rho(-\theta,\eta)\}$ for the antipodal distance. For a parameter region $\mathcal A\subset\mathbb S^d$ containing $\Theta_n$, the fill distance and antipodal separation distance are
\begin{equation*}
    h:=\sup_{\theta\in\mathcal A}\min_{1\leq j\leq n}\rho(\theta,\theta_j^*),\quad
    \underline h:=\min_{i\neq j}\rho_\pm(\theta_i^*,\theta_j^*).
\end{equation*}
In the localized setting, we use $\mathcal A=\mathcal A_\chi$; under antipodal quasi-uniformity, $h\eqsim\underline h\eqsim n^{-\frac{1}{d}}$.

Our analysis uses localized Fourier estimates, integral representations, and spherical approximation tools. The main contributions are as follows.

\begin{enumerate}
\item \textbf{Localized Fourier estimates.} We estimate the norms of individual dyadic Fourier blocks of the localized neurons and establish the localization bound
\begin{equation}\label{eq:intro-localization}
\left|\left\langle\Delta_{q,s}\varphi_\theta,\Delta_{q,s}\varphi_{\theta'}\right\rangle_{\mathcal{L}^2(\mathbb R^d)}\right|\lesssim_\gamma\frac{2^{-q(2(k-s)+1)}}{\big(2^q\rho_\pm(\theta,\theta')\big)^\gamma}.
\end{equation}
Here $\Delta_{q,s}$ denotes a Littlewood--Paley Sobolev block, with $q\geq1$, $\gamma>d$, $0\leq s<k+\frac{1}{2}$, and $\rho_\pm(\theta,\theta')>0$. These localized spectral estimates are key tools for approximation and harmonic analysis and underpin several results below.

\item \textbf{Sharp spectral estimates for Gram and stiffness matrices.} For $0\leq s<k+\frac{1}{2}$, the Sobolev Gram matrices are given by
\begin{equation}\label{eq:intro-stiffness-matrix}
\begin{aligned}
(K_s)_{ij}:=\langle\varphi_{\theta_i^*},\varphi_{\theta_j^*}\rangle_{\mathcal H^s(\mathbb R^d)}=\int_{\mathbb R^d}(I-\bigtriangleup)^{\frac{s}{2}}\varphi_{\theta_i^*}(x)(I-\bigtriangleup)^{\frac{s}{2}}\varphi_{\theta_j^*}(x)\,dx,\quad 1\leq i,j\leq n.
\end{aligned}
\end{equation}
We obtain configuration-dependent bounds for all eigenvalues, ordered nonincreasingly. Under antipodal quasi-uniformity, these bounds are sharp:
\begin{equation}\label{eq:intro-spectrum}
\lambda_j(K_s)\eqsim n j^{-\frac{d+2(k-s)+1}{d}},\quad \kappa(K_s)\eqsim n^{\frac{d+2(k-s)+1}{d}},\quad 1\leq j\leq n.
\end{equation}
The identity $\big\|\sum_{j=1}^n a_j\varphi_{\theta_j^*}\big\|_{\mathcal H^s(\mathbb R^d)}^2=a^\top K_s a$, with $a=(a_1,\ldots,a_n)^\top$, relates these spectral estimates to coefficient stability.

\item \textbf{Configuration-dependent approximation lower bounds.} For targets $f\in\mathcal H^r(\Omega)$ with $r>\frac{d+2k+1}{2}$, we prove, for sufficiently large $n$,
\begin{equation}\label{eq:intro-lower-bound}
\inf_{g_n\in\chi L_n^k}\|\chi f-g_n\|_{\mathcal{L}^2(\mathbb R^d)}\gtrsim n^{-\frac{1}{2}}\underline h^{k+\frac{1}{2}}\|\chi f\|_{\mathcal{L}^2(\mathbb R^d)}.
\end{equation}
This also bounds the approximation error on $\Omega$ from below and gives the order $n^{-\frac{d+2k+1}{2d}}$ under antipodal quasi-uniformity. As a complement, we adapt the upper estimates of \cite{liu2025integral} to parameters in $\mathcal A_\chi$, obtaining interior approximation bounds with controlled output coefficients.

\item \textbf{Bernstein and Marcinkiewicz--Zygmund inequalities.} For the localized network spaces with parameters in $\mathcal A_\chi$, we control higher Sobolev norms and establish equal-weight norm discretization:
\begin{equation}\label{eq:intro-bernstein-mz}
\begin{aligned}
\|g_n\|_{\mathcal H^t(\mathbb R^d)}&\lesssim\underline h^{-(t-s)}\|g_n\|_{\mathcal H^s(\mathbb R^d)},\\
\|g_n\|_{\mathcal L^2(\mathbb R^d)}^2&\eqsim\frac{1}{m}\sum_{i=1}^mg_n(x_i)^2,\quad g_n\in\chi L_n^k.
\end{aligned}
\end{equation}
The Bernstein inequality holds for $0\leq s<t<k+\frac{1}{2}$. The MZ inequality holds for antipodally quasi-uniform parameters and quasi-uniform samples in $\Omega$ when $k>\frac{d-1}{2}$ and $m\gtrsim n$, with sufficient oversampling conditions for smaller $k$.

\item \textbf{Discrete least-squares recovery.} For prescribed parameters covering the full sphere $\SS^d$, we obtain recovery estimates without assuming parameter quasi-uniformity. On bounded Lipschitz domains, with $k>\frac{d-1}{2}$ and quasi-uniform samples satisfying $m\gtrsim\big(h^{\frac{2k+1}{d+2k+1}}\underline h^{\frac{d}{d+2k+1}}\big)^{-d}$ with a sufficiently small implicit constant, regularized discrete least squares gives
\begin{equation}\label{eq:intro-discrete-rate}
\|f-f_{n,m}\|_{\mathcal H^s(\Omega')}\lesssim h^{r-s}\Big(\frac{h}{\underline h}\Big)^{\frac{ds}{d+2k+1}}
\begin{cases}
\|f\|_{\mathcal W^{r,p}(\Omega)},&\frac{d}{p}<r\leq\frac{d}{2},\quad 2<p<\infty,\\
\|f\|_{\mathcal H^r(\Omega)},&\frac{d}{2}<r\leq\frac{d+2k+1}{2},
\end{cases}
\end{equation}
where $0\leq s<\min\{r,k+\frac{1}{2}\}$. The scheme uses localized neurons and the coefficient penalty $h^{2k+1}\|a\|_2^2$. Under antipodal quasi-uniformity, $m\gtrsim n$ suffices and the rate reduces to $n^{-\frac{r-s}{d}}$.

\item \textbf{Inverse approximation and structural conditions.} Within the antipodally quasi-uniform setting, for $0<t<\frac{d+2k+1}{2}$, approximation by two disjoint parameter families whose union has antipodal separation $\gtrsim n^{-\frac{1}{d}}$ gives
\begin{equation}\label{eq:intro-inverse}
\begin{aligned}
&\inf_{g_n\in\chi L_n^k(\Theta_n)}\|f-g_n\|_{\mathcal{L}^2(\mathbb R^d)}+\inf_{\widetilde g_n\in\chi L_n^k(\widetilde\Theta_n)}\|f-\widetilde g_n\|_{\mathcal{L}^2(\mathbb R^d)}\lesssim n^{-\frac{t}{d}}\\
&\Longrightarrow\quad f\in\mathcal B_{2,\infty}^t(\mathbb R^d).
\end{aligned}
\end{equation}
A coefficient budget yields another inverse result without nestedness. Further results identify the limits of inverse conclusions from unconstrained rates or qualitative interior convergence.
\end{enumerate}

Section~\ref{sec_related} compares these results with related work. Sections~\ref{sec_preliminary} and~\ref{sec_main} give the preliminaries and main results. Section~\ref{sec_loc} proves the localized Fourier estimates, and Section~\ref{sec_proof} proves the remaining results.

\section{Related work}\label{sec_related}

We compare spectral, approximation, sampling, and inverse results, with particular attention to their domains, parameter configurations, and coefficient conditions.

\subsection{Spectral estimates and conditioning}\label{subsec:related-spectral}

Spectral perspectives on neural networks include the spectral bias and frequency principle \cite{rahaman2019spectral,xu2020frequency}, infinite-width neural tangent kernels \cite{jacot2018neural}, residual-network kernels on spheres \cite{belfer2024spectral}, and kernel-regression learning curves \cite{bordelon2020spectrum,canatar2021spectral}. Related results study integral-operator spectra approximated by randomly sampled kernel matrices \cite{koltchinskii2000random,rosasco2010learning} and kernel-matrix spectra when data dimension and sample size grow proportionally \cite{elkaroui2010spectrum}. These studies provide spectral descriptions of training, learning, and random kernel models, while our focus is the dependence of finite Gram and stiffness spectra on neuron parameter geometry.

For unit-slope ReLU neurons with uniformly spaced breakpoints on $[0,1]$, \cite{hong2022activation} establishes the Gram eigenvalue estimates $\lambda_j(K_0^{\mathrm R})\eqsim nj^{-4}$. For ordered breakpoints in one-dimensional Ritz discretizations, \cite{cai2025efficient} gives $\kappa(K_s^{\mathrm R})\lesssim nh_{\min}^{-(3-2s)}$, $s=0,1$, for the ReLU mass and derivative stiffness matrices, where $h_{\min}$ is the smallest partition interval. Uniform spacing gives upper bounds of order $n^4$ and $n^2$, consistent with our estimates for $k=1$ and $s=0,1$. Our results apply in arbitrary dimension, parameter geometry, and give configuration-dependent upper and lower bounds for all eigenvalues.

For shallow ReLU networks on $d$-dimensional unit balls, \cite{zhang2025high} obtains $\mu_j\eqsim j^{-\frac{d+3}{d}}$ for the continuous $\mathcal{L}^2$ Gram operator with the uniform direction--bias measure on $\mathbb S^{d-1}\times[-1,1]$. Our estimates concern discrete Sobolev matrices of localized ReLU$^k$ neurons and depend on parameter geometry without requiring a uniform direction--bias distribution.

For ReLU$^k$ neurons with both inputs and parameters on $\mathbb S^d$, \cite{liu2025condition} gives geometric bounds for the eigenvalues of the mass and integer-order stiffness matrices. Under antipodal quasi-uniformity,
\begin{equation}\label{eq:related-spherical-spectrum}
\lambda_j(K_s^{\mathbb S})\eqsim n j^{-\frac{d+2(k-s)+1}{d}},\quad \kappa(K_s^{\mathbb S})\eqsim n^{\frac{d+2(k-s)+1}{d}},\quad 1\leq j\leq n,
\end{equation}
for integer $0\leq s\leq k$. These matrices use inner products over the full sphere. Our estimates retain this spectral exponent but apply to localized neurons on bounded Euclidean domains.

For standard piecewise linear nodal bases on shape-regular, quasi-uniform simplicial meshes, the finite element mass and Dirichlet stiffness matrices satisfy $\kappa(K_s^{\mathrm{FE}})\eqsim n^{\frac{2s}{d}}$  ($s=0,1$), where $n$ is the number of degrees of freedom \cite{kamenski2014conditioning,xu1992iterative}. This comparison depends on the basis: already in one dimension, ReLU hinges supplemented by affine functions and nodal hat functions can represent the same piecewise linear space while having substantially different Gram conditioning \cite{zhang2025high}. Connections with numerical discretization also include ReLU representations of linear finite element functions \cite{he2020finiteelements} and the MgNet framework linking multigrid with convolutional neural networks \cite{he2019mgnet}.

\subsection{Approximation under prescribed parameter configurations}\label{subsec:related-approximation}

Universal approximation theorems establish density under suitable activation assumptions \cite{cybenko1989approximation,hornik1989multilayer,leshno1993multilayer}; surveys describe the broader approximation theory \cite{pinkus1999approximation,devore2021neural,mhaskar2026perspective}. Quantitative deep-network approximation theory relates accuracy to network size, depth, and target smoothness \cite{yarotsky2017error,yarotsky2018optimal,lu2021smooth}. Related results treat deep convolutional approximation on spheres \cite{fang2020spherical}, simultaneous approximation of functions and derivatives on manifolds \cite{zhou2025manifolds}, anisotropic and mixed smoothness \cite{yang2026anisotropic}, and sparse compositional targets under parameter norm budgets \cite{huang2026compositional}.

For shallow networks with freely chosen hidden parameters, quantitative rates have been studied in Barron and variation spaces \cite{barron1993universal,bach2017breaking,klusowski2018approximation,siegel2022sharp,siegel2022optimal} and Sobolev spaces \cite{mhaskar1993approximation,mao2023rates,yang2024optimal,mao2024approximation}. The Barron-type rate $\mathcal O(n^{-\frac{1}{2}})$ is related to Maurey's sampling principle \cite{maurey1973type,pisier1981remarques} and probabilistic or greedy approximation \cite{Jones1992,devore1996some,makovoz1996random,Barron2008}, with refined estimates accounting for the activation, parameter geometry, and target regularity \cite{bach2017breaking,klusowski2018approximation,xu2020finite,siegel2022sharp,siegel2022optimal,siegel2022high,mhaskar2023tractability}. Related greedy analyses are developed in \cite{li2024entropy,li2025empirical}. For shallow ReLU$^k$ networks on bounded domains, \cite{he2026radon} studies Radon-domain $\mathcal L^p$ ridge integral spaces, their Sobolev embeddings, and randomized approximation rates; at $p=2$, the integral space recovers the critical Sobolev space $\mathcal H^{\frac{d+2k+1}{2}}$ \cite{liu2025integral}.

Related work on prescribed ridge directions studies density conditions \cite{lin1993fundamentality}, Sobolev approximation on balls \cite{petrushev1998approximation}, and worst-case uniform approximation of Lipschitz functions \cite{jorgensen2021approximation}. Target-independent centers also appear in spherical zonal networks \cite{mhaskar2019zonal}, while fixed random neurons are used in Petrov--Galerkin and discontinuous Galerkin methods for differential equations \cite{shang2023petrov,sun2024dg}. For prescribed ReLU$^k$ parameters covering the full sphere, \cite{liu2025integral} gives bounded-domain $\mathcal L^2$ error of order $h^r$ for $f\in\mathcal H^r(\Omega)$ with $r\leq\frac{d+2k+1}{2}$. Our focus below is configuration-dependent lower bounds and saturation for individual smooth targets.

Saturation results identify limiting rates for classical approximation schemes \cite{lorentz1964inequalities,butzer1971fourier,schumaker2007spline}. For degree-$k$ piecewise polynomial spaces $V_h$ on quasi-uniform meshes of size $h$, \cite{lin2014lower} establishes
\begin{equation}\label{eq:related-polynomial-lower}
\inf_{v\in V_h}\|f-v\|_{\mathcal{L}^2(\Omega)}\gtrsim h^{k+1}
\end{equation}
for sufficiently smooth fixed targets with a nonzero derivative of order $k+1$. For conditionally positive definite thin plate splines associated with $\bigtriangleup^\ell$ on a bounded domain $\Omega$ satisfying an interior cone condition, \cite{schaback2002inverse} proves
\begin{equation}\label{eq:related-rbf-saturation}
\|f-I_Xf\|_{\mathcal{L}^\infty(K)}=o(h_X^{2\ell})\quad\Longrightarrow\quad\bigtriangleup^\ell f=0\ \text{in }\Omega,
\end{equation}
where $f\in\mathcal{C}^{2\ell}(\Omega)$, $I_Xf$ is the polynomially augmented interpolant at admissible centers $X$, and the hypothesis holds for every compact $K\Subset\Omega$ as the fill distance $h_X$ tends to zero.

Saturation has also been studied for neural network approximation operators. For one-dimensional max-product operators generated by sigmoidal functions, \cite{costarelli2017saturation} identifies the saturation order $n^{-1}$, with faster uniform convergence restricted to constant targets. For classical multivariate sigmoidal sampling operators, \cite{costarelli2025saturation} shows that interior uniform convergence of order $o(n^{-2})$ forces $\bigtriangleup f=0$ under suitable decay and moment conditions, where $n$ denotes the grid resolution. These results identify saturation of prescribed sigmoidal approximation schemes. Our results relate hidden-parameter geometry to approximation lower bounds for shallow ReLU$^k$ networks with general parameter configurations in $\mathcal A_\chi$ and unrestricted output coefficients.

For prescribed ReLU$^k$ neurons on $\mathbb S^d$, \cite{mao2025configuration} establishes the asymptotic lower bound
\begin{equation}\label{eq:related-spherical-lower}
\inf_{g_n\in L_n^{k,\mathbb S}}\|f-g_n\|_{\mathcal L^2(\mathbb S^d)}\gtrsim n^{-\frac{1}{2}}\underline h^{k+\frac{1}{2}}\|f\|_{\mathcal L^2(\mathbb S^d)},
\end{equation}
where $L_n^{k,\mathbb S}$ is the span of the prescribed neurons and $f\in\mathcal H^r(\mathbb S^d)$ is fixed with $r>\frac{d+2k+1}{2}$. Together with \cite{liu2025integral}, this identifies the saturation order $n^{-\frac{d+2k+1}{2d}}$ under antipodal quasi-uniformity on the corresponding smooth parity class. Our lower bounds concern bounded Euclidean domains and use separation in $\mathcal A_\chi$.

\subsection{Bernstein and Marcinkiewicz--Zygmund inequalities}\label{subsec:related-sampling}

Continuous and discrete norm estimates have been studied through Sobolev sampling inequalities and scattered-data approximation \cite{narcowich2005sobolev,narcowich2006sobolev,arcangeli2014sampling}, general MZ inequalities \cite{kolomoitsev2026marcinkiewicz}, and their connection with weighted least squares \cite{grochenig2020sampling}. Our focus is the dependence of Bernstein inequalities and equal-weight norm discretization on prescribed hidden-parameter geometry.

Bernstein inequalities for spherical basis function spaces $V_Z=\operatorname{span}\{\phi(z\cdot\eta):z\in Z\}$ generated by positive definite Green kernels and their admissible perturbations are established in \cite{mhaskar2010bernstein}. A representative $\mathcal{L}^2$ estimate is
\begin{equation}\label{eq:related-sbf-bernstein}
\|v\|_{\mathcal H^r(\mathbb S^d)}\lesssim q_Z^{-r}\|v\|_{\mathcal{L}^2(\mathbb S^d)},\quad v\in V_Z,\quad 0<r<\beta-\frac{d}{2},
\end{equation}
where $q_Z$ is the center separation distance, $\beta$ is the kernel order, and the constant may depend on the mesh ratio of $Z$. Our Bernstein estimates retain this separation-dependent scaling for localized affine ReLU$^k$ neurons, with antipodal separation replacing the separation of kernel centers.

MZ inequalities for spherical polynomials at scattered samples are established in \cite{mhaskar2001spherical} and extended to diffusion polynomials on compact manifolds in \cite{filbirMhaskar2011mz}. In the spherical setting, quasi-uniform samples $X=\{x_i\}_{i=1}^m$ with fill distance $h_X$ satisfy
\begin{equation}\label{eq:related-polynomial-mz}
\|p\|_{\mathcal{L}^2(\mathbb S^d)}^2\eqsim\sum_{i=1}^m\omega_i|p(x_i)|^2,\quad p\in\Pi_L,\quad L\lesssim h_X^{-1},
\end{equation}
where $\Pi_L$ is the space of spherical polynomials of degree at most $L$, $\omega_i$ are positive partition weights, and the constant in $L\lesssim h_X^{-1}$ is sufficiently small. For quasi-uniform nodes, $\omega_i\eqsim m^{-1}$ and the required sample size is of order $L^d$, matching the space dimension. These polynomial estimates also support sample-based encoders for learning nonlinear functionals on spheres \cite{yang2026spherical}. Our MZ estimates concern localized ReLU$^k$ spaces on bounded Euclidean domains instead of polynomial spaces.

\subsection{Discrete least-squares recovery}\label{subsec:related-least-squares}

Recovery from finite data is studied in statistical learning theory through empirical risk minimization, regularization, and capacity bounds \cite{anthony1999neural,bartlett2002rademacher,cucker2007learning}, including neural-network risk bounds and near-minimax guarantees \cite{barron1994approximation,parhi2022near}, random-design regression \cite{hsu2014random}, and random-feature learning \cite{rahimi2007random,rudi2017generalization}. Related analyses treat empirical-feature regularization \cite{guo2016empirical}, distributed regularized least squares with Gaussian kernels \cite{hu2021distributed}, and hyper-kernel ridge regression for multi-index models \cite{huang2025hyperkernel}.

Further sample-based guarantees cover spectral algorithms for functional linear regression \cite{fan2024spectral}, nonlinear functional regression with kernel-embedded deep networks \cite{shi2025functional}, and neural-network distribution regression with two-stage sampling \cite{shi2025distribution}. Iterative methods include early stopping and multi-pass stochastic gradient descent with general convex losses \cite{hu2022early,lei2021multipass}, and truncated kernel stochastic gradient descent for least squares on spheres \cite{bai2026truncated}. Least-squares and collocation formulations also appear in kernel methods \cite{hangelbroek2026extending}, extreme learning machines, and neural-network methods for differential equations \cite{HZS:2006,raissi2019physics,chen2022bridging,wang2024extreme}. Our focus is recovery of continuous approximation rates from quasi-uniform function samples using prescribed hidden parameters.

Discrete least-squares approximation in finite-dimensional spaces from random evaluations is studied in \cite{cohen2013stability,chkifa2015discrete}. For an $n$-dimensional trial space, the sampling measure and weights based on its Christoffel function in \cite{cohen2017optimal} yield high-probability stability in the regime
\begin{equation}\label{eq:related-random-ls}
m\gtrsim n\log n,\quad \frac{1}{2}I_n\preceq G_{n,m}\preceq\frac{3}{2}I_n,
\end{equation}
where $G_{n,m}$ is the weighted empirical Gram matrix in an $\mathcal{L}^2$-orthonormal basis. These results cover general trial spaces and give accuracy guarantees for stabilized least-squares estimators. Our result uses equal-weight quasi-uniform samples and a quadratic coefficient penalty. The sufficient sampling density is expressed through parameter fill distance and antipodal separation, without assuming parameter quasi-uniformity or requiring a separate spectral check of $G_{n,m}$.

For radial basis spaces on spheres, \cite{legia2006continuous} studies continuous and discrete least-squares surface fitting with separate center and data sets. On bounded Euclidean domains, \cite{li2019discrete} establishes continuous Sobolev error estimates for discrete least squares using positive definite kernels whose native spaces are equivalent to Sobolev spaces. Its stability conditions quantify the relationship between sampling density, center separation, and native-space regularization.

For ReLU$^k$ neurons on $\mathbb S^d$, \cite{liu2026optimal} proves that equal-weight least squares recovers the continuous approximation rate when $k>\frac{d-1}{2}$, the parameters are antipodally quasi-uniform, and the samples are quasi-uniform with $m\gtrsim n$. For targets satisfying the required parity condition, the least-squares approximant $f_{n,m}^{\mathbb S}$ satisfies
\begin{equation}\label{eq:related-spherical-ls}
\|f-f_{n,m}^{\mathbb S}\|_{\mathcal H^s(\mathbb S^d)}\lesssim n^{-\frac{r-s}{d}}
\begin{cases}
\|f\|_{\mathcal W^{r,p}(\SS^d)}, & \frac{d}{p}<r\leq\frac{d}{2},\quad 2<p<\infty,\\
\|f\|_{\mathcal H^r(\SS^d)}, & \frac{d}{2}<r\leq\frac{d+2k+1}{2},
\end{cases}
\end{equation}
for $0\leq s<\min\{r,k+\frac{1}{2}\}$ and a sufficiently large constant in the sampling condition. The same work treats residual least squares for elliptic spectral multipliers and bounded-domain recovery through spherical lifting. Our formulation uses localized neurons with parameters covering the full sphere and a scalar quadratic coefficient penalty. It gives interior error estimates on bounded Lipschitz domains for the same two target-regularity regimes, without a parity condition or parameter quasi-uniformity; the sampling density and error bounds depend explicitly on parameter coverage and separation.

\subsection{Inverse approximation theory}\label{subsec:related-inverse}

Inverse approximation theory relates convergence rates to target regularity and membership in approximation spaces. Classical results cover polynomial and spline approximation, nonlinear approximation on adaptive partitions, and scattered-data approximation \cite{bernstein1912quelques,nikolskii1975approximation,devore1993constructive,schumaker2007spline,devore1998nonlinear,wendland2005scattered}. Related neural-network studies describe approximation and variation spaces \cite{siegel2022sharp,siegel2023characterization} and establish converse results for recurrent models \cite{wang2023inverse}. Our focus is how hidden-parameter geometry and coefficient growth determine the regularity information contained in shallow-network approximation rates.

Bernstein inequalities yield inverse theorems for polynomial and spline approximation \cite{devore1993constructive,schumaker2007spline} and Besov characterizations through nested spherical basis function spaces \cite{mhaskar2010bernstein}. Related kernel results include native-space inverse theorems for radial basis interpolation \cite{schaback2002inverse} and localized Bernstein estimates on bounded Euclidean domains \cite{hangelbroek2018inverse}. For nested antipodally quasi-uniform parameter sets, our localized Bernstein inequality similarly gives
\begin{equation}\label{eq:related-bernstein-inverse}
\inf_{g_n\in\chi L_n^k(\Theta_n)}\|f-g_n\|_{\mathcal L^2(\mathbb R^d)}\lesssim n^{-\frac{t}{d}}\quad\Longrightarrow\quad f\in\mathcal B_{2,\infty}^t(\mathbb R^d),\quad 0<t<k+\frac{1}{2}.
\end{equation}
This conclusion is limited by the regularity of individual neurons. Our inverse results beyond this threshold require coefficient control or simultaneous approximation on two suitably separated parameter families.

For deep networks, \cite{huang2026compositional} gives approximation upper bounds under a budget on the product of layer Frobenius norms. Converse results for shallow networks characterize function spaces through bounded representations. For shallow ReLU networks, \cite{E2019} proves that pointwise convergence to a continuous target under a uniform path-norm bound implies Barron-space membership. For a normalized ReLU$^k$ dictionary $\mathcal D$ and approximants $g_n=\sum_{j=1}^n a_{j,n}\phi_{j,n}$ with $\phi_{j,n}\in\mathcal D$, the variation-space characterization in \cite{siegel2023characterization} gives
\begin{equation}\label{eq:related-variation-inverse}
\|f-g_n\|_{\mathcal L^2(\Omega)}\to0,\quad \sup_n\sum_{j=1}^n|a_{j,n}|\leq M\quad\Longrightarrow\quad \|f\|_{\mathcal D}\leq M,
\end{equation}
where $\|\cdot\|_{\mathcal D}$ denotes the variation norm. Sharp embeddings between quasi-Banach Besov spaces and shallow ReLU$^k$ variation spaces are established in \cite{li2026besov}. These results concern bounded representations and function-space embeddings; our inverse results relate growing coefficient budgets and quantitative approximation rates to Besov regularity.

For ReLU$^k$ networks on $\mathbb S^d$, \cite{liu2026optimal} derives Sobolev inverse conclusions below $k+\frac{1}{2}$ from a Bernstein inequality along nested antipodally quasi-uniform parameter sequences. The coefficient-controlled integral characterization in \cite{liu2025integral} reaches the critical Sobolev order $\frac{d+2k+1}{2}$, while \cite{liu2026parameter} treats growing coefficient budgets. For quasi-uniform parameters and $g_n(\eta)=\sum_{j=1}^n a_{j,n}\sigma_k(\theta_{j,n}^*\cdot\eta)$, the latter establishes
\begin{equation}\label{eq:related-spherical-inverse}
\begin{gathered}
\|f-g_n\|_{\mathcal L^2(\mathbb S^d)}\lesssim n^{-\frac{d+2k+1}{2d}+\alpha},\quad \Bigl(n\sum_{j=1}^n|a_{j,n}|^2\Bigr)^{\frac{1}{2}}\lesssim n^\alpha\\
\Longrightarrow\quad f\in\mathcal B_{2,\infty,*}^{\frac{d+2k+1}{2}-\alpha d}(\mathbb S^d),
\end{gathered}
\end{equation}
for $0<\alpha<\frac{d+2k+1}{2d}$, where the subscript $*$ records the spherical parity restriction. It also shows that an unconstrained rate alone can fail to imply regularity above the neuron threshold, and recovers the inverse conclusion from two parameter sequences whose union is antipodally separated. Our results establish corresponding whole-space Besov conclusions for localized affine neurons with parameters in $\mathcal A_\chi$.

\section{Preliminaries}\label{sec_preliminary}
Throughout the paper, for $A,B\geq0$, we write $A\lesssim B$ whenever $A\leq CB$ for some constant $C>0$. We write $A\gtrsim B$ when $B\lesssim A$, and $A\eqsim B$ when both $A\lesssim B$ and $A\gtrsim B$ hold. The implicit constants may change from one occurrence to another and depend only on the fixed parameters $d$, $k$, $s$, $r$, $\Omega',~U,~\Omega$ and $\chi$ unless otherwise stated. In the antipodal quasi-uniform setting $h\leq C^*\underline{h}$, they may additionally depend on the quasi-uniformity constant $C^*$, but not on $n$. Dependence on any other parameters is stated explicitly or indicated by a subscript on $\gtrsim,\lesssim,\eqsim$. For example, if $A\leq CB$ with $C$ depending on an index $\gamma$, we write $A\lesssim_\gamma B$.

We also use the notation $C_1,C_2,\dots$ and $c_1,c_2,\dots$ to denote constants, whose dependence will be clear from the context. Uppercase constants are considered to be large constants in the sense that the corresponding results remain valid if they are increased; lowercase constants are considered to be small constants in the sense that the results remain valid if they are decreased.

\subsection{Network spaces and localization}\label{subsec:network-spaces}
We retain the domains $\Omega,\Omega'$, the set $U$, and the cutoff $\chi$ introduced in Section~\ref{sec:introduction}. The prescribed network spaces, localized neurons, and localized spaces are defined in \eqref{shallowk}, \eqref{eq:localized-neuron}, and \eqref{eq:localized-network-space}. For the same prescribed parameters, the coefficient-constrained classes are
\begin{equation}
    \begin{split}
        &L_{n,M}^{k,\alpha}:=\Big\{\sul_{j=1}^na_j\sigma_k(w_j^*\cdot\circ+b_j^*):\Big(n\sul_{j=1}^na_j^2\Big)^{\frac{1}{2}}\leq n^\alpha M\Big\},\\
        &\chi L_{n,M}^{k,\alpha}:=\{\chi f_n:~f_n\in L_{n,M}^{k,\alpha}\}=\Big\{\sul_{j=1}^na_j\varphi_{\theta_j^*}:\Big(n\sul_{j=1}^na_j^2\Big)^{\frac{1}{2}}\leq n^\alpha M\Big\}.
    \end{split}
\end{equation}
The active parameter set is given in \eqref{eq:active-set}, and the associated Gram and stiffness matrices $K_s$ are defined in \eqref{eq:intro-stiffness-matrix}.

\subsection{Parameter and sample geometry}\label{subsec:parameter-geometry}
Throughout, $|\cdot|$ denotes the Euclidean norm on $\RR^d$ and $\rho$ denotes the geodesic distance on the unit sphere
\[
    \mathbb S^d=\{x\in\mathbb R^{d+1}:|x|=1\},
\]
that is,
\begin{equation}
    \rho(\theta,\eta)=\arccos(\theta\cdot\eta),\quad\theta,\eta\in\SS^d.
\end{equation}

\begin{definition}[Quasi-uniformity and antipodal quasi-uniformity]\label{def_quasiuniform}
Let $d\in\mathbb{N}$. For a finite set
$\Theta_n=\{\theta_{j}^*\}_{j=1}^n\subset\mathcal{A}\subset\mathbb{S}^d$, define
\begin{equation*}
    h=h_n:=\sup_{\theta\in\mathcal{A}}\min_{1\le j\le n}\rho(\theta,\theta_{j}^*),
    \qquad
    \tilde{h}=\tilde{h}_n:=\min_{i\ne j}\rho(\theta_{i}^*,\theta_{j}^*),\qquad \underline{h}=\underline{h}_n:=\min_{i\ne j}\rho_\pm(\theta_{i}^*,\theta_{j}^*),
\end{equation*}
where
\begin{equation*}
    \rho_\pm(\theta,\eta)=\min\Big\{
      \rho(\theta,\eta),
      \rho(\theta,-\eta)
    \Big\}.
\end{equation*}
The set $\Theta_n$ is called quasi-uniform or antipodally quasi-uniform on $\mathcal{A}$ if there are constants such that
\begin{equation}\label{eqn:quasi-uniform}
    h_n\lesssim\tilde{h}_n,\qquad h_n\lesssim\underline{h}_n,
\end{equation}
respectively.

Similarly, a collection $X=\{x_i\}_{i=1}^m\subset\Omega$ is called quasi-uniform on $\Omega$ if $h_X\lesssim\underline{h}_X$, where
\begin{equation*}
    h_X:=\sup\limits_{x\in\Omega}\min_{1\leq i\leq m}|x-x_i|,\qquad\underline{h}_X:=\mil_{i\neq j}|x_i-x_j|.
\end{equation*}
\end{definition}

\subsection{Besov spaces and Sobolev spaces}\label{subsec:besov-sobolev}
We describe the Sobolev inner product using the Fourier transform, then introduce the dyadic blocks used in the Besov norms.

For $f\in \mathcal{L}^1(\RR^d)\cap\mL(\RR^d)$, define the Fourier transform by
\begin{equation}
    \widehat{f}(\xi)=\int_{\RR^d}f(x)e^{-i\xi\cdot x}dx,\quad\xi\in\RR^d.
\end{equation}
The transform extends to $\mL(\RR^d)$, with inverse understood in the $\mathcal{L}^2$ sense:
\begin{equation}
    f(x)=\frac{1}{(2\pi)^d}\int_{\RR^d}\widehat{f}(\xi)e^{ix\cdot\xi}d\xi,\quad\text{in }\mathcal{L}^2(\RR^d).
\end{equation}

Let $r\geq0$, and let $\bigtriangleup=\displaystyle\sul_{j=1}^d\frac{\partial^2}{\partial x_j^2}$ denote the Laplace operator. Parseval's identity gives
\begin{equation*}
    \begin{split}
        \int_{\RR^d}(I-\bigtriangleup)^{\frac{r}{2}}f(x)\overline{(I-\bigtriangleup)^{\frac{r}{2}}g(x)}dx=\frac{1}{(2\pi)^d}\int_{\RR^d}\widehat{f}(\xi)\overline{\widehat{g}(\xi)}(1+|\xi|^2)^rd\xi.
    \end{split}
\end{equation*}
Therefore, the inner product on $\mH^r(\RR^d)$ is defined by
\begin{equation}\label{eqn_def_inner_sob}
    \begin{split}        \left<f,g\right>_{\mH^r(\RR^d)}:=\frac{1}{(2\pi)^d}\int_{\RR^d}\widehat{f}(\xi)\overline{\widehat{g}(\xi)}(1+|\xi|^2)^rd\xi.
    \end{split}
\end{equation}
We use \eqref{eqn_def_inner_sob} for the standard Sobolev inner product; in particular, $\mL(\RR^d)=\mH^0(\RR^d)$.
For any measurable $D\subset\RR^d$, we use $\|f\|_{\mL(D)}^2:=\int_D|f(x)|^2dx$; in particular, $\|f\|_{\mL(\RR^d)}^2=\frac{1}{(2\pi)^d}\int_{\RR^d}|\widehat f(\xi)|^2d\xi$.

To define the Besov norms, fix a smooth cutoff function $\zeta$ satisfying the following conditions (see, e.g., \cite[(3.6)]{petrushev2005localized}):
\begin{eqnarray}\label{eqn_def_zeta}
&&\zeta\in\mathcal{C}^\infty(\RR),\quad\zeta\geq0,\quad\mathrm{supp}(\zeta)\subset[1/2,2],\\
&&\zeta(t)>c_2>0,\qquad t\in[3/5,5/3],\\
&&\zeta(t)+\zeta(2t)=1,\quad t\in[1/2,1].
\end{eqnarray}
A standard consequence of these properties is the partition-of-unity identity
\begin{equation}\label{eqn_zeta_sum1}
    \begin{split}
        1=&\sul_{q=0}^\infty\zeta(2^{-q}t),\qquad t\geq1,\\
        1=&(1-\zeta)(t)+\zeta(t)=(1-\zeta)(t)+\sul_{q=0}^\infty\zeta(2^{-q}t),\qquad 0\leq t<1.
    \end{split}
\end{equation}

\begin{definition}[Littlewood–Paley Sobolev operators]
For $s\geq0$, the Littlewood–Paley Sobolev operators $\Delta_{q,s}$ are the Fourier multiplier operators
\begin{equation}
\label{eq:lp-blocks}
    \widehat{\Delta_{q,s} f}(\xi)
    =\zeta_{q,s}(\xi)\widehat f(\xi),
    \quad q=0,1,\dots,
\end{equation}
given by the multipliers
\begin{equation}
    \zeta_{0,s}(\xi)=\Big[1-\sum_{q=1}^{\infty}\zeta\Big(\frac{|\xi|}{2^q}\Big)\Big](1+|\xi|^2)^{s/2},\qquad\zeta_{q,s}(\xi)=\zeta\Big(\frac{|\xi|}{2^q}\Big)(1+|\xi|^2)^{s/2},\quad q\geq1.
\end{equation}
For $q\geq1$, these operators localize to $|\xi|\eqsim2^q$; the block $q=0$ contains the low frequencies. We call $\Delta_{q,s}f$ a Littlewood--Paley Sobolev block.
\end{definition}

\begin{definition}[Besov and Sobolev spaces]\label{def_bes_sob}
    For $r\geq0$, $1\leq p\leq\infty$, the Besov spaces $\mathcal B_{2,p}^r(\RR^d)$ are defined by the following norms:
    \begin{equation}
        \|f\|_{\mathcal B_{2,p}^r(\RR^d)}=\left\{\begin{array}{cc}
            \left(\sul_{q=0}^\infty\|\Delta_{q,r}f\|_{\mL(\RR^d)}^p\right)^{\frac1p}, &\quad1\leq p<\infty  \\
            \sup\limits_{q\in\NN_0}\|\Delta_{q,r}f\|_{\mL(\RR^d)}, &\quad p=\infty. 
        \end{array}\right.
    \end{equation}
    In particular, for $p=2$, the Besov space $\mathcal B_{2,2}^r(\RR^d)$ is equivalent to the Sobolev space $\mH^r(\RR^d)$.

    For any bounded domain $\Omega\subset\RR^d$, the Besov spaces $\mathcal B_{2,p}^r(\Omega)$ and Sobolev spaces $\mH^r(\Omega)$ are defined by the norms
    \begin{equation}
        \|g\|_{\mathcal B_{2,p}^r(\Omega)}=\inf\limits_{f|_\Omega=g}\|f\|_{\mathcal B_{2,p}^r(\RR^d)},\qquad\|g\|_{\mH^r(\Omega)}=\inf\limits_{f|_\Omega=g}\|f\|_{\mH^r(\RR^d)}.
    \end{equation}

    We also use the Bessel potential Sobolev spaces $\mathcal W^{r,p}$ for $r\geq0$ and $1<p<\infty$, defined by
\begin{equation}
\|f\|_{\mathcal W^{r,p}(\mathbb R^d)}:=\|(I-\bigtriangleup)^{\frac{r}{2}}f\|_{\mathcal L^p(\mathbb R^d)},\quad \|g\|_{\mathcal W^{r,p}(\Omega)}:=\inf_{f|_\Omega=g}\|f\|_{\mathcal W^{r,p}(\mathbb R^d)}.
\end{equation}
For $p=2$ or $r\in\NN$, this norm is equivalent to the usual weak-derivative Sobolev norm on Lipschitz domains; for $p=2$, $\mathcal W^{r,2}=\mathcal H^r$ with equivalent norms.
\end{definition}

\begin{remark}
In the summation \eqref{eqn_zeta_sum1}, the dyadic supports have finite overlap, with at most two nonzero terms at each frequency. Hence the Littlewood--Paley square sum is equivalent to the Sobolev norm
     \begin{equation}
         \|f\|_{\mH^r(\RR^d)}^2=\frac{1}{(2\pi)^d}\int_{\RR^d}\big|\widehat{f}(\xi)\big|^2(1+|\xi|^{2})^rd\xi.
     \end{equation}
\end{remark}

\section{Main results}\label{sec_main}

We present the main results in the network settings introduced above. These results relate hidden-parameter geometry to approximation accuracy, matrix spectra, recovery from discrete data, and inverse regularity.

We begin with the dyadic norm estimate for individual localized neurons and the off-diagonal estimate for their interactions. These are used in the spectral, approximation lower-bound, and norm-inequality proofs. Section~\ref{sec_loc} proves the two estimates separately.

\begin{theorem}[Localized Fourier estimates]\label{thm:localized-estimates}
For every $0\leq s<k+\frac{1}{2}$, there exists $q_0\in\mathbb N$ such that
\[
\|\Delta_{q,s}\varphi_\theta\|_{\mathcal{L}^2(\mathbb R^d)}^2\eqsim 2^{-q(2(k-s)+1)},\quad q\ge q_0
\]
uniformly for $\theta\in\mathcal A_\chi$.

For every $\gamma>d$ and $\theta,\theta'\in\mathcal A_\chi$ with $\rho_\pm(\theta,\theta')>0$,
\begin{equation}
\left| \left\langle \Delta_{q,s}\varphi_{\theta}, \Delta_{q,s}\varphi_{\theta'} \right\rangle_{\mL(\mathbb R^d)} \right| \lesssim_\gamma\frac{2^{-q(2(k-s)+1)}}{ \left( 2^q \rho_\pm(\theta,\theta') \right)^{\gamma}},\qquad q\geq1.
\end{equation}
\end{theorem}

We next consider approximation in the prescribed network spaces and the spectra of the associated matrices.

Approximation upper bounds on bounded Euclidean domains were established in \cite{liu2025integral} using parameters covering the full sphere. The next theorem adapts these bounds to coverage of $\mathcal A_\chi$, with the same rates. This makes the coverage condition compatible with the spectral and approximation lower bounds, which require each singular hyperplane to intersect $U$, where $\chi$ is uniformly positive.

\begin{theorem}[Approximation rates]\label{thm_appr_rate}
Let $d,n\in\NN$, $k\in\NN_0$, $r\leq\frac{d+2k+1}{2}$, $0\leq s<\min\{k+\frac{1}{2},r\}$, and $\{\theta_j^*\}_{j=1}^n\subset\mathcal A_\chi$. Then there exists $h_\delta$, independent of $n$ and $\{\theta_j^*\}_{j=1}^n$, such that, whenever $h\leq h_\delta$,
\begin{equation}\label{eqn_upp_bound}
\inf\limits_{g_n\in L_n^k}\|f-g_n\|_{\mathcal H^s(\Omega')}\lesssim h^{r-s}\|f\|_{\mathcal H^r(\Omega)},\quad f\in\mathcal H^r(\Omega).
\end{equation}
In particular, if $\{\theta_j^*\}_{j=1}^n$ is quasi-uniform on $\mathcal A_\chi$, then with $\alpha=\frac{d+2k+1-2r}{2d}$, there exists $M\eqsim\|f\|_{\mathcal H^r(\Omega)}$ such that
\begin{equation}\label{eqn_upp_coef_bound}
\inf\limits_{g_n\in L_{n,M}^{k,\alpha}}\|f-g_n\|_{\mathcal H^s(\Omega')}\lesssim n^{-\frac{r-s}{d}}\|f\|_{\mathcal H^r(\Omega)},\quad f\in\mathcal H^r(\Omega).
\end{equation}
\end{theorem}

\begin{remark}[Parameter coverage and separation]
The upper bound \eqref{eqn_upp_bound} depends on the fill distance $h$ and requires no antipodal separation. Quasi-uniformity gives $h\eqsim n^{-\frac{1}{d}}$. The spectral stability and lower bounds instead depend on $\underline h$: adding nearly coincident neurons can reduce separation without substantially improving coverage. Under antipodal quasi-uniformity,
\begin{equation}
h\eqsim\underline h\eqsim n^{-\frac{1}{d}},
\end{equation}
so both kinds of estimates can be expressed in terms of the network width.
\end{remark}

With hidden parameters prescribed, approximation reduces to choosing the output coefficients. Continuous $\mathcal H^s$ least squares in $\chi L_n^k$ leads to a system with matrix $K_s$, whose spectrum controls conditioning and coefficient stability.

\begin{theorem}[Eigenvalues of Gram and stiffness matrices]\label{thm_gram}
Let $d,n,k,s$, and $\{\theta_j^*\}_{j=1}^n$ be as in Theorem~\ref{thm_appr_rate}. Then the eigenvalues of $K_s$, ordered nonincreasingly, satisfy
\begin{equation}
\max\left\{\underline h^{2(k-s)+1}\Big(\frac{n}{j}\Big)^{\frac{2(k-s)+1}{d}},(n^{\frac{1}{d}}\underline h)^{d+2(k-s)+1}nj^{-\frac{d+2(k-s)+1}{d}}\right\}\lesssim\lambda_j(K_s)\lesssim nj^{-\frac{d+2(k-s)+1}{d}},\quad 1\leq j\leq n.
\end{equation}
In particular, if $\{\theta_j^*\}_{j=1}^n$ is antipodally quasi-uniform,
\begin{equation}
\lambda_j(K_s)\eqsim nj^{-\frac{d+2(k-s)+1}{d}},\quad 1\leq j\leq n.
\end{equation}
\end{theorem}

The connection with coefficient stability follows from the identity
\begin{equation}\label{eqn_gram_norm_eq}
a^\top K_sa=\|f_n\|_{\mathcal H^s(\RR^d)}^2,\quad f_n=\sum_{j=1}^na_j\varphi_{\theta_j^*},\quad a=(a_1,\dots,a_n)^\top.
\end{equation}
Thus the smallest eigenvalue controls the coefficient norm, giving
\begin{equation}
\|a\|_2\lesssim\underline h^{-(k-s+\frac{1}{2})}\|f_n\|_{\mathcal H^s(\RR^d)},\quad 0\leq s<k+\frac{1}{2},
\end{equation}
including the $\mathcal{L}^2$ estimate at $s=0$. Under antipodal quasi-uniformity, $\kappa(K_s)\eqsim n^{\frac{d+2(k-s)+1}{d}}$. This conditioning depends on the neuron representation: an invertible change of basis preserves the approximation space and best errors but can change the condition number.

The next theorem complements the approximation upper bounds by relating antipodal separation to a lower bound on the approximation error for fixed smooth targets.

\begin{theorem}[Configuration-dependent lower bounds]\label{thm_saturation}
    Let $d,n,k,s$, and $\{\theta_j^*\}_{j=1}^n$ be as in Theorem~\ref{thm_appr_rate}, and suppose that $r>\frac{d+2k+1}{2}$. Then, for every $f\in\mathcal{H}^r(\Omega)$, there exists $n_0$, depending only on $d,r$ and $\|\chi f\|_{\mH^r(\Omega)}\|\chi f\|_{\mL(\Omega)}^{-1}$, such that, whenever $n\geq n_0$,
    \begin{equation}
        \inf\limits_{g_n\in L_n^k}\|f-g_n\|_{\mathcal H^s(\Omega)}\gtrsim\inf\limits_{f_n\in\chi L_n^k}\|\chi f-f_n\|_{\mL(\RR^d)}\gtrsim n^{-\frac12}\underline{h}^{k+\frac12}\|\chi f\|_{\mathcal{L}^2(\Omega)}\geq n^{-\frac12}\underline{h}^{k+\frac12}\| f\|_{\mathcal{L}^2(\Omega')}.
    \end{equation}
    In particular, if $\{\theta_j^*\}_{j=1}^n$ is antipodally quasi-uniform,
    \begin{equation}
        \inf\limits_{g_n\in L_n^k}\|f-g_n\|_{\mathcal H^s(\Omega)}\gtrsim\inf\limits_{f_n\in\chi L_n^k}\|\chi f-f_n\|_{\mL(\RR^d)}\gtrsim n^{-\frac{d+2k+1}{2d}}\|\chi f\|_{\mathcal{L}^2(\Omega)}\geq n^{-\frac{d+2k+1}{2d}}\|f\|_{\mathcal{L}^2(\Omega')}.
    \end{equation}
\end{theorem}

\begin{remark}[Comparison of upper and lower bounds]
For a fixed target $f\in\mathcal H^r(\Omega)$ with $r>\frac{d+2k+1}{2}$, antipodal quasi-uniformity gives
\begin{equation}
\begin{aligned}
\inf_{g_n\in L_n^k}\|f-g_n\|_{\mathcal L^2(\Omega')}&\lesssim n^{-\frac{d+2k+1}{2d}}\|f\|_{\mathcal H^{\frac{d+2k+1}{2}}(\Omega)},\\
\inf_{g_n\in L_n^k}\|f-g_n\|_{\mathcal L^2(\Omega)}&\gtrsim n^{-\frac{d+2k+1}{2d}}\|\chi f\|_{\mathcal L^2(\RR^d)},
\end{aligned}
\end{equation}
The lower bound holds for sufficiently large $n$, with the threshold depending on $f$. The powers of $n$ match, but the domains differ: the upper bound is on $\Omega'$ and the lower bound on $\Omega$. Thus the estimates give matching orders rather than a two-sided bound on one domain.
\end{remark}

Bernstein inequalities are fundamental tools in approximation theory, controlling stronger norms of approximants in terms of weaker norms and the approximation scale.

\begin{theorem}[Bernstein inequality]\label{thm_bernstein}
Let $d,n,k,s$, and $\{\theta_j^*\}_{j=1}^n$ be as in Theorem~\ref{thm_appr_rate}. Then, for $s<r<k+\frac{1}{2}$,
\begin{equation}
\|f_n\|_{\mH^r(\RR^d)}\lesssim\underline h^{-(r-s)}\|f_n\|_{\mH^s(\RR^d)},\quad f_n\in\chi L_n^k.
\end{equation}
\end{theorem}

The restriction $r<k+\frac{1}{2}$ reflects the Sobolev regularity of the individual localized neurons. This threshold differs from the target smoothness $\frac{d+2k+1}{2}$ allowed in Theorem~\ref{thm_appr_rate}: the approximation spaces can exploit target regularity beyond the regularity of their basis functions.

Marcinkiewicz--Zygmund inequalities provide stable discretizations of continuous norms. Given $X=\{x_i\}_{i=1}^m$, the following theorem applies the equal-weight norm
\begin{equation}
\|f\|_{\ell^2(X)}^2:=\frac{1}{m}\sum_{i=1}^mf(x_i)^2.
\end{equation}
to the localized network spaces.

\begin{theorem}[Marcinkiewicz--Zygmund inequality]\label{thm_mz}
    Let $d,n,k$, and $\{\theta_j^*\}_{j=1}^n$ be as in Theorem~\ref{thm_appr_rate}, and fix $\frac d2<\tau<\frac{d+1}{2}$. There exists $C_1>0$ such that, if $X=\{x_i\}_{i=1}^m\subset\Omega$ is quasi-uniform and
    \begin{equation}\label{eqn_mz_sample_size_tau}
        m\geq C_1
        \begin{cases}
            \underline h^{-d\left(1+ \frac{(d-1)(\tau-k-\frac12)} {2\tau(k+1)-2k-1}\right)}, &k\leq\frac{d-1}{2},\\[0.4em]
            \underline h^{-d},&k>\frac{d-1}{2},
        \end{cases}
    \end{equation}
    then
    \begin{equation}
        \|f_n\|_{\ell^2(X)} \eqsim\|f_n\|_{\mL(\RR^d)} =\|f_n\|_{\mL(\Omega)}, \qquad f_n\in\chi L_n^k.
    \end{equation}
    The constants depend on $d,k,\tau,\Omega,U,\Omega',\chi$, and the quasi-uniformity constant of $X$.
\end{theorem}

Under antipodal quasi-uniformity and $k>\frac{d-1}{2}$, the order $m\gtrsim n$ is optimal for an MZ inequality: the lower bound requires the evaluation map into $\RR^m$ to be injective, hence $m\geq n$. For $k\leq\frac{d-1}{2}$, \eqref{eqn_mz_sample_size_tau} is sufficient; we do not establish the sharpness of the extra oversampling exponent.

The preceding Bernstein and MZ inequalities concern parameters in $\mathcal A_\chi$. For discrete least-squares recovery, we use parameters covering the full sphere and add a quadratic coefficient penalty. To be precise, we denote the discrete least square solution
\begin{equation}\label{eq:domain-ls-objective}
f_{n,m}=\sul_{j=1}^n\alpha_j\varphi_{\theta_j^*},
\end{equation}
where $\boldsymbol\alpha$ is the unique minimizer of
$$\boldsymbol\alpha=\arg\min_{a\in\RR^n}\left\{\frac{1}{m}\sum_{i=1}^m\Big(\chi(x_i)f(x_i)-\sum_{j=1}^na_j\varphi_{\theta_j^*}(x_i)\Big)^2+h^{2k+1}\|a\|_2^2\right\}.$$

The following result separates the roles of parameter coverage, antipodal separation, and sample density.

\begin{theorem}[Regularized discrete least squares]\label{thm_ls}
Let $d,n,k,r,s$ be as in Theorem~\ref{thm_appr_rate}, and $\{\theta_j^*\}_{j=1}^n\subset\SS^d$. There exist $C_5>0$ such that for any 
$$m\geq C_5\big(h^{\frac{2k+1}{d+2k+1}}\underline h^{\frac{d}{d+2k+1}}\big)^{-d}$$
and $X=\{x_i\}_{i=1}^m\subset\Omega$ being quasi-uniform, the unique minimizer $f_{n,m}$ of \eqref{eq:domain-ls-objective} satisfies
\begin{equation}\label{eq:domain-ls-rate}
\|f-f_{n,m}\|_{\mathcal H^s(\Omega')}\lesssim h^{r-s}\Big(\frac{h}{\underline h}\Big)^{\frac{ds}{d+2k+1}}
\begin{cases}
\|f\|_{\mathcal W^{r,p}(\Omega)},&\frac{d}{p}<r\leq\frac{d}{2},\quad 2<p<\infty,\\
\|f\|_{\mathcal H^r(\Omega)},&\frac{d}{2}<r\leq\frac{d+2k+1}{2}.
\end{cases}
\end{equation}
The constants depend only on $d,k,r,s,p,\Omega,\Omega',\chi$, and the quasi-uniformity constant of $X$, independently of $n,m$, and the parameter configuration.
\end{theorem}

\begin{remark}[Parameter geometry and computation]
The penalty strength depends on parameter coverage rather than on the target regularity or sample locations. Writing $A_{ij}:=\varphi_{\theta_j^*}(x_i)$ and $y_i:=\chi(x_i)f(x_i)$, the coefficients satisfy
\begin{equation}
\left(\frac{1}{m}A^\top A+h^{2k+1}I_n\right)\boldsymbol\alpha=\frac{1}{m}A^\top y.
\end{equation}
This system is positive definite even if the localized neurons are linearly dependent. On $\Omega'$, where $\chi=1$, the approximant is the shallow network $\sum_{j=1}^n\alpha_j\sigma_k(w_j^*\cdot x+b_j^*)$.
\end{remark}

Finally, we ask when approximation rates imply target regularity under coefficient or separation conditions.

\begin{theorem}[Inverse approximation]\label{thm_inverse}
Let $d\in\NN,~k\in\NN_0$, $0<\alpha<\frac{d+2k+1}{2d}$, and let $\{\Theta_n\}_{n=1}^\infty$ be a sequence of antipodally quasi-uniform sets on $\mathcal{A}_\chi$.
\begin{enumerate}
\item\label{item1} Let $f\in\mathcal L^2(\RR^d)$. If there exists $M>0$, independent of $n$, such that
\begin{equation}\label{eq:euclidean-constrained-rate}\inf_{f_n\in\chi L_{n,M}^{k,\alpha}(\Theta_n)}\|f-f_n\|_{\mathcal L^2(\RR^d)}\lesssim n^{-\frac{d+2k+1}{2d}+\alpha},\end{equation}
then
\begin{equation*}f\in\mathcal B_{2,\infty}^{\frac{d+2k+1}{2}-\alpha d}(\RR^d).\end{equation*}

\item\label{item2} For every $M>0$, there exists
\begin{equation*}f\in\mathcal L^2(\RR^d)\setminus\mathcal B_{2,\infty}^{\frac{d+2k+1}{2}-\alpha d}(\RR^d)\end{equation*}
such that
\begin{equation}\label{eq:euclidean-mere-convergence}\lim_{n\to\infty}\inf_{f_n\in\chi L_{n,M}^{k,\alpha}(\Theta_n)}\|f-f_n\|_{\mathcal L^2(\Omega')}=0.\end{equation}

\item\label{item3} Suppose further that $\Theta_n\subset\Theta_{n+1}$ and
\begin{equation*}\underline h_n:=\min_{i\neq j}\rho_\pm(\theta_{i,n}^*,\theta_{j,n}^*)\gtrsim n^{-\frac1d},\qquad n\geq2.\end{equation*}
If $\alpha>\frac12$, then every $f\in\mathcal L^2(\RR^d)$ satisfying
\begin{equation}\label{eqn:rate_alpha}\inf_{f_n\in\chi L_n^k(\Theta_n)}\|f-f_n\|_{\mathcal L^2(\RR^d)}\lesssim n^{-\frac{d+2k+1}{2d}+\alpha}\end{equation}
belongs to $\mathcal B_{2,\infty}^{\frac{d+2k+1}{2}-\alpha d}(\RR^d)$. If $0<\alpha\leq\frac12$, there exists $f\in\mathcal L^2(\RR^d)$ satisfying \eqref{eqn:rate_alpha} but not belonging to this Besov space.

\item\label{item4} Let $f\in\mathcal L^2(\RR^d)$, and suppose there exists another parameter sequence $\widetilde\Theta_n=\{\widetilde\theta_{j,n}^*\}_{j=1}^n\subset\mathcal A_\chi$ such that
\begin{equation*}\#(\Theta_n\cup\widetilde\Theta_n)=2n,\qquad \min_{\substack{\theta,\theta'\in\Theta_n\cup\widetilde\Theta_n\\\theta\neq\theta'}}\rho_\pm(\theta,\theta')\gtrsim n^{-\frac1d},\end{equation*}
and
\begin{equation}\label{eq:euclidean-two-rates}\inf_{f_n\in\chi L_n^k(\Theta_n)}\|f-f_n\|_{\mathcal L^2(\RR^d)}+\inf_{\widetilde f_n\in\chi L_n^k(\widetilde\Theta_n)}\|f-\widetilde f_n\|_{\mathcal L^2(\RR^d)}\lesssim n^{-\frac{d+2k+1}{2d}+\alpha}.\end{equation}
Then $f\in\mathcal B_{2,\infty}^{\frac{d+2k+1}{2}-\alpha d}(\RR^d)$.
\end{enumerate}
\end{theorem}

\begin{remark}[Coefficient and separation conditions]
Item~\ref{item1} requires the coefficient bound
\begin{equation}
\Big(n\sum_{j=1}^n a_j^2\Big)^{\frac{1}{2}}\leq Mn^\alpha
\end{equation}
with $M$ independent of $n$. Its error is measured on $\RR^d$, while item~\ref{item2} concerns only qualitative convergence on $\Omega'$. Item~\ref{item4} requires no coefficient bound, but its union must contain $2n$ distinct parameters with the stated antipodal separation, including cross-family pairs; separate quasi-uniformity is insufficient.
\end{remark}

\begin{remark}[Sobolev regularity at the endpoint]
At $\alpha=0$, \cite[Theorem~2.4]{liu2025integral} gives the stronger conclusion
\begin{equation}
\lim_{n\to\infty}\inf_{g_n\in L_{n,M}^{k,0}}\|f-g_n\|_{\mathcal L^2(\Omega)}=0\quad\Longrightarrow\quad f\in\mathcal H^{\frac{d+2k+1}{2}}(\Omega),
\end{equation}
for $M$ independent of $n$. This result assumes parameters quasi-uniform on the entire $\SS^d$ and qualitative convergence on $\Omega$; the present theorem concerns parameters in $\mathcal A_\chi$ and whole-space approximation by localized networks.
\end{remark}

\section{Localization of the Littlewood–Paley Sobolev blocks}\label{sec_loc}

This section proves the two parts of Theorem~\ref{thm:localized-estimates}. We retain their separate statements below to organize the proofs.

In Sections \ref{sec_loc} and \ref{sec_proof}, we assume that $\Omega\subset\BB^d$. The case of a general bounded $\Omega$ follows by rescaling the spatial variable. We switch from the spherical parametrization to the unit-normal parametrization. For $\theta=\binom{w}{b}\in\mathcal A_\chi$, the active condition implies $|w|\ge c_3>0$; hence the normalization map
\[
\binom{w}{b}\longmapsto\left(\frac{w}{|w|},\frac{b}{|w|}\right)
\]
is well defined on $\mathcal A_\chi$. By positive homogeneity,
\[
\sigma_k(w^\top x+b)=|w|^k\sigma_k\left(\frac{w}{|w|}^\top x+\frac{b}{|w|}\right),
\]
and the normalized parameters form a compact subset $\mathcal B_\chi\subset\mathbb S^{d-1}\times[-1,1]$. Since the spherical distance on $\mathbb S^d$ is equivalent to the Euclidean chordal distance and the above normalization is bi-Lipschitz on the active set, we use the normalized parametrization throughout this section. For $\theta=\binom{w}{b},\theta'=(w',b')\in\mathcal B_\chi$, define
\[
\varrho(\theta,\theta'):=\left(|w-w'|^2+|b-b'|^2\right)^{1/2}
\]
and
\[
\varrho_\pm(\theta,\theta'):=\min\{\varrho(\theta,\theta'),\varrho(\theta,-\theta')\}.
\]
In the Fourier proofs below, separation is measured by $\varrho$ and $\varrho_\pm$, with $|w|=1$ for $\theta=\binom{w}{b}\in\mathcal B_\chi$. Positive homogeneity and the bi-Lipschitz normalization transfer the estimates to $\mathcal A_\chi$.

\begin{theorem}[Localization property]
\label{thm:gram}
For every $0\leq s<k+\frac{1}{2},~\gamma>d$, and $\theta,\theta'\in\mathcal A_\chi$ with $\rho_\pm(\theta,\theta')>0$,
\begin{equation}
\label{eq:gram-estimate}
\left|
    \left\langle
        \Delta_{q,s}\varphi_{\theta},
        \Delta_{q,s}\varphi_{\theta'}
    \right\rangle_{\mL(\mathbb R^d)}
\right|
\lesssim_\gamma\frac{2^{-q(2(k-s)+1)}}{
\left(
    2^q \rho_\pm(\theta,\theta')
\right)^{\gamma}},\qquad q\geq1.
\end{equation} 
\end{theorem}

\subsection{Active-parameter normalization and the high-frequency representation}

We first express the inner product in \eqref{eq:gram-estimate} using Parseval's identity:
\begin{equation}
\begin{split}
    \left\langle\Delta_{q,s}\varphi_{\theta},\Delta_{q,s}\varphi_{\theta'}\right\rangle_{\mL(\mathbb R^d)}=&\frac{1}{(2\pi)^d}\int_{\RR^d}\zeta_{q,s}(\xi)^2\widehat{\varphi_{\theta}}(\xi)\overline{\widehat{\varphi_{\theta'}}(\xi)}\,d\xi\\
    =&\frac{1}{(2\pi)^d}\int_{\RR^d}\zeta\Big(\frac{|\xi|}{2^q}\Big)^2(1+|\xi|^2)^s\widehat{\varphi_{\theta}}(\xi)\overline{\widehat{\varphi_{\theta'}}(\xi)}\,d\xi.
\end{split}
\end{equation}

We derive a Fourier representation of $\varphi_\theta$ from the distributional Fourier transform of $\sigma_k$. The activation has polynomial growth and is not in
$\mL(\mathbb R)$, so its transform is a tempered distribution. For $k=0$, the inversion identity below is understood distributionally and almost everywhere; symmetric inversion assigns the midpoint value at the jump. Substituting
$t=w^\top x+b$ into the Fourier inversion formula
\[
    \sigma_k(t)
    =
    \frac{1}{2\pi}\int_{\mathbb R}
    \widehat{\sigma_k}(\tau)e^{i\tau t}
    \,d\tau
\]
yields
\begin{equation}
\label{eq:one-dimensional-ridge-fourier}
\begin{aligned}
\sigma_k(w^\top x+b)=\frac{1}{2\pi}\int_{\mathbb R}
\widehat{\sigma_k}(\tau)
e^{i\tau(w^\top x+b)}
\,d\tau=
\frac{1}{2\pi}\int_{\mathbb R}
\widehat{\sigma_k}(\tau)
e^{i\tau b}
e^{i(\tau w)^\top x}
\,d\tau.
\end{aligned}
\end{equation}
Using the standard modulation formula for the Fourier transform,
\begin{equation*}
\mathcal F[\chi(\circ)e^{i(\tau w)^\top\circ}](\xi)=\widehat\chi(\xi-\tau w),
\end{equation*}
we have
\begin{equation*}
    \begin{split}
        \varphi_\theta(x)=&\chi(x)\sigma_k(w^\top x+b)=\frac{1}{2\pi}\int_{\mathbb R}\widehat{\sigma_k}(\tau)e^{i\tau b}e^{i(\tau w)^\top x}\chi(x)d\tau\\
        =&\frac{1}{(2\pi)^{d+1}}\int_{\mathbb R}\widehat{\sigma_k}(\tau)e^{i\tau b}\int_{\RR^d}\widehat\chi(\xi-\tau w)e^{i\xi\cdot x}d\xi d\tau.
    \end{split}
\end{equation*}
That is,
\begin{equation}
    \widehat{\varphi_\theta}(\xi)
    =\frac{1}{2\pi}\int_{\mathbb R}\widehat{\sigma_k}(\tau)e^{i\tau b}\widehat\chi(\xi-\tau w)d\tau.
\end{equation}
We split the Fourier transform into three parts. The first is estimated through pairwise interactions; the other two decay rapidly. Let $\vartheta,\varsigma\in \mathcal C^\infty(\mathbb R)$ be smooth symmetric functions satisfying
\begin{equation*}
    \begin{split}
        &\vartheta(\tau)=0
    \quad\text{for }|\tau|\leq1,
    \qquad
    \vartheta(\tau)=1
    \quad\text{for }|\tau|\geq2,\\
    &\varsigma(\tau)=0\quad\text{for }|\tau|\leq\frac18\ \text{and}\ |\tau|\geq8,\qquad
    \varsigma(\tau)=1
    \quad\text{for }\frac14\leq|\tau|\leq4
    \end{split}
\end{equation*}
We then write
\begin{equation}\label{eqn_varphi_decom}
    \widehat{\varphi_\theta}(\xi)=\frac{1}{2\pi}\big(I_1(\theta,q;\xi)+I_2(\theta,q;\xi)+I_3(\theta;\xi)\big),
\end{equation}
where
\begin{equation}\label{eqn_def_I123}
    \begin{split}
        &I_1(\theta,q;\xi)=\int_\RR\varsigma\big(\frac{\tau}{2^q}\big)\vartheta(\tau)\widehat{\sigma_k}(\tau)e^{i\tau b}\widehat\chi(\xi-\tau w)d\tau,\\
        &I_2(\theta,q;\xi)=\int_\RR\Big(1-\varsigma\big(\frac{\tau}{2^q}\big)\Big)\vartheta(\tau)\widehat{\sigma_k}(\tau)e^{i\tau b}\widehat\chi(\xi-\tau w)d\tau,\\
        &I_3(\theta;\xi)=\int_{\mathbb R}(1-\vartheta(\tau))\widehat{\sigma_k}(\tau)e^{i\tau b}\widehat\chi(\xi-\tau w)d\tau
    \end{split}
\end{equation}

In the sense of distributions,
\[
    \partial_t^{k+1}\sigma_k
    =
    k!\,\delta_0.
\]
Consequently, away from the origin,
\begin{equation}\label{eqn_sigma_fourier}
    \widehat{\sigma_k}(\tau)=k!(i\tau)^{-k-1},
\end{equation}
Thus $\widehat{\sigma_k}$ is homogeneous of degree $-(k+1)$ away from the origin, and $\vartheta\widehat{\sigma_k}$ is an ordinary smooth function. For every integer
$\ell\geq0$,
\begin{equation}
\label{eq:multiplier-decay}
    \Big|\Big(\frac{d}{d\tau}\Big)^\ell(\vartheta\widehat{\sigma_k})(\tau)\Big|
    \lesssim_\ell|\tau|^{-k-1-\ell},
    \qquad |\tau|\geq1.
\end{equation}

\begin{lemma}\label{lem_decay_I3}
For every $\gamma>0$,
\[
    \left|I_3(\theta;\xi)\right|
    \lesssim_\gamma(1+|\xi|)^{-\gamma},
\]
uniformly for $\theta\in\mathcal B_\chi$.
\end{lemma}
\begin{proof}
Applying the Fourier modulation formula again, we obtain
\[
\mathcal F^{-1}\left[I_3(\theta;\circ)\right](x)=2\pi\chi(x)
\mathcal F^{-1}
\left[
    (1-\vartheta)\widehat{\sigma_k}
\right](w^\top x+b).
\]
    The compactly supported distribution $(1-\vartheta)\widehat{\sigma_k}$ has finite order, so its inverse transform is smooth. Hence $\mathcal F^{-1}[I_3(\theta;\circ)]$ is smooth and supported in $\Omega$, with all seminorms $|\mathcal F^{-1}[I_3(\theta;\circ)]|_{\mathcal C^\gamma(\RR^d)}$ uniformly bounded for $\theta\in\mathcal B_\chi$. Its Fourier transform therefore decays rapidly:
    \begin{equation*}
        |I_3(\theta;\xi)|
    \lesssim_\gamma(1+|\xi|)^{-\gamma},\quad\xi\in\RR^d.
    \end{equation*}
\end{proof}

\begin{lemma}\label{lem_decay_I2}
For every $\gamma>0$,
\begin{equation}\label{eq:tau-localization}
    \left\|\zeta_{q,s}(\circ)I_2(\theta,q;\circ)\right\|_{\mL(\RR^d)}\lesssim_\gamma2^{-q\gamma},\quad q\geq1.
\end{equation}
uniformly for $\theta\in\mathcal B_\chi$. Thus we may insert smooth cutoffs in $\tau$
to restrict $|\tau|\eqsim2^q$, with a remainder that decays uniformly
faster than any power of $2^{-q}$.
\end{lemma}

\begin{proof}
By \eqref{eqn_def_zeta}, $\zeta_{q,s}(\circ)I_2(\theta,q;\circ)$ is supported in $\{\xi\in\RR^d:2^{q-1}\leq|\xi|\leq2^{q+1}\}$.

If $|\tau|\leq 2^{q-2}$, then, since $|w|=1$, we have
$$|\xi-\tau w|
\geq|\xi|-|\tau|
\geq
2^{q-2}.$$
Since $\chi\in\mathcal{C}^\infty_c(\RR^d)$, $\widehat\chi$ is a Schwartz function:
\begin{equation}
\label{eq:normal-line-concentration}
    |\widehat\chi(\xi-\tau w)|
    \lesssim
    (1+|\xi-\tau w|)^{-\gamma-1}\lesssim
    2^{-q(\gamma+1)}.
\end{equation}
As $\vartheta=0$ for $|\tau|\leq1$, the bound \eqref{eq:multiplier-decay} gives
\begin{equation*}
    \int_{-2^{q-2}}^{2^{q-2}}\left|\vartheta(\tau)\widehat{\sigma_k}(\tau)e^{i\tau b}\widehat\chi(\xi-\tau w)\right|d\tau\lesssim_\gamma\int_{-2^{q-2}}^{2^{q-2}}2^{-q(\gamma+1)}d\tau\lesssim2^{-q\gamma}.
\end{equation*}

If $|\tau|\geq2^{q+2}$, then
$$|\xi-\tau w|\geq
|\tau|-|\xi|\geq
|\tau|-2^{q+1}\geq
2^{q+1}.$$
Using \eqref{eq:multiplier-decay} and
\eqref{eq:normal-line-concentration}, we obtain
\begin{equation*}
\begin{split}
        \int_{|\tau|\geq 2^{q+2}}\left|
\vartheta(\tau)\widehat{\sigma_k}(\tau)e^{i\tau b}
\widehat\chi(\xi-\tau w)\right|d\tau
\lesssim\int_{|\tau|\geq 2^{q+2}}
|\tau|^{-k-1}
(1+|\tau|)^{-\gamma}
\,d\tau\lesssim
2^{-q(k+\gamma)}.
\end{split}
\end{equation*}
Thus $I_2(\theta,q;\circ)$ decays pointwise to any prescribed order on $|\xi|\in[2^{q-1},2^{q+1}]$. Increasing that order by $\lceil \frac{d}{2}+s\rceil$ gives \eqref{eq:tau-localization}.
\end{proof}

\subsection{Preliminary estimates for $I_1$}
\label{subsec:coherent-separation}
Recall that the normalization above gives
$$|w|=|w'|=1.$$
We aim to show
\begin{equation}\label{eqn_I1_prod}
    \left|\int_{\RR^d}\zeta_{q,s}(\xi)^2I_1(\theta,q;\xi)\overline{I_1(\theta',q;\xi)}d\xi\right|\lesssim\frac{2^{-q(2(k-s)+1)}}{\left(
    2^q \varrho_\pm(\theta,\theta')\right)^{\gamma}}.
\end{equation}
It suffices to prove the estimate for $q\geq4$. On the support of the cutoffs defining $I_1$, $\vartheta(\tau)=\vartheta(\tau')=1$, so
\begin{equation*}
    \begin{split}
        &\int_{\RR^d}\zeta_{q,s}(\xi)^2I_1(\theta,q;\xi)\overline{I_1(\theta',q;\xi)}d\xi\\
        =&\int_{\RR^d}\zeta_{q,s}(\xi)^2\iint_{\RR^2}\varsigma\big(\frac{\tau}{2^q}\big)\varsigma\big(\frac{\tau'}{2^q}\big)\widehat{\sigma_k}(\tau)\overline{\widehat{\sigma_k}(\tau')}e^{i(\tau b-\tau' b')}\widehat\chi(\xi-\tau w)\overline{\widehat\chi(\xi-\tau'w')}d\tau d\tau' d\xi.
    \end{split}
\end{equation*}
Let $\varsigma_+(t)=\left\{\begin{array}{ll}
    \varsigma(t), &\quad t\geq0,  \\
    0, &\quad t<0, 
\end{array}\right.$ and define $J_+(\theta,\theta',q)$ and $J_-(\theta,\theta',q)$ by
\begin{equation}\label{eqn_def_J+}
    \begin{split}
        &J_+(\theta,\theta',q)=\iint_{\RR^2}\varsigma_+\big(\frac{\tau}{2^q}\big)\varsigma_+\big(\frac{\tau'}{2^q}\big)e^{i(\tau b-\tau' b')}\int_{\RR^d}\zeta_{q,s}(\xi)^2\widehat{\sigma_k}(\tau)\overline{\widehat{\sigma_k}(\tau')}\widehat\chi(\xi-\tau w)\overline{\widehat\chi(\xi-\tau'w')}d\xi d\tau d\tau',\\
        &J_-(\theta,\theta',q)=\iint_{\RR^2}\varsigma_+\big(\frac{\tau}{2^q}\big)\varsigma_+\big(-\frac{\tau'}{2^q}\big)e^{i(\tau b-\tau' b')}\int_{\RR^d}\zeta_{q,s}(\xi)^2\widehat{\sigma_k}(\tau)\overline{\widehat{\sigma_k}(\tau')}\widehat\chi(\xi-\tau w)\overline{\widehat\chi(\xi-\tau'w')}d\xi d\tau d\tau'.
    \end{split}
\end{equation}

Using the symmetry properties of $\vartheta$, $\widehat{\sigma_k}$, $\widehat\chi$, and the integration domain, we obtain
\begin{equation*}
    \begin{split}
        &\iint_{\RR^2}\varsigma_+\big(-\frac{\tau}{2^q}\big)\varsigma_+\big(-\frac{\tau'}{2^q}\big)\int_{\RR^d}\zeta_{q,s}(\xi)^2\widehat{\sigma_k}(\tau)\overline{\widehat{\sigma_k}(\tau')}e^{i(\tau b-\tau' b')}\widehat\chi(\xi-\tau w)\overline{\widehat\chi(\xi-\tau'w')}d\xi d\tau d\tau'=\overline{J_+(\theta,\theta',q)},\\
        &\iint_{\RR^2}\varsigma_+\big(-\frac{\tau}{2^q}\big)\varsigma_+\big(\frac{\tau'}{2^q}\big)\int_{\RR^d}\zeta_{q,s}(\xi)^2\widehat{\sigma_k}(\tau)\overline{\widehat{\sigma_k}(\tau')}e^{i(\tau b-\tau' b')}\widehat\chi(\xi-\tau w)\overline{\widehat\chi(\xi-\tau'w')}d\xi d\tau d\tau'=\overline{J_-(\theta',\theta,q)}.
    \end{split}
\end{equation*}
The substitution $\tau'\mapsto-\tau'$ and the same symmetries give
\begin{equation}\label{eqn_def_J-}
\begin{split}
    J_-(\theta,\theta',q)=(-1)^{k+1}\iint_{\RR^2}\varsigma_+\left(\frac{\tau}{2^q}\right)\varsigma_+\left(\frac{\tau'}{2^q}\right)e^{i(\tau b+\tau'b')}\int_{\mathbb R^d}\zeta_{q,s}(\xi)^2\widehat{\sigma_k}(\tau)\overline{\widehat{\sigma_k}(\tau')}\widehat\chi(\xi-\tau w)\overline{\widehat\chi(\xi+\tau'w')}d\xi d\tau d\tau'.
\end{split}
\end{equation}
Thus both $J_+$ and $J_-$ are reduced to interactions between two positive normal frequencies. The two cases differ only in whether the second affine parameter is $(w',b')$ or $(-w',-b')$.

To prove \eqref{eqn_I1_prod}, it therefore suffices to show
\begin{equation}\label{eqn_Jpm_bound}
    |J_+(\theta,\theta',q)|\lesssim\frac{2^{-q(2(k-s)+1)}}{\left(
    2^q \varrho_\pm(\theta,\theta')\right)^{\gamma}},\quad |J_-(\theta,\theta',q)|\lesssim\frac{2^{-q(2(k-s)+1)}}{\left(
    2^q \varrho_\pm(\theta,\theta')\right)^{\gamma}}.
\end{equation}

We estimate these terms using direction separation and bias separation. Since
\[
\varrho_\pm(\theta,\theta')
=
\min\{
\varrho(\theta,\theta'),
\varrho(\theta,-\theta')\}\eqsim
\min\left\{
|w-w'|+|b-b'|,
|w+w'|+|b+b'|
\right\},
\]
the separation of the two affine parameters can be detected either
by the directional component or by the bias component.

\paragraph{Case 1:} When $|w-w'|\gtrsim\varrho(\theta,\theta')\geq\varrho_\pm(\theta,\theta')$, the decay of $J_+(\theta,\theta',q)$ is obtained from the overlap of the two Fourier
packets
\[
    \widehat\chi(\xi-\tau w),
    \qquad
    \widehat\chi(\xi-\tau'w').
\]
Indeed, under dyadic localization, $|\tau|\eqsim |\tau'|\eqsim 2^q$, so the distance between the centers of the two packets is
given by
\[
    |\tau w-\tau'w'|
    \quad\text{or}\quad
    |\tau w+\tau'w'|,
\]
depending on the corresponding sign. Since $\widehat\chi$ is a Schwartz
function, the overlap of the two packets decays rapidly as this
distance increases, which provides the decay factor $(1+2^q\varrho_\pm(\theta,\theta'))^{-\gamma}$.

\paragraph{Case 2:} When $|b-b'|\gtrsim\varrho(\theta,\theta')\geq\varrho_\pm(\theta,\theta')$, the decay must come from the oscillation of the phase factor
\[
    e^{i(\tau b-\tau'b')}.
\]
Such oscillatory cancellation is standard in the literature (see, e.g., \cite[Chapter 8]{stein1993harmonic}).

To separate the oscillatory variable from the frequency mismatch, we
introduce the change of variables
\[
    u=\tau+\tau',
    \qquad
    v=\tau-\tau'.
\]
Then
\[
    \tau=\frac{u+v}{2},
    \qquad
    \tau'=\frac{u-v}{2},
    \qquad
    d\tau\,d\tau'=\frac12\,du\,dv .
\]
For the two coherent sign configurations, the phase can be written as
$$
    \tau b-\tau'b'
 =
    \frac{u}{2}(b-b')
    +
    \frac{v}{2}(b+b'),\quad\tau b+\tau'b'=
    \frac{u}{2}(b+b')
    +
    \frac{v}{2}(b-b').
$$
A similar argument applies to $J_-(\theta,\theta',q)$. Thus, the two coherent contributions \eqref{eqn_def_J+} and \eqref{eqn_def_J-} can be written as
\begin{equation}
\label{eq:coherent-amplitude-representation}
J_\pm(\theta,\theta',q)=\frac12\int_{\mathbb R}\int_{\mathbb R}e^{\frac{iu}{2}(b\mp b')}e^{\frac{iv}{2}(b\pm b')}A_{q,\pm}(u,v)dudv,
\end{equation}
where
\begin{equation}
\begin{aligned}
A_{q,\pm}(u,v)=&(\pm1)^{k+1}\varsigma_+\left(\frac{u+v}{2^{q+1}}\right)\varsigma_+\left(\frac{u-v}{2^{q+1}}\right)\vartheta\left(\frac{u+v}{2}\right)\widehat{\sigma_k}\left(\frac{u+v}{2}\right)\vartheta\left(\frac{u-v}{2}\right)\overline{\widehat{\sigma_k}\left(\frac{u-v}{2}\right)}\\
&\times\int_{\mathbb R^d}\zeta_{q,s}(\xi)^2\widehat\chi\left(\xi-\frac{u+v}{2}w\right)\overline{\widehat\chi\left(\xi\mp\frac{u-v}{2}w'\right)}\,d\xi.
\end{aligned}
\end{equation}

\begin{lemma}
\label{lem:coherent-amplitude}
For all integers $\gamma_1,\gamma_2\geq0$,
\begin{equation}
\label{eq:amplitude-bound}
\begin{split}
    &\iint_{\mathbb R^2}\left|\partial_u^{\gamma_1} A_{q,\pm}(u,v)
\right|dudv
\lesssim_{\gamma_1,\gamma_2}2^{-q(2(k-s)+1)}(2^{-q}+|w\mp w'|)^{\gamma_1}(1+2^q|w\mp w'|)^{-\gamma_2}.
\end{split}
\end{equation}
\end{lemma}
\begin{proof}
For either sign, Leibniz's rule reduces the derivative estimates for $A_{q,\pm}$ to the following factors:
\begin{align*}
&K_{1,q,\pm}(u,v)=\int_{\mathbb R^d}\zeta_{q,s}(\xi)^2\widehat\chi\left(\xi-\frac{u+v}{2}w\right)\overline{\widehat\chi\left(\xi\mp\frac{u-v}{2}w'\right)}d\xi,\\
&K_{2,q}(u,v)=\vartheta\left(\frac{u+v}{2}\right)\widehat{\sigma_k}\left(\frac{u+v}{2}\right)\vartheta\left(\frac{u-v}{2}\right)\overline{\widehat{\sigma_k}\left(\frac{u-v}{2}\right)},\\
&K_{3,q}(u,v)=\varsigma_+\big(\frac{u+v}{2^{q+1}}\big)\varsigma_+\big(\frac{u-v}{2^{q+1}}\big).
\end{align*}
Making the change of variables $\eta=\xi-\frac{u+v}{2}w$, we write
\begin{equation*}
K_{1,q,\pm}(u,v)=\int_{\mathbb R^d}\zeta_{q,s}\left(\eta+\frac{u+v}{2}w\right)^2\widehat\chi(\eta)\overline{\widehat\chi\left(\eta+\frac{u+v}{2}w\mp\frac{u-v}{2}w'\right)}d\eta,
\end{equation*}
and by Leibniz's rule,
\begin{equation*}
    \begin{split}
        |\partial_u^\beta K_{1,q,\pm}(u,v)|\lesssim&\sul_{\beta_1+\beta_2=\beta}\int_{\mathbb R^d}\left|\partial_u^{\beta_1}\left[\zeta_{q,s}\left(\eta+\frac{u+v}{2}w\right)^2\right]\partial_u^{\beta_2}\left[\widehat\chi(\eta)\overline{\widehat\chi\left(\eta+\frac{u+v}{2}w\mp\frac{u-v}{2}w'\right)}\right]\right|d\eta\\
        \lesssim&\sul_{\beta_1+\beta_2=\beta}\int_{\mathbb R^d}2^{q(2s-\beta_1)}\left|\widehat\chi(\eta)\overline{\left[\left(\frac{w\mp w'}{2}\cdot\nabla\right)^{\beta_2}\widehat\chi\right]\left(\eta+\frac{u+v}{2}w\mp\frac{u-v}{2}w'\right)}\right|d\eta.
    \end{split}
\end{equation*}
On the support of $K_{3,q}$, the relations
$$\tau=\frac{u+v}{2},\qquad\tau'=\frac{u-v}{2}$$
give $\tau\eqsim\tau'\eqsim2^q>0$, and hence
\begin{equation*}
    \left|\frac{u+v}{2}w\mp\frac{u-v}{2}w'\right|=|\tau w\mp\tau'w'|=\left(|\tau-\tau'|^2+\tau\tau'|w\mp w'|^2\right)^{1/2}\gtrsim|v|+2^q|w\mp w'|.
\end{equation*}
Here $\nabla^{\beta_2}\widehat\chi$ denotes the collection of derivatives of order $\beta_2$. Schwartz decay gives
\begin{equation*}
    |\widehat\chi(\eta)|\left|\nabla^{\beta_2}\widehat\chi\left(\eta+\frac{u+v}{2}w\mp\frac{u-v}{2}w'\right)\right|\lesssim_{\beta_2,\gamma_2}(1+|\eta|)^{-d-1}(1+|v|+2^q|w\mp w'|)^{-\gamma_2-2}.
\end{equation*}
Since $\int_{\mathbb R^d}(1+|\eta|)^{-d-1}d\eta<\infty$, it follows that
\begin{equation*}
    |\partial_u^\beta K_{1,q,\pm}(u,v)|\lesssim_{\beta,\gamma_2}2^{2qs}(2^{-q}+|w\mp w'|)^\beta(1+|v|+2^q|w\mp w'|)^{-\gamma_2-2}.
\end{equation*}

On the other hand, \eqref{eq:multiplier-decay} gives, on the support of $A_{q,\pm}$,
\begin{equation*}
    \begin{split}
        |\partial_u^\alpha K_{2,q}(u,v)|&=\left|\partial_u^\alpha\left[\vartheta\left(\frac{u+v}{2}\right)\widehat{\sigma_k}\left(\frac{u+v}{2}\right)\vartheta\left(\frac{u-v}{2}\right)\overline{\widehat{\sigma_k}\left(\frac{u-v}{2}\right)}\right]\right|\\
        &\lesssim_\alpha\sul_{\alpha_1+\alpha_2=\alpha}\left|\partial_\tau^{\alpha_1}\left(\vartheta(\tau)\widehat{\sigma_k}(\tau)\right)\partial_{\tau'}^{\alpha_2}\left(\vartheta(\tau')\widehat{\sigma_k}(\tau')\right)\right|\lesssim2^{-q(2k+2+\alpha)}.
    \end{split}
\end{equation*}
By Leibniz's rule, on its support,
\begin{equation*}
    \begin{split}
        |\partial_u^{\gamma_1}A_{q,\pm}(u,v)|&\lesssim_{\gamma_1}\sul_{\alpha+\beta+\nu=\gamma_1}|\partial_u^\beta K_{1,q,\pm}(u,v)\partial_u^\alpha K_{2,q}(u,v)\partial_u^\nu K_{3,q}(u,v)|\\
        &\lesssim\sul_{\alpha+\beta+\nu=\gamma_1}2^{2qs}(2^{-q}+|w\mp w'|)^\beta(1+|v|+2^q|w\mp w'|)^{-\gamma_2-2}\times2^{-q(2k+2+\alpha)}\times2^{-q\nu}\\
        &\lesssim(2^{-q}+|w\mp w'|)^{\gamma_1}(1+|v|+2^q|w\mp w'|)^{-\gamma_2-2}2^{-q(2(k-s)+2)}.
    \end{split}
\end{equation*}

Integrating first with respect to $v$, we obtain, for every $\gamma_2\geq0$,
\begin{equation*}
    \int_{\mathbb R}(1+|v|+2^q|w\mp w'|)^{-\gamma_2-2}dv=\frac{2}{\gamma_2+1}(1+2^q|w\mp w'|)^{-\gamma_2-1}\lesssim_{\gamma_2}(1+2^q|w\mp w'|)^{-\gamma_2}.
\end{equation*}
The $u$-interval has length $O(2^q)$, so
\begin{equation*}
    \begin{split}
        \iint_{\mathbb R^2}|\partial_u^{\gamma_1}A_{q,\pm}(u,v)|dudv&\lesssim_{\gamma_1,\gamma_2}\int_{-2^{q+4}}^{2^{q+4}}(2^{-q}+|w\mp w'|)^{\gamma_1}(1+2^q|w\mp w'|)^{-\gamma_2}2^{-q(2(k-s)+2)}du\\
        &\lesssim(1+2^q|w\mp w'|)^{-\gamma_2}2^{-q(2(k-s)+1)}(2^{-q}+|w\mp w'|)^{\gamma_1}.
    \end{split}
\end{equation*}
\end{proof}

\subsection{Proof of Theorem \ref{thm:gram}}

\begin{proof}
By \eqref{eqn_varphi_decom},
\begin{equation*}
        \big|\left\langle
        \Delta_{q,s}\varphi_{\theta},
        \Delta_{q,s}\varphi_{\theta'}
    \right\rangle_{\mL(\mathbb R^d)}\big|\leq\frac{1}{(2\pi)^{d+2}}\sul_{i,j=1,2,3}\left|\int_{\RR^d}\zeta_{q,s}(\xi)^2I_i(\theta,q;\xi)\overline{I_j(\theta',q;\xi)}d\xi\right|,
\end{equation*}
where $I_3(\theta,q;\circ)$ denotes $I_3(\theta;\circ)$. By Lemmas~\ref{lem_decay_I3} and~\ref{lem_decay_I2}, it remains to prove
\begin{equation}\label{eqn_I1_prod_bound}
    \left|\int_{\RR^d}\zeta_{q,s}(\xi)^2I_1(\theta,q;\xi)\overline{I_1(\theta',q;\xi)}d\xi\right|\lesssim\frac{2^{-q(2(k-s)+1)}}{
\left(
    2^q \varrho_\pm(\theta,\theta')
\right)^{\gamma}}.
\end{equation}

For $J_+(\theta,\theta',q)$, we use the two cases in Section~\ref{subsec:coherent-separation}.

\paragraph{Case 1:} $|w-w'|\gtrsim\varrho(\theta,\theta')\geq\varrho_\pm(\theta,\theta')$.

In this case, the decay follows from the overlap of the two Fourier packets
\[
    \widehat\chi(\xi-\tau w),
    \qquad
    \widehat\chi(\xi-\tau'w').
\]
Indeed, applying the estimate in Lemma \ref{lem:coherent-amplitude} with $\gamma_1=0,\gamma_2=\gamma$ to \eqref{eq:coherent-amplitude-representation}, and recalling that $\mathrm{supp}(A_{q,+})\subset[-2^{q+4},2^{q+4}]^2$, we obtain
\begin{equation*}
    |J_+(\theta,\theta',q)|\lesssim2^{-q(2(k-s)+1)}(1+2^q|w- w'|)^{-\gamma}.
\end{equation*}
Together with $|w-w'|\gtrsim\varrho_\pm(\theta,\theta')$, this proves the first estimate in \eqref{eqn_Jpm_bound}.

\paragraph{Case 2:} $|b-b'|\gtrsim\varrho(\theta,\theta')\geq\varrho_\pm(\theta,\theta')$.

In this case, the decay comes from the oscillation of the phase factor $e^{i(\tau b-\tau'b')}$. To see this, we integrate by parts $\gamma$ times in $u$ in \eqref{eq:coherent-amplitude-representation} and apply Lemma \ref{lem:coherent-amplitude} with $\gamma_1=\gamma_2=\gamma$ to obtain
\begin{equation*}
    \begin{split}
        |J_+(\theta,\theta',q)|=&\left|\frac12\Big(\frac{2i}{b- b'}\Big)^\gamma\iint_{\mathbb R^2}e^{\frac{iu}{2}(b- b')}e^{\frac{iv}{2}(b+ b')}\partial_u^\gamma A_{q,+}(u,v)dudv\right|\\
        \lesssim&|b-b'|^{-\gamma}2^{-q(2(k-s)+1)}(2^{-q}+|w- w'|)^{\gamma}(1+2^q|w- w'|)^{-\gamma}\\
        =&2^{-q(2(k-s)+1)}\Big(\frac{1+2^q|w- w'|}{2^q|b-b'|}\Big)^\gamma(1+2^q|w- w'|)^{-\gamma}=2^{-q(2(k-s)+1)}(2^q|b-b'|)^{-\gamma}.
    \end{split}
\end{equation*}
Together with $|b-b'|\gtrsim\varrho_\pm(\theta,\theta')$, this proves the first estimate in \eqref{eqn_Jpm_bound}.

\bigskip

For $J_-(\theta,\theta',q)$, replace $(w',b')$ by $(-w',-b')$ in the same argument. This gives \eqref{eqn_Jpm_bound} for all $\gamma\in\NN_0$, and hence all $\gamma>0$, proving \eqref{eqn_I1_prod_bound} and Theorem~\ref{thm:gram}.

\end{proof}

\subsection{Dyadic estimates for individual localized neurons}

\begin{theorem}\label{thm:diagonal-estimate}
For every $0\leq s<k+\frac{1}{2}$, there exists $q_0\in\mathbb N$ such that
\[
\|\Delta_{q,s}\varphi_\theta\|_{\mathcal{L}^2(\mathbb R^d)}^2\eqsim 2^{-q(2(k-s)+1)},\quad q\ge q_0
\]
uniformly for $\theta\in\mathcal A_\chi$.
\end{theorem}

\begin{proof}
Again, it suffices to consider $\theta=\binom{w}{b}\in\mathcal{B}_\chi$. Fix $\theta$ and, for $y\in w^\perp$, define
\[
\chi_\theta(y)=\chi(y-bw).
\]
Using the orthogonal decomposition
\[
x=y+(t-b)w,\qquad y\in w^\perp,\quad t=w^\top x+b,
\]
we have
\[
\varphi_\theta(y+(t-b)w)=\chi(y+(t-b)w)\sigma_k(t).
\]
By Leibniz's rule and the identity
\[
\partial_t^{k+1}\sigma_k(t)=k!\delta_0
\]
in the sense of distributions,
\[
\partial_t^{k+1}\left[\chi(y+(t-b)w)\sigma_k(t)\right]=k!\chi_\theta(y)\delta_0(t)+R_\theta(y,t),
\]
where
\[
R_\theta(y,t)=\sum_{j=1}^{k+1}\binom{k+1}{j}\frac{k!}{(j-1)!}(w^\top\nabla)^j\chi(y+(t-b)w)t_+^{j-1}.
\]
Since $\chi\in\mathcal C_c^\infty(\mathbb R^d)$ and the normalized active parameter set is bounded, $R_\theta$ is uniformly compactly supported and smooth with respect to $y$, and its derivative with respect to $t$ is a finite measure. Consequently, for every $\gamma>0$,
\[
|\widehat{\chi_\theta}(\eta)|\lesssim_\gamma(1+|\eta|)^{-\gamma},\qquad|\widehat{R_\theta}(\eta,\lambda)|\lesssim_\gamma |\lambda|^{-1}(1+|\eta|)^{-\gamma}
\]
uniformly for $\theta\in\mathcal B_\chi$ and $|\lambda|\geq1$.

Taking the Fourier transform in the variables $(y,t)$ gives
\[
( i\lambda)^{k+1}e^{- i\lambda b}\widehat{\varphi_\theta}(\eta+\lambda w)=k!\widehat{\chi_\theta}(\eta)+\widehat{R_\theta}(\eta,\lambda).
\]
Therefore,
\begin{equation}\label{eqn_FT_main_rem}
    \widehat{\varphi_\theta}(\eta+\lambda w)=\frac{e^{ i\lambda b}k!}{( i\lambda)^{k+1}}\widehat{\chi_\theta}(\eta)+\mathcal O_\gamma\left(|\lambda|^{-k-2}(1+|\eta|)^{-\gamma}\right)
\end{equation}
uniformly for $\theta\in\mathcal B_\chi$ and $|\lambda|\geq1$.

We first prove the upper bound. By the definition of $\Delta_{q,s}$ and Parseval's identity,
\[
\|\Delta_{q,s}\varphi_\theta\|_2^2=\frac{1}{(2\pi)^d}\int_{\mathbb R^d}\zeta\left(\frac{|\xi|}{2^q}\right)^2(1+|\xi|^2)^s|\widehat{\varphi_\theta}(\xi)|^2\,d\xi .
\]
Writing $\xi=\eta+\lambda w$ with $\eta\in w^\perp$, we have $|\xi|\eqsim2^q$ on the support of the multiplier. We split this region into
\[
|\lambda|\geq2^{q-2}\quad\mathrm{and}\quad |\lambda|<2^{q-2}.
\]

For $|\lambda|\geq2^{q-2}$, the Fourier representation \eqref{eqn_FT_main_rem} yields
\[
|\widehat{\varphi_\theta}(\eta+\lambda w)|\lesssim_\gamma |\lambda|^{-k-1}(1+|\eta|)^{-\gamma}
\]
and we obtain
\begin{equation*}
    \begin{split}
        &\int_{|\lambda|\geq2^{q-2}}\zeta\left(\frac{|\xi|}{2^q}\right)^2(1+|\xi|^2)^s|\widehat{\varphi_\theta}(\xi)|^2d\xi\leq\int_{|\lambda|\geq2^{q-2}}\int_{w^\perp}(1+|\eta+\lambda w|^2)^s|\widehat{\varphi_\theta}(\eta+\lambda w)|^2d\eta d\lambda\\
        \lesssim&\int_{|\lambda|\geq2^{q-2}}|\lambda|^{-2(k-s)-2}d\lambda\lesssim2^{-q(2(k-s)+1)}.
    \end{split}
\end{equation*}

For $|\lambda|<2^{q-2}$, the multiplier is supported on $2^{q-1}\leq|\xi|\leq2^{q+1}$, so $|\eta|\geq2^{q-2}$. Tangential smoothness of $\varphi_\theta$ gives
\[
|\widehat{\varphi_\theta}(\eta+\lambda w)|\lesssim_\gamma (1+|\eta|)^{-\gamma}.
\]
Choosing $\gamma$ sufficiently large gives
\[
\int_{|\lambda|<2^{q-2}}\zeta\left(\frac{|\xi|}{2^q}\right)^2(1+|\xi|^2)^s|\widehat{\varphi_\theta}(\xi)|^2\,d\xi\lesssim2^{-q(2(k-s)+1)}.
\]
Hence, uniformly for $\theta\in\mathcal B_\chi$,
\[
\|\Delta_{q,s}\varphi_\theta\|_2^2\lesssim2^{-q(2(k-s)+1)}.
\]

We next prove the lower bound. The Fourier transform of $\chi_\theta$ on $w^\perp\simeq\mathbb R^{d-1}$ is
\[
\widehat{\chi_\theta}(\eta)=\int_{w^\perp}\chi(y-bw)e^{- i\eta\cdot y}\,dy,\qquad \eta\in w^\perp .
\]
Since each active hyperplane intersects $U$, the uniform continuity of $\chi$ provides a hyperplane disk of fixed radius on which $\chi\geq\frac{c_1}{2}$; hence Parseval's identity on $w^\perp$ gives
\[
\inf_{\theta\in\mathcal B_\chi}\int_{w^\perp}|\widehat{\chi_\theta}(\eta)|^2\,d\eta=(2\pi)^{d-1}\inf_{\theta\in\mathcal B_\chi}\|\chi_\theta\|_{\mathcal{L}^2(w^\perp)}^2>0.
\]
Therefore, there exists $R>0$, independent of $\theta$, such that
\[
\int_{|\eta|\leq R}|\widehat{\chi_\theta}(\eta)|^2\,d\eta\geq c>0.
\]

Consider the region
\[
|\eta|\leq R,\qquad \frac45\,2^q\leq\lambda\leq\frac54\,2^q .
\]
For sufficiently large $q$, this region satisfies
\[
\frac35\leq\frac{|\eta+\lambda w|}{2^q}\leq\frac53,
\]
and hence
\[
\zeta\left(\frac{|\eta+\lambda w|}{2^q}\right)\geq c_2>0.
\]
The leading term in \eqref{eqn_FT_main_rem} satisfies
\[
2^{2qs}\int_{|\eta|\leq R}\int_{\frac45 2^q}^{\frac54 2^q}|\lambda|^{-2k-2}|\widehat{\chi_\theta}(\eta)|^2\,d\lambda d\eta\gtrsim2^{-q(2k-2s+1)},
\]
while the remainder is bounded by
\[
2^{2qs}\int_{|\eta|\leq R}\int_{\frac45 2^q}^{\frac54 2^q}|\lambda|^{-2k-4}(1+|\eta|)^{-2\gamma}\,d\lambda d\eta\lesssim2^{-q}2^{-q(2k-2s+1)}.
\]
Hence, for sufficiently large $q$, uniformly for $\theta\in\mathcal B_\chi$,
\[
\|\Delta_{q,s}\varphi_\theta\|_2^2\gtrsim2^{-q(2k-2s+1)}.
\]
\end{proof}

\section{Proofs of the main results}\label{sec_proof}
\subsection{Proof of Theorem~\ref{thm_appr_rate}}
\begin{proof}

    The set $\{\theta_j^*\}_{j=1}^n$ can be extended to a collection $\{\eta_i^*\}_{i=1}^N\subset\SS^d$ with $N\eqsim n$, $\mal_{\eta\in\SS^d}\mil_{1\leq i\leq N}\rho(\eta,\eta_i^*)=h$ and $\{\eta_i^*\}_{i=1}^N\cap\mathcal{A}_\chi=\{\theta_j^*\}_{j=1}^n$. Then, by \cite{liu2025integral}, there exists $g_N(x)=\sul_{i=1}^Na_i\sigma_k(\eta_i^*\cdot\widetilde x)$ such that
    \begin{equation}\label{eqn_upper_rate_relu}
        \|f-g_N\|_{\mathcal H^s(\Omega')}\leq\|f-g_N\|_{\mathcal H^s(\Omega)}\lesssim h^{r-s}\|f\|_{\mathcal H^r(\Omega)}.
    \end{equation}
    Moreover, the coefficients can be chosen to satisfy
$$\Big(N\sul_{j=1}^Na_j^2\Big)^{\frac{1}{2}}\lesssim N^\alpha\|f\|_{\mH^r(\Omega)}$$
    if $\{\eta_i^*\}_{i=1}^N$ is quasi-uniform.

    Neurons with $\eta_i^*\notin\mathcal A_\chi$ are either zero or equal to $(\eta_i^*\cdot\widetilde x)^k$ on $U$. Hence
    \begin{equation*}
        \varphi_{\eta_i^*}(x)\equiv0\quad\hbox{or}\quad\varphi_{\eta_i^*}(x)=(\eta_i^*\cdot\widetilde x)^k,\qquad x\in\Omega'.
    \end{equation*}
    Let $$\mathcal{A}_{\Omega'}:=\left\{
        \theta=\binom{w}{b}\in\SS^d:
        \{x:w^\top x+b=0\}\cap \Omega'\ne\varnothing
    \right\},$$
    so that we can write
    $$g_N(x)=\sul_{\{i:\eta_i^*\in\mathcal{A}_{\Omega'}\}}a_i\varphi_{\eta_i^*}(x)+p(x),\quad x\in\Omega',$$
    where $p$ is a polynomial of degree $\leq k$. We now show that $p$ can be represented by the remaining basis functions. To this end, we prove
    \begin{equation}\label{eqn_span_poly}
        \mathrm{span}\{\varphi_{\eta_i^*}|_{\Omega'}:\eta_i^*\in\mathcal{A}_\chi\setminus\mathcal{A}_{\Omega'}\}=\mathbb P_k(\Omega')
    \end{equation}
    where $\mathbb P_k(\Omega')$ is the space of polynomials of degree $\leq k$ on $\Omega'$.

    Since $\Omega'\Subset U^\circ$, there exists $\delta$ such that
    $$\{x+y:~x\in\Omega',~|y|\leq\delta\}\subset U.$$
    Define
    \begin{equation*}
        \theta(\omega,t):=\frac{1}{\sqrt{1+(t-\mil_{x\in\overline{\Omega'}}\omega\cdot x)^2}}\binom{\omega}{t-\mil_{x\in\overline{\Omega'}}\omega\cdot x},\qquad\omega\in\SS^{d-1},~t\geq0,
    \end{equation*}
    so that $\theta(\omega,t)\in\SS^d$ and, for $0<t<\delta$, $\theta(\omega,t)\in\mathcal{A}_\chi\setminus\mathcal{A}_{\Omega'}$:
    \begin{enumerate}
    \item For any $x\in\Omega'$,
        $$\binom{x}{1}\cdot\binom{\omega}{t-\mil_{x\in\overline{\Omega'}}\omega\cdot x}=\omega\cdot x+t-\mil_{x\in\overline{\Omega'}}\omega\cdot x>0.$$
        \item Take $x_0\in\overline{\Omega'}$ with $\omega\cdot x_0=\mil_{x\in\overline{\Omega'}}\omega\cdot x$; then $x_0-t\omega\in U$ and
        $$\binom{x_0-t\omega}{1}\cdot\binom{\omega}{t-\mil_{x\in\overline{\Omega'}}\omega\cdot x}=\omega\cdot x_0-t\omega\cdot\omega+t-\mil_{x\in\overline{\Omega'}}\omega\cdot x=0.$$
    \end{enumerate}
    Define
    \begin{equation}
        \mathcal{S}_\delta:=\left\{\theta(\omega,t):~\omega\in\SS^{d-1},~\frac{\delta}{3}<t<\frac{2\delta}{3}\right\}
    \end{equation}
    and $K=\binom{d+k}{k}$. The class $\big\{(\vartheta\cdot\widetilde x)^k:~\vartheta\in\mathcal S_\delta\big\}$ spans $\PP_k$: any annihilating linear functional induces a homogeneous polynomial in $\vartheta$ vanishing on the open spherical set $\mathcal S_\delta$, so its coefficients, and hence the functional, vanish. Thus there exist $\vartheta_1^*,\dots,\vartheta_K^*\in\mathcal S_\delta$ such that $(\vartheta_u^*\cdot\widetilde x)^k$, $u=1,\dots,K$, are linearly independent. Equivalently,
    \begin{equation*}
        \mathrm{det}\left(\int_{\Omega'}(\vartheta_u^*\cdot\widetilde x)^k(\vartheta_v^*\cdot\widetilde x)^kdx\right)_{u,v=1}^K>0.
    \end{equation*}
    Choose $h_{\delta,1}>0$ such that 
    $$\left\{\theta\in\SS^d:~\rho(\theta,\mathcal{S}_\delta)\leq h_{\delta,1}\right\}\subset\left\{\theta(\omega,t):~\omega\in\SS^{d-1},~0<t<\delta\right\}$$
    and choose $h_{\delta,2}>0$ such that
    \begin{equation}\label{eqn_det_vartheta}
        \mathrm{det}\left(\int_{\Omega'}(\vartheta_u\cdot\widetilde x)^k(\vartheta_v\cdot\widetilde x)^kdx\right)_{u,v=1}^K>0
    \end{equation}
    whenever $|\vartheta_u-\vartheta_u^*|<h_{\delta,2}$.
    
    Set $h_\delta=\min\left\{\frac12\mil_{u\neq v}\{\rho(\vartheta_u^*,\vartheta_v^*)\},h_{\delta,1},h_{\delta,2}\right\}$. Then, for the extended collection $\{\eta_i^*\}_{i=1}^N$, whose active subset is $\{\theta_j^*\}_{j=1}^n$, whenever $h<h_\delta$ there exists a collection
    $$\{\eta_{i_u}^*\}_{u=1}^K\subset\left\{\theta(\omega,t):~\omega\in\SS^{d-1},~0<t<\delta\right\}\subset\mathcal{A}_\chi\setminus\mathcal{A}_{\Omega'}$$
    such that $|\eta_{i_u}^*-\vartheta_u^*|<h_\delta$, $u=1,\dots,K$. The corresponding determinant in \eqref{eqn_det_vartheta} is therefore positive, so \eqref{eqn_span_poly} holds. We next verify that replacing the polynomial part preserves the coefficient scale. By decreasing $h_{\delta,2}$ if necessary, the continuity of the Gram matrix and \eqref{eqn_det_vartheta} imply that there exists $c_4=c_4(\delta)>0$ such that
\begin{equation}\label{eqn_uniform_poly_gram}
\lambda_{\min}\left(\int_{\Omega'}(\vartheta_u\cdot\widetilde x)^k(\vartheta_v\cdot\widetilde x)^kdx\right)_{u,v=1}^K\geq c_4
\end{equation}
whenever $|\vartheta_u-\vartheta_u^*|<h_{\delta,2}$, $u=1,\dots,K$.

Therefore, there exists a sufficiently small $\varepsilon>0$, independent of $n$, such that
\begin{equation}
    \begin{split}
        &B(\vartheta_u^*,\varepsilon):=\left\{\theta\in\SS^d:~\rho(\theta,\vartheta_u^*)\leq\varepsilon\right\}\subset\mathcal S_\delta,\qquad u=1,\dots,K,\\
        &\mathrm{dist}\big(B(\vartheta_u^*,\varepsilon),B(\vartheta_v^*,\varepsilon)\big)\geq\varepsilon,\qquad u\neq v.
    \end{split}
\end{equation}
Decrease $h_\delta$ so that $h_\delta\leq\varepsilon$. By the fill-distance condition, $B(\vartheta_u^*,\varepsilon)\cap\{\eta_i^*\}_{i=1}^N\neq\varnothing$ for every $u$ whenever $h<h_\delta$. Choose $\eta_{i_u}^*$ in each intersection. Then \eqref{eqn_uniform_poly_gram} gives coefficients $\beta_1,\dots,\beta_K$ such that
\begin{equation*}
    p(x)=\sul_{u=1}^K\beta_{u}\varphi_{\eta_{i_{u}}^*}(x),\qquad x\in\Omega',
\end{equation*}
which proves \eqref{eqn_upp_bound}.

Next, we assume quasi-uniformity and prove \eqref{eqn_upp_coef_bound}. Since each $B(\vartheta_u^*,\varepsilon)$ is independent of $n$, for sufficiently large $n$ we have
$$n^*:=\mil_{1\leq u\leq K}\#\Big(B(\vartheta_u^*,\varepsilon)\cap\{\eta_i^*\}_{i=1}^N\Big)\eqsim n.$$
Choose $n^*$ elements from each $B(\vartheta_u^*,\varepsilon)\cap\{\eta_i^*\}_{i=1}^N$, denoted by
\[
\{\eta_{i_{u,\ell}}^*:u=1,\dots,K\},\qquad \ell=1,\dots,n^*,
\]
such that $\eta_{i_{u,\ell}}^*\in B(\vartheta_u^*,\varepsilon)\subset\mathcal A_\chi\setminus\mathcal A_{\Omega'}$. Again, for every $\ell$, \eqref{eqn_uniform_poly_gram} shows that there exist coefficients $\beta_{u,\ell}$ satisfying
\[
p(x)=\sul_{u=1}^K\beta_{u,\ell}\varphi_{\eta_{i_{u,\ell}}^*}(x),\qquad x\in\Omega',
\]
and $\sul_{u=1}^K|\beta_{u,\ell}|^2\lesssim\|p\|_{\mL(\Omega')}^2$.

To prove \eqref{eqn_upp_coef_bound}, note that $h\eqsim n^{-\frac{1}{d}}$ in the quasi-uniform setting, so it suffices to establish the coefficient bound. Averaging these representations gives
\[
p(x)=\sum_{\ell=1}^{n^*}\sum_{u=1}^K\alpha_{u,\ell}\varphi_{\eta_{i_{u,\ell}}^*}(x),\qquad
\alpha_{u,\ell}:=\frac{\beta_{u,\ell}}{n^*},
\]
and hence
\begin{equation}\label{eqn_poly_coeff_average}
\sum_{\ell=1}^{n^*}\sum_{u=1}^K|\alpha_{u,\ell}|^2
\lesssim\frac1{n^*}\|p\|_{\mL(\Omega')}^2
\lesssim\frac1n\|p\|_{\mL(\Omega')}^2.
\end{equation}
Moreover, since every neuron contributing to $p$ is either zero or a polynomial $(\eta_i^*\cdot\widetilde x)^k$ on $\Omega'$ and $\Omega'$ is bounded, we have
\[
\|p\|_{\mL(\Omega')}
\lesssim\sum_{i=1}^N|a_i|
\leq\sqrt N\left(\sum_{i=1}^Na_i^2\right)^{1/2}.
\]
It follows that
\begin{equation}\label{eqn_poly_coeff_final}
n\sum_{\ell=1}^{n^*}\sum_{u=1}^K|\alpha_{u,\ell}|^2
\lesssim\|p\|_{\mL(\Omega')}^2\lesssim N\sum_{i=1}^Na_i^2.
\end{equation}

Consequently, replacing the polynomial part of $g_N$ by the above active neurons preserves, up to a constant, the coefficient bound of the original $g_N$.
\end{proof}

\subsection{Proof of Theorem~\ref{thm_gram}}
\subsubsection{Upper bound}
\begin{proof}
The upper estimate in Theorem~\ref{thm:diagonal-estimate} implies
\begin{equation*}\sup_{\theta\in\mathcal A_\chi}\|\varphi_\theta\|_{\mB^{k+\frac12}(\RR^d)}\lesssim1.\end{equation*}
Thus there are subspaces $V_m\subset\mH^s(\RR^d)$ of dimension at most $m$ such that
\begin{equation}\label{eq:common-fourier-approximation}\dim V_m\leq m,\qquad \sup_{\theta\in\mathcal A_\chi}\inf_{v\in V_m}\|\varphi_\theta-v\|_{\mH^s(\RR^d)}^2\lesssim m^{-\frac{2(k-s)+1}{d}},\qquad m\geq1.\end{equation}
To see this, extend $\varphi_\theta$ periodically, with $\Omega$ contained in the interior of a period cell $[-A,A]^d$. Then standard Fourier series approximation (see, e.g., \cite[Chapters~2 and~9]{triebel1983theory}) of the extended periodic function yields the approximation rate $\mathcal{O}(m^{-\frac{2(k-s)+1}{2d}})$. Finally, multiplying the periodic approximant by another smooth cutoff $\chi_e\in\mathcal{C}^\infty(\RR^d)$ such that
\begin{equation*}
    \chi_e(x)=1,\quad x\in\Omega,\qquad\chi_e(x)=0,\quad x\notin[-A,A]^d
\end{equation*}
yields the approximant $v$ and the space $V_m$.

Let $\mathrm{Proj}_{V_m}$ be the $\mH^s(\RR^d)$-orthogonal projection onto $V_m$, and define
\begin{equation*}E_m:=\left(\left\langle(I-\mathrm{Proj}_{V_m})\varphi_{\theta_i^*},(I-\mathrm{Proj}_{V_m})\varphi_{\theta_j^*}\right\rangle_{\mH^s(\RR^d)}\right)_{i,j=1}^n.\end{equation*}
Orthogonality gives
\begin{equation*}\left\langle\mathrm{Proj}_{V_m}\varphi_{\theta_i^*},(I-\mathrm{Proj}_{V_m})\varphi_{\theta_j^*}\right\rangle_{\mH^s(\RR^d)}=0,\qquad 1\leq i,j\leq n,\end{equation*}
and hence
\begin{equation*}K_s-E_m=\left(\left\langle\mathrm{Proj}_{V_m}\varphi_{\theta_i^*},\mathrm{Proj}_{V_m}\varphi_{\theta_j^*}\right\rangle_{\mH^s(\RR^d)}\right)_{i,j=1}^n.\end{equation*}
Therefore,
\begin{equation*}0\preceq E_m\preceq K_s,\qquad \operatorname{rank}(K_s-E_m)\leq\dim V_m\leq m.\end{equation*}
Moreover, by the best-approximation property of $\mathrm{Proj}_{V_m}$ and \eqref{eq:common-fourier-approximation},
\begin{equation*}
\operatorname{tr}E_m=\sum_{i=1}^n\|(I-\mathrm{Proj}_{V_m})\varphi_{\theta_i^*}\|_{\mH^s(\RR^d)}^2=\sum_{i=1}^n\inf_{v\in V_m}\|\varphi_{\theta_i^*}-v\|_{\mH^s(\RR^d)}^2\lesssim n\,m^{-\frac{2(k-s)+1}{d}}.
\end{equation*}

Since $E_m$ is positive semidefinite,
\begin{equation*}\sum_{\ell=1}^m\lambda_\ell(K_s)\geq\sum_{\ell=1}^m\lambda_\ell(K_s-E_m)=\operatorname{tr}(K_s-E_m)\end{equation*}
and
\begin{equation}\label{eq:sobolev-gram-spectral-tail}\begin{aligned}\sum_{\ell=m+1}^n\lambda_\ell(K_s)&=\operatorname{tr}K_s-\sum_{\ell=1}^m\lambda_\ell(K_s)\leq\operatorname{tr}K_s-\operatorname{tr}(K_s-E_m)=\operatorname{tr}E_m\lesssim nm^{-\frac{2(k-s)+1}{d}}.\end{aligned}\end{equation}

For $2\leq j\leq n$, take $m=\lfloor j/2\rfloor$. Since the eigenvalues are nonincreasing,
\begin{equation*}(j-m)\lambda_j(K_s)\leq\sum_{\ell=m+1}^j\lambda_\ell(K_s)\leq\sum_{\ell=m+1}^n\lambda_\ell(K_s)\lesssim n\,m^{-\frac{2(k-s)+1}{d}}.\end{equation*}
Using $m\geq j/3$ and $j-m\geq j/2$, we obtain
\begin{equation*}\lambda_j(K_s)\lesssim\frac{n}{j-m}\,m^{-\frac{2(k-s)+1}{d}}\lesssim n\,j^{-\frac{d+2(k-s)+1}{d}},\qquad 2\leq j\leq n.\end{equation*}
For $j=1$, the uniform $\mH^s$ bound gives
\begin{equation*}\lambda_1(K_s)\leq\operatorname{tr}K_s=\sum_{i=1}^n\|\varphi_{\theta_i^*}\|_{\mH^s(\RR^d)}^2\lesssim n.\end{equation*}
This proves the upper bound. The constants depend only on $d,k,s$ and the fixed localization data.
\end{proof}

\subsubsection{Lower bound}
\begin{proof}
Define
\begin{equation*}Q_s(q):=\left(\left\langle\Delta_{q,s}\varphi_{\theta_i^*},\Delta_{q,s}\varphi_{\theta_j^*}\right\rangle_{\mL(\RR^d)}\right)_{i,j=1}^n,\qquad K_s^*:=\sum_{q=0}^\infty Q_s(q).\end{equation*}
By the definition of the Sobolev norm,
$$(K_s)_{ij}=\int_{\RR^d}(I-\bigtriangleup)^{\frac s2}\varphi_{\theta_i^*}(x)(I-\bigtriangleup)^{\frac s2}\varphi_{\theta_j^*}(x)dx=\left\langle\varphi_{\theta_i^*},\varphi_{\theta_j^*}\right\rangle_{\mH^s(\RR^d)},\qquad i,j=1,\dots,n.$$

For each $f_n=\sum\limits_{i=1}^n a_i\varphi_{\theta_i^*}$, the Littlewood--Paley norm equivalence gives
\begin{equation*}\begin{aligned}
a^\top K_sa=\|f_n\|_{\mH^s(\RR^d)}^2\eqsim\sum_{q=0}^\infty\|\Delta_{q,s}f_n\|_{\mL(\RR^d)}^2=\sum_{q=0}^\infty\sum_{i,j=1}^n a_i a_j\left\langle\Delta_{q,s}\varphi_{\theta_i^*},\Delta_{q,s}\varphi_{\theta_j^*}\right\rangle_{\mL(\RR^d)}=a^\top K_s^*a.
\end{aligned}\end{equation*}
Then the Courant--Fischer principle \cite[Chapter~III]{bhatia1997matrix} yields
\begin{equation}\label{eq:sobolev-gram-spectral-equivalence}
\lambda_j(K_s)\eqsim \lambda_j(K_s^*),\qquad 1\leq j\leq n,\end{equation}
where the eigenvalues are arranged in nonincreasing order. Hence it suffices to prove
\begin{equation*}\lambda_j(K_s^*)\gtrsim \max\left\{\underline h^{2(k-s)+1}\Big(\frac{n}{j}\Big)^{\frac{2(k-s)+1}{d}},(n^{\frac{1}{d}}\underline h)^{d+2(k-s)+1}nj^{-\frac{d+2(k-s)+1}{d}}\right\},\qquad 1\leq j\leq n.\end{equation*}

By Theorem~\ref{thm:gram}, for any $\gamma>0$, we have
\begin{equation}\label{eq:lower-proof-block-localization}
|Q_s(q)_{ij}|\lesssim_\gamma2^{-q(2(k-s)+1)}\left(2^q\rho_\pm(\theta_i^*,\theta_j^*)\right)^{-\gamma},\qquad q\geq 1.\end{equation}
Let $q_0\in\NN$ be as in Theorem~\ref{thm:diagonal-estimate}. Then
\begin{equation}\label{eq:lower-proof-block-diagonal}
Q_s(q)_{ii}\eqsim2^{-q(2(k-s)+1)},\qquad q\geq q_0,\quad 1\leq i\leq n.\end{equation}

Since spherical cap packing yields the annular count
\begin{equation*}
\#\left\{j:2^\ell\underline h\leq\rho_\pm(\theta_i^*,\theta_j^*)<2^{\ell+1}\underline h\right\}\lesssim2^{\ell d},\qquad \ell\leq\log_2\frac{2\pi}{\underline h},\end{equation*}
Thus we obtain, for $\gamma>d$,
\begin{equation}\label{eqn_nondiag_sum}
\begin{aligned}
\max_i\sum_{j\neq i}|Q_s(q)_{ij}|&\lesssim_\gamma2^{-q(2(k-s)+1)}\sum_{\ell=0}^\infty2^{\ell d}\min\left\{1,(2^q2^\ell\underline h)^{-\gamma}\right\}\\
&\lesssim_\gamma2^{-q(2(k-s)+1)}
\begin{cases}
(2^q\underline h)^{-d},&0<2^q\underline h\leq1,\\
(2^q\underline h)^{-\gamma},&2^q\underline h>1.
\end{cases}
\end{aligned}
\end{equation}

Together with \eqref{eq:lower-proof-block-diagonal}, this yields
\begin{equation}\label{eqn_Q_norm}
    \|Q_s(q)\|_2\leq\max_i\sum_{j=1}^n|Q_s(q)_{ij}|\lesssim2^{-q(2(k-s)+1)}\left(1+(2^q\underline h)^{-d}\right),\qquad q\geq q_0.
\end{equation}

We establish the lower bound for $\lambda_j(K_s^*)$ by choosing $q=q(j)$ and bounding $\lambda_j(Q_s(q))$ from below. Let $j\leq m< n$. The trace of $Q_s(q)$ can be split into three parts:
\begin{equation}
    \operatorname{tr}Q_s(q)=\sul_{\ell=1}^n\lambda_\ell(Q_s(q))=\sum_{\ell=j}^{m}\lambda_\ell(Q_s(q))+\sum_{\ell=1}^{j-1}\lambda_\ell(Q_s(q))+\sum_{\ell=m+1}^n\lambda_\ell(Q_s(q)),
\end{equation}
The first estimate below follows from \eqref{eq:lower-proof-block-diagonal}, the second from \eqref{eqn_Q_norm}, and the third from \eqref{eq:sobolev-gram-spectral-tail}. All corresponding constants are independent of $n,q,m$ and $j$.
\begin{equation}
    \begin{split}
    \sul_{\ell=1}^n\lambda_\ell(Q_s(q))=&\operatorname{tr}Q_s(q)\eqsim n2^{-q(2(k-s)+1)},\\
        \sum_{\ell=1}^{j-1}\lambda_\ell(Q_s(q))\leq&(j-1)\|Q_s(q)\|_2\lesssim j\big(1+(2^q\underline h)^{-d}\big)2^{-q(2(k-s)+1)},\\
        \sum_{\ell=m+1}^n\lambda_\ell(Q_s(q))\leq&\sum_{\ell=m+1}^n\lambda_\ell(K_s^*)\lesssim \sum_{\ell=m+1}^n\lambda_\ell(K_s)\lesssim n m^{-\frac{2(k-s)+1}{d}}.
    \end{split}
\end{equation}
Thus, there exist constants $c_5,C_2>0$ such that whenever $j\leq c_5n$, $n\geq C_2j(2^q\underline h)^{-d}$, $m\geq\min\{n,\lceil C_22^{qd}\rceil\}$,
\begin{equation*}
    \sum_{\ell=1}^{j-1}\lambda_\ell(Q_s(q))\leq\frac14\sul_{\ell=1}^n\lambda_\ell(Q_s(q)),\quad\sum_{\ell=m+1}^n\lambda_\ell(Q_s(q))\leq\frac14\sul_{\ell=1}^n\lambda_\ell(Q_s(q)).
\end{equation*}
Take $q=\lceil\log_2(C_2\underline h^{-1}\big(\frac{j}{n}\big)^{\frac1d})\rceil$ and $m=\min\{n-1,\lceil C_22^{qd}\rceil\}$, and increase $C_2$ if necessary so that $m\geq j$. We then obtain
\begin{equation*}
    \begin{split}
        \sum_{\ell=j}^{m}\lambda_\ell(Q_s(q))\geq&\sul_{\ell=1}^n\lambda_\ell(Q_s(q))-\sum_{\ell=1}^{j-1}\lambda_\ell(Q_s(q))-\sum_{\ell=m+1}^n\lambda_\ell(Q_s(q))\\
        \geq&\frac{1}{2}\sul_{\ell=1}^n\lambda_\ell(Q_s(q))\gtrsim n2^{-q(2(k-s)+1)}
    \end{split}
\end{equation*}
and consequently,
\begin{equation}\label{eqn_eigen_lower}
    \begin{split}
        \lambda_j(K_s)\eqsim&\lambda_j(K_s^*)\geq\lambda_j(Q_s(q))\geq\frac{1}{m-j+1}\sum_{\ell=j}^{m}\lambda_\ell(Q_s(q))\gtrsim \frac{n2^{-q(2(k-s)+1)}}{\min\{n,C_22^{qd}\}}\\
    \eqsim&\max\left\{\underline h^{2(k-s)+1}\Big(\frac{n}{j}\Big)^{\frac{2(k-s)+1}{d}},(n^{\frac{1}{d}}\underline h)^{d+2(k-s)+1}nj^{-\frac{d+2(k-s)+1}{d}}\right\}.
    \end{split}
\end{equation}

For $j>c_5n$, it suffices to consider the smallest eigenvalue $\lambda_n(K_s)$. In fact, by \eqref{eqn_nondiag_sum} and \eqref{eq:lower-proof-block-diagonal}, there exists $C_3$ such that for $2^q\geq C_3\underline h^{-1}$,
\begin{equation}\label{eqn_dig_dom}
    Q_s(q)_{ii}-\sum_{j\neq i}|Q_s(q)_{ij}|\gtrsim2^{-q(2(k-s)+1)}.
\end{equation}
Taking $q=\lceil\log_2(C_3\underline h^{-1})\rceil$ gives
\begin{equation}\label{eqn_eigen_Q_lower}
    \lambda_n(Q_s(q))\geq\mil_{a\in\RR^n,~|a|=1}a^\top Q_s(q)a\gtrsim2^{-q(2(k-s)+1)}\simeq\underline h^{2(k-s)+1}
\end{equation}
and
\begin{equation}
    \lambda_j(K_s)\gtrsim\lambda_j(K_s^*)\geq\lambda_j(Q_s(q))\geq\lambda_n(Q_s(q))\gtrsim\underline h^{2(k-s)+1},\quad j>c_5n,
\end{equation}
which agrees with the estimate in \eqref{eqn_eigen_lower} and completes the proof.

\end{proof}

\subsection{Proof of Theorem~\ref{thm_saturation}}
\begin{proof}
    If $\underline h=0$ or $\|\chi f\|_{\mathcal{L}^2(\Omega)}=0$ then there is nothing to prove. Now we only consider $\underline h>0$ and $\|\chi f\|_{\mathcal{L}^2(\Omega)}>0$. Let $f_n=\sul_{j=1}^na_j\varphi_{\theta_j^*}\in \chi L_n^k$ satisfy $\|\chi f-f_n\|_{\mathcal{L}^2(\Omega)}\leq\frac12\|\chi f\|_{\mathcal{L}^2(\Omega)}$. Since $\sup\limits_{x\in\Omega}\varphi_{\theta}(x)^2\lesssim1$ uniformly,
    \begin{equation*}
            n\|a\|_2^2\gtrsim\sup\limits_{x\in\Omega}\Big(n\sum_{j=1}^na_j^2\Big)\Big(\frac{1}{n}\sum_{j=1}^n\varphi_{\theta_j^*}(x)^2\Big)\geq\sup\limits_{x\in\Omega}\Big(\sum_{j=1}^na_j\varphi_{\theta_j^*}(x)\Big)^2\gtrsim\|f_n\|_{\mathcal{L}^2(\Omega)}^2\geq\frac{\|\chi f\|_{\mathcal{L}^2(\Omega)}^2}{4}.
    \end{equation*}
Let $\mq=\lceil\log_2(C_3\underline h^{-1})\rceil$ where $C_3$ is the constant for which \eqref{eqn_dig_dom} holds. We estimate the high-frequency part of $f_n$ as follows:
\begin{equation*}
    \begin{split}
        \int_{|\xi|\geq2^\mq}|\widehat{f_n}(\xi)|^2d\xi\geq&\int_{\RR^d}\Big|\sul_{q=\mq+1}^\infty\zeta\big(\frac{|\xi|}{2^q}\big)\widehat{f_n}(\xi)\Big|^2d\xi\eqsim\int_{\RR^d}\sul_{q=\mq+1}^\infty\big|\zeta\big(\frac{|\xi|}{2^q}\big)\widehat{f_n}(\xi)\big|^2d\xi=\int_{\RR^d}\sul_{q=\mq+1}^\infty\big|\widehat{\Delta_{q,0}f_n}(\xi)\big|^2d\xi\\
        =&(2\pi)^d\sul_{q=\mq+1}^\infty a^\top Q_0(q)a\gtrsim2^{-\mq(2k+1)}\|a\|_2^2\gtrsim n^{-1}\underline h^{2k+1}\|\chi f\|_{\mL(\Omega)}^2.
    \end{split}
\end{equation*}
On the other hand,
\begin{equation*}
        \int_{|\xi|\geq2^\mq}|\widehat{\chi f}(\xi)|^2d\xi\lesssim2^{-2\mq r}\int_{\RR^d}|\widehat{\chi f}(\xi)|^2(1+|\xi|^2)^rd\xi=(2\pi)^d2^{-2\mq r}\|\chi f\|_{\mH^r(\RR^d)}^2\eqsim\underline h^{2r}\|\chi f\|_{\mH^r(\Omega)}^2.
\end{equation*}
Since $r>\frac{d+2k+1}{2}$, the ratio $\displaystyle\frac{\underline h^{2r}}{n^{-1}\underline h^{2k+1}}$ tends to zero uniformly over parameter configurations. Thus there exists $n_0$, depending on $d,r$ and $\|\chi f\|_{\mH^r(\Omega)}\|\chi f\|_{\mL(\Omega)}^{-1}$, such that, for $n\geq n_0$,
\begin{equation*}
    \begin{split}
        \|\chi f-f_n\|_{\mL(\RR^d)}^2\geq&\frac{1}{(2\pi)^d}\int_{|\xi|\geq2^\mq}|\widehat{f_n}(\xi)-\widehat{\chi f}(\xi)|^2d\xi\geq\frac{1}{(2\pi)^d}\int_{|\xi|\geq2^\mq}\frac{1}{2}|\widehat{f_n}(\xi)|^2-2|\widehat{\chi f}(\xi)|^2d\xi\\
        \gtrsim& n^{-1}\underline h^{2k+1}\|\chi f\|_{\mL(\Omega)}^2.
    \end{split}
\end{equation*}

\end{proof}

\subsection{Proof of Theorem~\ref{thm_bernstein}}
\begin{proof}
    Using the Sobolev norms in Definition~\ref{def_bes_sob}, for $t\in\{s,r\}$ we obtain
    \begin{equation}\label{eqn_sobnorm_rewrite}
        \|f_n\|_{\mH^t(\RR^d)}^2\eqsim\sul_{q=0}^\infty\|\Delta_{q,t}f_n\|_{\mL(\RR^d)}^2=\sul_{q=0}^\infty\Big\|\sul_{j=1}^na_j\Delta_{q,t}\varphi_{\theta_j^*}\Big\|_{\mL(\RR^d)}^2=\sul_{q=0}^\infty a^\top Q_t(q)a.
    \end{equation}
    Let $\mq=\lceil\log_2(C_3\underline h^{-1})\rceil$ where $C_3$ is chosen so that \eqref{eqn_dig_dom} holds for both $r$ and $s$. Then, for $1\leq q\leq\mq$, the low-frequency blocks satisfy
    \begin{equation*}
        \begin{split}
            \|\Delta_{q,r}f_n\|_{\mL(\RR^d)}^2=&\frac{1}{(2\pi)^d}\int_{2^{q-1}\leq|\xi|\leq2^{q+1}}(1+|\xi|^2)^r\zeta\big(\frac{|\xi|}{2^q}\big)^2|\widehat{f_n}(\xi)|^2d\xi\\
            \lesssim&\frac{2^{2(q+1)(r-s)}}{(2\pi)^d}\int_{2^{q-1}\leq|\xi|\leq2^{q+1}}(1+|\xi|^2)^s\zeta\big(\frac{|\xi|}{2^q}\big)^2|\widehat{f_n}(\xi)|^2d\xi\lesssim\underline h^{-2(r-s)}\|\Delta_{q,s}f_n\|_{\mL(\RR^d)}^2.
        \end{split}
    \end{equation*}
    For $q=0$, the same argument applies, with the multipliers replaced by $\zeta_{0,r}$ and $\zeta_{0,s}$.
    
    For $q\geq\mq+1$, the high-frequency blocks satisfy
    \begin{equation*}
        \begin{split}
            \sul_{q=\mq+1}^\infty a^\top Q_r(q)a&\eqsim\sul_{q=\mq+1}^\infty2^{-q(2(k-r)+1)}\|a\|_2^2\eqsim2^{-\mq(2(k-r)+1)}\|a\|_2^2\eqsim\underline h^{-2(r-s)}2^{-\mq(2(k-s)+1)}\|a\|_2^2\\
            \eqsim&\underline h^{-2(r-s)}\sul_{q=\mq+1}^\infty a^\top Q_s(q)a
        \end{split}
    \end{equation*}
    Combining these two estimates with \eqref{eqn_sobnorm_rewrite} completes the proof.
\end{proof}

\subsection{Proof of Theorem~\ref{thm_mz}}\label{subsec_pf_mz}

    Define $$\mathcal{P}_\mq=\sul_{q=0}^\mq\Delta_{q,0},\qquad \pc_\mq=I-\mathcal{P}_\mq=\sul_{q=\mq+1}^\infty\Delta_{q,0},\qquad\mq>0,$$
    for the low- and high-frequency cutoffs, respectively. We first bound $\|\pc_\mq f_n\|_{\ell^2(X)}$ for large $\mq$. In this subsection, we use the original spherical parameters in $\mathcal A_\chi\subset\SS^d$.
    \begin{equation}\label{eqn_pc_mq_explicit}
        \begin{split}
            \|\pc_\mq f_n\|_{\ell^2(X)}^2=\frac{1}{m}\sul_{i=1}^m\pc_\mq f_n(x_i)^2=\frac{1}{m}\sul_{i=1}^m\Big(\sul_{j=1}^na_j\pc_\mq\varphi_{\theta_j^*}(x_i)\Big)^2=a^\top A_\mq^ca,
        \end{split}
    \end{equation}
    where
    \begin{equation*}
        (A_\mq^c)_{j,\ell}=\frac{1}{m}\sul_{i=1}^m\pc_\mq\varphi_{\theta_j^*}(x_i)\pc_\mq\varphi_{\theta_\ell^*}(x_i).
    \end{equation*}

\begin{lemma}\label{lem_high_freq_bound}
    Let $d,n,k$, and $\{\theta_j^*\}_{j=1}^n$ be as in Theorem~\ref{thm_appr_rate}. Let $X=\{x_i\}_{i=1}^m\subset\Omega$ be quasi-uniform with $h_X\leq\min\{1,\underline h\}$, and fix $\frac d2<\tau<\frac{d+1}{2}$. Then, for $\mq\geq\lceil\log_2(C_3\underline h^{-1})\rceil$ and $f_n=\sul_{j=1}^na_j\varphi_{\theta_j^*}$,
    \begin{equation}\label{eqn_high_freq_bound_tau}
        \begin{split}
            \|\pc_\mq f_n\|_{\ell^2(X)} \lesssim{}&\|\pc_\mq f_n\|_{\mL(\RR^d)} +\|a\|_2
            \begin{cases}
                h_X^{k+\frac12} \left[1+(h_X\underline h^{-d})^{ \frac{\tau-k-\frac12}{2\tau-1}}\right], &k\leq\frac{d-1}{2},\\[0.4em]
                2^{-\mq(k+\frac12)}(h_X2^\mq)^\tau, &k>\frac{d-1}{2}.
            \end{cases}
        \end{split}
    \end{equation}
    Here $h_X:=\sup_{x\in\Omega}\mil_{1\leq i\leq m}|x-x_i|$, and the implicit constant may depend on $d,k,\tau$, and the quasi-uniformity constant.
\end{lemma}

\begin{proof}
Denote by $K_\mq:=\mathcal F^{-1}(\sum_{q=0}^\mq\zeta_{q,0})$ the convolution kernel of $\mathcal P_\mq$. Then
\begin{equation*}
    \pc_\mq f(x)=f(x)-\int_{\RR^d}K_\mq(y)f(x-y)dy,\qquad x\in\RR^d.
\end{equation*}
By the standard Littlewood--Paley kernel estimates (see \cite[Chapter~2]{bahouri2011fourier}), for any $\alpha\in\NN_0^d$ and $\beta\geq0$,
\begin{equation*}
    |\partial_y^\alpha K_\mq(y)|\lesssim_{\alpha,\beta}2^{\mq(d+|\alpha|)}(1+2^\mq|y|)^{-\beta}.
\end{equation*}
Moreover, since the multiplier of $\mathcal P_\mq$ equals one in a neighborhood of the origin,
\begin{equation*}
    \int_{\RR^d}K_\mq(y)dy=1,\qquad \int_{\RR^d}y^\alpha K_\mq(y)dy=0,\quad |\alpha|\geq1.
\end{equation*}
Consequently, for any $\gamma\geq0$, the smooth compactly supported function satisfies
\begin{equation*}
    \Big|\pc_\mq\big[\chi(x)(w\cdot x+b)^k\big]\Big|\lesssim_\gamma2^{-\mq(k+\gamma)}(1+|x|)^{-\gamma}\lesssim_\gamma2^{-\mq k}(1+2^\mq|w\cdot x+b|)^{-\gamma},
\end{equation*}
uniformly in $\theta$, since the parameters are bounded and $\chi$ is fixed.

If $w\cdot x+b<0$, then $\varphi_\theta(x)=0$, and the convolution integral only involves points $x-y$ on the opposite side of the hyperplane. If $w\cdot x+b>0$, the same observation applies to $\varphi_\theta-\chi(\circ)(w\cdot\circ+b)^k$. In either case, crossing the hyperplane requires $|y|\geq c|w\cdot x+b|$, and the corresponding truncated power is bounded by a constant times $|y|^k$. Hence, for $w\cdot x+b\neq0$,
\begin{equation*}
    \begin{split}
        |\pc_\mq\varphi_\theta(x)|&\lesssim\Big|\pc_\mq\big[\chi(x)(w\cdot x+b)^k\big]\Big|+\int_{|y|\geq c|w\cdot x+b|}|K_\mq(y)||y|^kdy\\
        &\lesssim_\gamma2^{-\mq k}(1+2^\mq|w\cdot x+b|)^{-\gamma},
    \end{split}
\end{equation*}
where the integral is estimated using the kernel bound with $\beta>d+k+\gamma$. On the hyperplane $w\cdot x+b=0$, the same bound follows from $|\varphi_\theta(x-y)|\lesssim|y|^k$ and $\varphi_\theta(x)=0$ when $k\geq1$; when $k=0$, it follows from the uniform boundedness of $\varphi_\theta$ and $\|K_\mq\|_{\mathcal{L}^1(\RR^d)}$. Therefore, for every $k\in\NN_0$,
\begin{equation}\label{eqn_local_vanish}
    \Big|\pc_\mq\varphi_\theta(x)\Big|\lesssim_\gamma2^{-\mq k}(1+2^\mq|w\cdot x+b|)^{-\gamma},\qquad x\in\RR^d.
\end{equation}

    Define
    \begin{equation}\label{eqn_Aqc_sum1}
        \begin{split}
            &\mathcal{I}_j(0):=\big\{\ell:~\rho_\pm(\theta_j^*,\theta_\ell^*)<\underline h\big\},\quad\mathcal{I}_j(u):=\big\{\ell:~2^{u-1}\underline h\leq\rho_\pm(\theta_j^*,\theta_\ell^*)<2^{u}\underline h\big\},\qquad u\geq1,\\
            &\mathcal{I}_{j,\ell}(0):=\big\{i:~\mal\left\{|w_j^*\cdot x_i+b_j^*|,|w_\ell^*\cdot x_i+b_\ell^*|\right\}<2^{-\mq}\big\}\\
            &\mathcal{I}_{j,\ell}(v):=\big\{i:~2^{v-\mq-1}\leq\mal\left\{|w_j^*\cdot x_i+b_j^*|,|w_\ell^*\cdot x_i+b_\ell^*|\right\}<2^{v-\mq}\big\},\qquad v\geq1,
        \end{split}
    \end{equation}
    so that, for each $j$, dyadic summation yields
    \begin{equation*}
         \begin{split}
             \sul_{\ell=1}^n\big|(A_\mq^c)_{j,\ell}\big|&\leq\frac{1}{m}\sul_{u=0}^{\lceil\log_2\frac{\pi}{\underline h}\rceil}\sul_{\ell\in\mathcal{I}_j(u)}\sul_{v=0}^{\mq+1}\sul_{i\in\mathcal{I}_{j,\ell}(v)}\Big|\pc_\mq\varphi_{\theta_j^*}(x_i)\pc_\mq\varphi_{\theta_\ell^*}(x_i)\Big|\\
             &\lesssim_\gamma\frac{2^{-2\mq k}}{m}\sul_{u=0}^{\lceil\log_2\frac{\pi}{\underline h}\rceil}\sul_{\ell\in\mathcal{I}_j(u)}\sul_{v=0}^{\mq+1}\sul_{i\in\mathcal{I}_{j,\ell}(v)}(1+2^\mq|w_j^*\cdot x_i+b_j^*|)^{-\gamma}(1+2^\mq|w_\ell^*\cdot x_i+b_\ell^*|)^{-\gamma}\\
             &\lesssim\frac{2^{-2\mq k}}{m}\sul_{u=0}^{\lceil\log_2\frac{\pi}{\underline h}\rceil}\sul_{\ell\in\mathcal{I}_j(u)}\sul_{v=0}^{\mq+1}\sul_{i\in\mathcal{I}_{j,\ell}(v)}2^{-\gamma v}.
         \end{split}   
    \end{equation*}
        Since $|w\cdot x+b|\leq\epsilon$ places $x$ in a belt satisfying $\mathrm{dist}(x,\{w\cdot x+b=0\})\lesssim \epsilon$, the points in $\mathcal{I}_{j,\ell}(v)$ satisfy
        \begin{equation*}
            \mathrm{dist}(x,\{w_j^*\cdot x+b_j^*=0\})+\mathrm{dist}(x,\{w_\ell^*\cdot x+b_\ell^*=0\})\leq C_42^{v-\mq}
        \end{equation*}
        for some $C_4$. Consequently, the points lie in the set
        \begin{equation*}
            \Xi_{j,\ell}(v):=\left\{x\in\Omega:~|\theta_j^*\cdot\widetilde x|\leq C_42^{v-\mq},~|\theta_\ell^*\cdot\widetilde x|\leq C_42^{v-\mq}\right\},\qquad1\leq j,\ell\leq n.
        \end{equation*}
        For $d\geq2$ and $\ell\neq j$, let $\theta_{\ell,j}$ be the orthogonal projection of $\theta_\ell^*$ onto the direction $\theta_j^*$ and $\theta_{\ell,j}^\perp=\theta_\ell^*-\theta_{\ell,j}$; then $|\theta_{\ell,j}^\perp|=\sqrt{1-(\theta_j^*\cdot\theta_\ell^*)^2}\eqsim\rho_\pm(\theta_j^*,\theta_\ell^*)$ and
        consequently, there exists $C_4'$ such that
        \begin{equation*}
            \begin{split}
                \Xi_{j,\ell}(v)\subset&\left\{x\in\Omega:~|\theta_j^*\cdot\widetilde x|\leq C_42^{v-\mq},~|\theta_{\ell,j}^\perp\cdot\widetilde x|\leq 2C_42^{v-\mq}\right\}\\
                \subset&\left\{x\in\Omega:~|\theta_j^*\cdot\widetilde x|\leq C_42^{v-\mq},~\left|\frac{\theta_{\ell,j}^\perp}{|\theta_{\ell,j}^\perp|}\cdot\widetilde x\right|\leq C_4'\frac{2^{v-\mq}}{\rho_\pm(\theta_j^*,\theta_\ell^*)}\right\}.
            \end{split}
        \end{equation*}
        Since $\theta_{\ell,j}^\perp\perp\theta_j^*$, the normalized points $\frac{1}{\sqrt{|x|^2+1}}\binom{x}{1}\in\SS^d$ lie in a rectangular cylinder with widths bounded by $C_42^{v-\mq}$ and $C_4'2^{v-\mq}\rho_\pm(\theta_j^*,\theta_\ell^*)^{-1}$. This normalization preserves distances on $\Omega$ up to constants, so the number of samples in $\Xi_{j,\ell}(v)$ satisfies
        \begin{equation}
            \begin{split}
                &\#\mathcal{I}_{j,\ell}(v)\leq\#\big(X\cap\Xi_{j,\ell}(v)\big)\lesssim m(2^{(v-\mq)}+h_X)(2^{(v-\mq)}\rho_\pm(\theta_j^*,\theta_\ell^*)^{-1}+h_X),\\
                &\#\mathcal{I}_{j,j}(v)\leq\#\big(X\cap\Xi_{j,j}(v)\big)\lesssim m(2^{v-\mq}+h_X).
            \end{split}
        \end{equation}
    For $d\geq2$, the sum in \eqref{eqn_Aqc_sum1} is bounded by
    \begin{equation}
        \begin{split}
            \sul_{\ell=1}^n\big|(A_\mq^c)_{j,\ell}\big|&\lesssim_\gamma\frac{2^{-2\mq k}}{m}\sul_{u=1}^{\lceil\log_2\frac{\pi}{\underline h}\rceil}\sul_{\ell\in\mathcal{I}_j(u)}\sul_{v=0}^{\mq+1}2^{-\gamma v}\#\mathcal{I}_{j,\ell}(v)+\frac{2^{-2\mq k}}{m}\sul_{v=0}^{\mq+1}2^{-\gamma v}\#\mathcal{I}_{j,j}(v)\\
            &\lesssim2^{-2\mq k}\sul_{u=1}^{\lceil\log_2\frac{\pi}{\underline h}\rceil}\sul_{\ell\in\mathcal{I}_j(u)}\Big(\frac{2^{-2\mq}}{\rho_\pm(\theta_j^*,\theta_\ell^*)}+\frac{2^{-\mq}h_X}{\rho_\pm(\theta_j^*,\theta_\ell^*)}+h_X^2\Big)+2^{-2\mq k}(2^{-\mq}+h_X)\\
            &\lesssim2^{-2\mq k}\sul_{u=0}^{\lceil\log_2\frac{\pi}{\underline h}\rceil}2^{ud}(2^{-2\mq}2^{-u}\underline h^{-1}+2^{-\mq}2^{-u}\underline h^{-1}h_X+h_X^2)+2^{-2\mq k}(2^{-\mq}+h_X)\\
            &\eqsim2^{-2\mq k}\Big(2^{-2\mq}+h_X^2\Big)\underline h^{-d}+2^{-\mq(2k+1)}+2^{-2\mq k}h_X.
        \end{split}
    \end{equation}
For $d=1$, the original spherical parameters satisfy $c_3\leq|w_j^*|\leq1$. Their antipodal separation implies that the breakpoints $-b_j^*/w_j^*$ are separated by $\gtrsim\underline h$. Since $2^\mq\underline h\gtrsim1$ and the samples are separated by $\gtrsim h_X$, for $\gamma>1$ we have
\begin{equation*}
    \sup_{x\in\Omega}\sul_{j=1}^n(1+2^\mq|w_j^*x+b_j^*|)^{-\gamma}\lesssim1,\qquad\sup_{1\leq j\leq n}\frac1m\sul_{i=1}^m(1+2^\mq|w_j^*x_i+b_j^*|)^{-\gamma}\lesssim2^{-\mq}+h_X.
\end{equation*}
Applying weighted Cauchy--Schwarz to $\pc_\mq f_n(x_i)=\sul_{j=1}^na_j\pc_\mq\varphi_{\theta_j^*}(x_i)$ and using \eqref{eqn_local_vanish}, we obtain
\begin{equation*}
    \|\pc_\mq f_n\|_{\ell^2(X)}^2\lesssim2^{-2\mq k}(2^{-\mq}+h_X)\|a\|_2^2.
\end{equation*}
Thus, the estimate below also holds when $d=1$.

Substituting the corresponding estimates into \eqref{eqn_pc_mq_explicit} gives, for $\mq\geq\lceil\log_2(C_3\underline h^{-1})\rceil$,
\begin{equation}\label{eq:MZ_belt_tail}
\begin{split}
\|\mathcal P_\mq^cf_n\|_{\ell^2(X)}^2&\lesssim2^{-2\mq k}\Big[\Big(2^{-2\mq}+h_X^2\Big)\underline h^{-d}+2^{-\mq}+h_X\Big]\|a\|_2^2.
\end{split}
\end{equation}

Applying the sampling inequality in \cite{arcangeli2014sampling} with $\tau>\frac{d}{2}$ to $\Delta_{q,0}f_n$ gives
\begin{equation}\label{eq:MZ_block}
    \begin{split}
        \|\Delta_{q,0}f_n\|_{\ell^2(X)} \lesssim{}&\|\Delta_{q,0}f_n\|_{\mL(\RR^d)} +h_X^\tau\|\Delta_{q,0}f_n\|_{\mH^\tau(\RR^d)}\\
        \lesssim{}&\|\Delta_{q,0}f_n\|_{\mL(\RR^d)} +h_X^\tau2^{q(\tau-k-\frac12)}\|a\|_2, \qquad q\geq\lceil\log_2(C_3\underline h^{-1})\rceil.
    \end{split}
\end{equation}
By the block upper estimate and a diagonally dominant block whose frequency support lies above that of the multiplier of $\mathcal P_\mq$,
\begin{equation*}
    \sul_{q=\mq+1}^{\infty} \|\Delta_{q,0}f_n\|_{\mL(\RR^d)} \lesssim2^{-\mq(k+\frac12)}\|a\|_2 \lesssim\|\pc_\mq f_n\|_{\mL(\RR^d)}.
\end{equation*}
Therefore, for $\mathfrak p\geq\mq$,
\begin{equation}\label{eqn_high_freq_split_tau}
    \begin{split}
        &\|\pc_\mq f_n\|_{\ell^2(X)} =\Big\|\sul_{q=\mq+1}^{\mathfrak p} \Delta_{q,0}f_n+\pc_{\mathfrak p}f_n\Big\|_{\ell^2(X)}\\
        \lesssim&\|\pc_\mq f_n\|_{\mL(\RR^d)} +\Bigg[h_X^\tau\sul_{q=\mq+1}^{\mathfrak p} 2^{q(\tau-k-\frac12)}+2^{-\mathfrak p k} \left((2^{-\mathfrak p}+h_X)\underline h^{-\frac{d}{2}} +(2^{-\mathfrak p}+h_X)^{\frac{1}{2}}\right)\Bigg]\|a\|_2.
    \end{split}
\end{equation}

If $k\leq\frac{d-1}{2}$, then $\tau>k+1/2$. Set $\displaystyle\mathfrak p= \left\lceil\log_2\left(h_X^{-1}\max\big\{1, (h_X\underline h^{-d})^{\frac1{2\tau-1}}\big\} \right)\right\rceil$,
so that $2^{-\mathfrak p}\lesssim h_X$ and
\begin{equation*}
    \begin{split}
        h_X^\tau2^{\mathfrak p(\tau-k-\frac12)} +2^{-\mathfrak p k} \left(h_X\underline h^{-\frac{d}{2}}+h_X^{\frac{1}{2}}\right)\lesssim h_X^{k+\frac12} \left[1+(h_X\underline h^{-d})^{ \frac{\tau-k-\frac12}{2\tau-1}}\right].
    \end{split}
\end{equation*}
If $\mq\leq\mathfrak p$, substitution into \eqref{eqn_high_freq_split_tau} proves the first case of \eqref{eqn_high_freq_bound_tau}. If $\mq>\mathfrak p$, apply \eqref{eq:MZ_belt_tail} directly; its right-hand side is nonincreasing in $\mq$, so the same bound follows.

For $k>\frac{d-1}{2}$, we take $\frac{d}{2}<\tau<\min\{k+\frac{1}{2},\frac{d+1}{2}\}$. Letting $\mathfrak p\to\infty$ in \eqref{eqn_high_freq_split_tau}, we obtain
\begin{equation*}
    \begin{split}
        \|\pc_\mq f_n\|_{\ell^2(X)} \lesssim{}&\|\pc_\mq f_n\|_{\mL(\RR^d)} +h_X^\tau2^{\mq(\tau-k-\frac12)}\|a\|_2=\|\pc_\mq f_n\|_{\mL(\RR^d)} +2^{-\mq(k+\frac12)}(h_X2^\mq)^\tau\|a\|_2.
    \end{split}
\end{equation*}
This completes the proof.
\end{proof}

\begin{proof}[Proof of Theorem~\ref{thm_mz}]
    We first establish the sampling estimates for the low-frequency part. If $\operatorname{supp}\widehat g\subset B(0,2^{\mathfrak p+1})$ and $2^{\mathfrak p}\leq h_X^{-1}$, then
    \begin{equation}\label{eqn_mz_bandlimited}
        \begin{split}
            \|g\|_{\ell^2(X)}\lesssim\|g\|_{\mL(\RR^d)},\qquad
            \|g\|_{\mL(\Omega)}\lesssim\|g\|_{\ell^2(X)} +h_X2^{\mathfrak p}\|g\|_{\mL(\RR^d)}.
        \end{split}
    \end{equation}
    Indeed, let $\Omega_i$ be the Voronoi cells of $X$ in $\Omega$. Then $|x-x_i|\leq h_X$ for $x\in\Omega_i$ and $|\Omega_i|\lesssim h_X^d\eqsim m^{-1}$. A smooth reproducing kernel and the mean value theorem give
    \begin{equation*}
        |g(x)-g(x_i)|\lesssim h_X2^{\mathfrak p} \int_{\RR^d} \frac{2^{\mathfrak p d}|g(y)|} {(1+2^{\mathfrak p}|x-y|)^{d+1}}dy, \qquad x\in\Omega_i.
    \end{equation*}
    Young's inequality proves the second estimate in \eqref{eqn_mz_bandlimited}. For the first estimate, the same kernel and the separation of $X$ give
    \begin{equation*}
            |g(x_i)|^2\lesssim\int_{\RR^d} \frac{2^{\mathfrak p d}|g(y)|^2} {(1+2^{\mathfrak p}|x_i-y|)^{d+1}}dy,
    \end{equation*}
    and the kernel satisfies 
    $$\sup\limits_{y\in\RR^d}\frac1m\sul_{i=1}^m \frac{2^{\mathfrak p d}} {(1+2^{\mathfrak p}|x_i-y|)^{d+1}}\lesssim1.$$

    Let $f_n=\sul_{j=1}^na_j\varphi_{\theta_j^*}$. Choose a sufficiently small constant $c_6>0$ and take $\mathfrak p=\lfloor\log_2(c_6h_X^{-1})\rfloor$. By increasing $C_1$, we may assume
    \begin{equation}\label{eqn_hX_estim}
        \mathfrak p\geq\lceil\log_2(C_3\underline h^{-1})\rceil+2, \qquad h_X2^{\mathfrak p}\leq c_6, \qquad2^{-\mathfrak p}\eqsim c_6^{-1}h_X.
    \end{equation}
    Then, by Lemma~\ref{lem_high_freq_bound},
    \begin{equation*}
        \begin{split}
            \|\pc_{\mathfrak p}f_n\|_{\ell^2(X)}\lesssim\|a\|_2
            \begin{cases}
                h_X^{k+\frac12} \left[1+(h_X\underline h^{-d})^{ \frac{\tau-k-\frac12}{2\tau-1}}\right], &k\leq\frac{d-1}{2},\\[0.4em]
                h_X^{k+\frac12},&k>\frac{d-1}{2}.
            \end{cases}
        \end{split}
    \end{equation*}
    On the other hand, by \eqref{eqn_gram_norm_eq} and Theorem~\ref{thm_gram}, $\|f_n\|_{\mL(\RR^d)}\gtrsim\underline h^{k+\frac{1}{2}}\|a\|_2$. Thus, for $2^{\mathfrak p}\gtrsim\underline h^{-1}$ with a sufficiently large implicit constant, we have
\begin{equation}
    \|f_n\|_{\mL(\RR^d)}\eqsim\|\mathcal{P}_{\mathfrak p} f_n\|_{\mL(\RR^d)}\gtrsim \underline h^{k+\frac{1}{2}}\|a\|_2.
\end{equation}
Since $m^{\frac{1}{d}}\eqsim2^{\mathfrak p}$, we can choose a sufficiently large constant $C_1$ such that
    \begin{equation*}
        m\geq C_1
        \begin{cases}
            \underline h^{-d\left(1+ \frac{(d-1)(\tau-k-\frac12)} {2\tau(k+1)-2k-1}\right)}, &k\leq\frac{d-1}{2},\\[0.4em]
            \underline h^{-d},&k>\frac{d-1}{2},
        \end{cases}
    \end{equation*}
    implies $\|\pc_{\mathfrak p}f_n\|_{\ell^2(X)}\lesssim\|\mathcal P_{\mathfrak p}f_n\|_{\mL(\RR^d)}$
and \eqref{eqn_mz_bandlimited} yields
    \begin{equation*}
        \begin{split}
            \|f_n\|_{\ell^2(X)}\leq\|\mathcal P_{\mathfrak p}f_n\|_{\ell^2(X)} +\|\pc_{\mathfrak p}f_n\|_{\ell^2(X)}\lesssim\|\mathcal P_{\mathfrak p}f_n\|_{\mL(\RR^d)}\lesssim\|f_n\|_{\mL(\RR^d)}.
        \end{split}
    \end{equation*}

    For the reverse inequality, the high-frequency block estimate and the coefficient lower bound give
\begin{equation*}
    \|\mathcal P_{\mathfrak p}^cf_n\|_{\mL(\RR^d)}\lesssim2^{-\mathfrak p(k+\frac12)}\|a\|_2\lesssim(2^{\mathfrak p}\underline h)^{-(k+\frac12)}\|f_n\|_{\mL(\RR^d)}.
\end{equation*}
Increasing $C_1$ if necessary, we may assume that the last expression is at most $\frac14\|f_n\|_{\mL(\RR^d)}$. Since $\operatorname{supp}(f_n)\subset\Omega$, the triangle inequality yields
\begin{equation*}
    \begin{split}
        \|\mathcal P_{\mathfrak p}f_n\|_{\mL(\Omega)}&\geq\|f_n\|_{\mL(\Omega)}-\|\mathcal P_{\mathfrak p}^cf_n\|_{\mL(\Omega)}\geq\frac34\|f_n\|_{\mL(\RR^d)},\\
        \|\mathcal P_{\mathfrak p}f_n\|_{\mL(\RR^d)}&\leq\|f_n\|_{\mL(\RR^d)}+\|\mathcal P_{\mathfrak p}^cf_n\|_{\mL(\RR^d)}\leq\frac54\|f_n\|_{\mL(\RR^d)}.
    \end{split}
\end{equation*}
Consequently, $\|\mathcal P_{\mathfrak p}f_n\|_{\mL(\RR^d)}\leq\frac53\|\mathcal P_{\mathfrak p}f_n\|_{\mL(\Omega)}$. Taking $c_6$ sufficiently small and then increasing $C_1$ if necessary, we can therefore absorb the remainder in \eqref{eqn_mz_bandlimited} to obtain
\begin{equation*}
    \frac12\|\mathcal P_{\mathfrak p}f_n\|_{\mL(\Omega)}\lesssim\|\mathcal P_{\mathfrak p}f_n\|_{\ell^2(X)}\leq\|f_n\|_{\ell^2(X)}+\|\mathcal P_{\mathfrak p}^cf_n\|_{\ell^2(X)}.
\end{equation*}
It follows that there exists $c_7>0$ such that
\begin{equation*}
    c_7\|f_n\|_{\mL(\Omega)}\leq\|f_n\|_{\ell^2(X)}+\|\mathcal P_{\mathfrak p}^cf_n\|_{\ell^2(X)}.
\end{equation*}
Increasing $C_1$ further so that $\|\mathcal P_{\mathfrak p}^cf_n\|_{\ell^2(X)}\leq\frac{c_7}{2}\|f_n\|_{\mL(\Omega)}$, we obtain
\begin{equation*}
    \frac{c_7}{2}\|f_n\|_{\mL(\Omega)}\leq\|f_n\|_{\ell^2(X)},
\end{equation*}
as desired.

\end{proof}

\subsection{Proof of Theorem~\ref{thm_ls}}

The proof uses high-frequency upper estimates for parameters on the full sphere. Unlike the spectral lower bounds, these estimates do not require the singular hyperplanes to intersect $U$. We retain the Fourier cutoffs $\mathcal P_{\mq}$ and $\mathcal P_{\mq}^c$ defined in Section~\ref{subsec_pf_mz}.

\begin{lemma}\label{lem:ls-full-sphere-tail}
Let $L_n^k=L_n^k(\Theta_n)$ and $\underline h>0$ be as in Theorem \ref{thm_ls}, let $0\leq t<k+\frac{1}{2}$ and $q\geq1$. Then for any $v_n=\sul_{j=1}^na_j\varphi_{\theta_j^*}\in L_n^k$,
\begin{equation}\label{eq:ls-full-sphere-block}
\|\Delta_{q,t}v_n\|_{\mathcal L^2(\RR^d)}\lesssim 2^{-q(k+\frac{1}{2}-t)}\left(1+(2^q\underline h)^{-d}\right)^{\frac{1}{2}}\|a\|_2.
\end{equation}
Consequently, for $\mq\geq1$,
\begin{equation}\label{eq:ls-full-sphere-tail}
\|\mathcal P_{\mq}^cv_n\|_{\mathcal H^t(\RR^d)}\lesssim 2^{-\mq(k+\frac{1}{2}-t)}\left(1+(2^{\mq}\underline h)^{-d}\right)^{\frac{1}{2}}\|a\|_2.
\end{equation}
\end{lemma}

\begin{proof}
As in the preceding proofs, assume that $\Omega\subset\BB^d$. Write $v_n=v_n^{(0)}+v_n^{(1)}$, where $v_n^{(0)}$ contains the terms with $|b_j^*|\leq|w_j^*|$. For these parameters, $|w_j^*|\geq\frac{1}{\sqrt{2}}$ and $\frac{|b_j^*|}{|w_j^*|}\leq1$. The single-neuron upper bound and the off-diagonal estimate used to prove \eqref{eqn_Q_norm} therefore remain valid, since their proofs require neither intersection with $U$ nor positivity of $\chi$ on the singular hyperplane. The same matrix upper-bound argument gives
\begin{equation*}
\|\Delta_{q,t}v_n^{(0)}\|_{\mathcal L^2(\RR^d)}^2\lesssim 2^{-q(2(k-t)+1)}\left(1+(2^q\underline h)^{-d}\right)\|a\|_2^2.
\end{equation*}
For the remaining parameters, each neuron is either zero or a polynomial on $\BB^d$. After multiplication by $\chi$, these functions have uniformly bounded $\mathcal H^\gamma(\RR^d)$ norms for every fixed integer $\gamma$. Choose $\gamma>k+\frac{d+1}{2}$. Antipodal packing gives $n\lesssim\underline h^{-d}$, and hence
\begin{equation*}
\|\Delta_{q,t}v_n^{(1)}\|_{\mathcal L^2(\RR^d)}\lesssim\sqrt n\,2^{-q(\gamma-t)}\|a\|_2\lesssim 2^{-q(k+\frac{1}{2}-t)}(2^q\underline h)^{-\frac{d}{2}}\|a\|_2.
\end{equation*}
Combining these estimates proves \eqref{eq:ls-full-sphere-block}. Finite overlap and geometric summation give
\begin{equation*}
\|\mathcal P_{\mathfrak q}^cv_n\|_{\mathcal H^t(\RR^d)}^2\lesssim\sum_{q=\mathfrak q+1}^{\infty}2^{-q(2(k-t)+1)}\left(1+(2^q\underline h)^{-d}\right)\|a\|_2^2\lesssim 2^{-\mathfrak q(2(k-t)+1)}\left(1+(2^{\mathfrak q}\underline h)^{-d}\right)\|a\|_2^2,
\end{equation*}
which proves \eqref{eq:ls-full-sphere-tail}.
\end{proof}

\begin{lemma}\label{lem:ls-sampling}
Let $X\subset\Omega$ be quasi-uniform with fill distance $h_X$, let $2\leq p<\infty$, and let $\frac{d}{p}<\tau<\frac{d}{p}+1$. For sufficiently small $h_X$,
\begin{equation}\label{eq:ls-sampling}
\|v\|_{\ell^2(X)}\lesssim\|v\|_{\mathcal L^p(\RR^d)}+h_X^\tau\|v\|_{\mathcal W^{\tau,p}(\RR^d)},\qquad v\in\mathcal W^{\tau,p}(\RR^d).
\end{equation}
\end{lemma}

\begin{proof}
Rescaling Sobolev embedding on balls of radius comparable to $h_X$ centred at the samples and summing with bounded overlap gives
\begin{equation*}
\frac{1}{m}\sum_{i=1}^m|v(x_i)|^p\lesssim\|v\|_{\mathcal L^p(\RR^d)}^p+h_X^{\tau p}\|v\|_{\mathcal W^{\tau,p}(\RR^d)}^p,
\end{equation*}
where $mh_X^d\eqsim1$ follows from quasi-uniformity. Finally, since $p\geq2$, the standard norm bound $\|\circ\|_{\ell^2(X)}\leq\|\circ\|_{\ell^p(X)}$ completes the proof.
\end{proof}

\begin{proof}[Proof of Theorem~\ref{thm_ls}]
For $r>\frac{d}{2}$, set $p=2$; otherwise, let $2<p<\infty$ satisfy $\frac{d}{p}<r\leq\frac{d}{2}$. In this proof, we take the Fourier cutoff scale
\begin{equation}\label{eq:ls-cutoff-scale}
\varepsilon:=h^{\frac{2k+1}{d+2k+1}}\underline h^{\frac{d}{d+2k+1}}\leq h,
\end{equation}
and choose $\mathfrak q\geq1$ such that $2^{-\mathfrak q}\eqsim\varepsilon$. By quasi-uniformity of $X$ and the sampling condition $m\geq C_5\varepsilon^{-d}$, we have $h_X\eqsim m^{-\frac{1}{d}}\lesssim C_5^{-\frac{1}{d}}\varepsilon$, and hence $h_X2^{\mathfrak q}\lesssim C_5^{-\frac{1}{d}}$. By increasing $C_5$, we may therefore assume that the sampling estimates apply and $h_X2^{\mathfrak q}$ is sufficiently small. The identity
\begin{equation*}
\varepsilon^{2k+1}\left(1+\Big(\frac{\varepsilon}{\underline h}\Big)^d\right)=\varepsilon^{2k+1}+h^{2k+1}\lesssim h^{2k+1}
\end{equation*}
and Lemma~\ref{lem:ls-full-sphere-tail} imply for $ 0\leq t<k+\frac{1}{2}$,
\begin{equation*}
\Big\|\sum_{j=1}^na_j\mathcal P_{\mathfrak q}^c\varphi_{\theta_j^*}\Big\|_{\mathcal H^t(\RR^d)}\lesssim\varepsilon^{-t}h^{k+\frac{1}{2}}\|a\|_2,\quad a\in\RR^n.
\end{equation*}
Choose $\frac{d}{2}<\tau_2<\min\{k+\frac{1}{2},\frac{d}{2}+1\}$. Applying Lemma~\ref{lem:ls-sampling} with $p=2$, we obtain
\begin{equation}\label{eq:ls-network-discrete-tail}
\Big\|\sum_{j=1}^na_j\mathcal P_{\mathfrak q}^c\varphi_{\theta_j^*}\Big\|_{\ell^2(X)}\lesssim\big(1+(h_X\varepsilon^{-1})^{\tau_2}\big)h^{k+\frac{1}{2}}\|a\|_2\lesssim h^{k+\frac{1}{2}}\|a\|_2,\quad a\in\RR^n.
\end{equation}
Similarly, choose $\frac{d}{p}<\tau_p<\min\{r,\frac{d}{p}+1\}$. Lemma~\ref{lem:ls-sampling} and the standard Fourier cutoff estimates yield
\begin{equation}\label{eq:ls-target-discrete-tail}
\begin{aligned}
\|\mathcal P_{\mathfrak q}^c(\chi f)\|_{\ell^2(X)}&\lesssim\|\mathcal P_{\mathfrak q}^c(\chi f)\|_{\mathcal L^p(\RR^d)}+h_X^{\tau_p}\|\mathcal P_{\mathfrak q}^c(\chi f)\|_{\mathcal W^{\tau_p,p}(\RR^d)}\\
&\lesssim\big(\varepsilon^r+h_X^{\tau_p}\varepsilon^{r-\tau_p}\big)\|f\|_{\mathcal W^{r,p}(\Omega)}\lesssim h^r\|f\|_{\mathcal W^{r,p}(\Omega)}.
\end{aligned}
\end{equation}

By \cite[Theorem~2.2]{liu2025integral}, there exists $g_n=\sul_{j=1}^na_j\sigma_k(w_j^*\cdot x+b_j^*)$ such that
\begin{equation}\label{eqn_coef_corr_est}
    \|g_n-f\|_{\mL(\Omega)}\lesssim n^{-\frac{r}{d}}\|f\|_{\mH^r(\Omega)},\qquad\|a\|_2\lesssim h^{r-k-\frac{1}{2}}\|f\|_{\mH^r(\Omega)}.
\end{equation}
Taking $v_n=\chi g_n$, then
\begin{equation}
    \|v_n-\chi f\|_{\mL(\Omega)}=\|\chi (g_n-f)\|_{\mL(\Omega)}\lesssim n^{-\frac{r}{d}}\|f\|_{\mH^r(\Omega)}
\end{equation}
and consequently
\begin{equation*}
\begin{aligned}
\|\chi f-v_n\|_{\ell^2(X)}&\leq\|\mathcal P_{\mathfrak q}(\chi f-v_n)\|_{\ell^2(X)}+\|\mathcal P_{\mathfrak q}^c(\chi f)\|_{\ell^2(X)}+\|\mathcal P_{\mathfrak q}^cv_n\|_{\ell^2(X)}\\
&\lesssim\|\chi f-v_n\|_{\mathcal L^2(\RR^d)}+h^r\|f\|_{\mathcal W^{r,p}(\Omega)}+h^{k+\frac{1}{2}}\|a\|_2\lesssim h^r\|f\|_{\mathcal W^{r,p}(\Omega)}.
\end{aligned}
\end{equation*}
where the first term is follows by the sampling estimate for the low-frequency functions \eqref{eqn_mz_bandlimited}, second by \eqref{eq:ls-target-discrete-tail}, and third by \eqref{eq:ls-network-discrete-tail}. Together with the coefficient estimate in \eqref{eqn_coef_corr_est},
$$\|\chi f-v_n\|_{\ell^2(X)}^2+h^{2k+1}\|a\|_2^2\lesssim h^{2r}\|f\|_{\mathcal W^{r,p}(\Omega)}^2.$$
Since $f_{n,m}$ is the minimizer of \eqref{eq:domain-ls-objective},
\begin{equation}\label{eq:ls-minimizer-bounds}
\|\chi f-f_{n,m}\|_{\ell^2(X)}^2+h^{2k+1}\|\boldsymbol\alpha\|_2^2\lesssim h^{2r}\|f\|_{\mathcal W^{r,p}(\Omega)}^2.
\end{equation}
In particular,
\begin{equation}\label{eqn_alpha_norm_est}
    \|\boldsymbol\alpha\|_2\lesssim h^{r-k-\frac{1}{2}}\|f\|_{\mathcal W^{r,p}(\Omega)}
\end{equation}

Now we are ready to consider $\|\chi f-f_{n,m}\|_{\mathcal L^2(\Omega)}$. Since $\chi f-f_{n,m}$ is supported in $\operatorname{supp}\chi\subset\Omega$, the bandlimited lower sampling estimate in \eqref{eqn_mz_bandlimited} gives
\begin{equation*}
\begin{aligned}
&\|\chi f-f_{n,m}\|_{\mathcal L^2(\Omega)}\leq\|\mathcal P_{\mathfrak q}(\chi f-f_{n,m})\|_{\mathcal L^2(\Omega)}+\|\mathcal P_{\mathfrak q}^c(\chi f-f_{n,m})\|_{\mathcal L^2(\Omega)}\\
\lesssim&\|\mathcal P_{\mathfrak q}(\chi f-f_{n,m})\|_{\ell^2(X)}+h_X2^{\mathfrak q}\|\mathcal P_{\mathfrak q}(\chi f-f_{n,m})\|_{\mathcal L^2(\RR^d)}+\|\mathcal P_{\mathfrak q}^c(\chi f-f_{n,m})\|_{\mathcal L^2(\RR^d)}\\
\lesssim& \|\mathcal P_{\mathfrak q}(\chi f-f_{n,m})\|_{\ell^2(X)}+h_X2^{\mathfrak q}\|\chi f-f_{n,m}\|_{\mathcal L^2(\RR^d)}.
\end{aligned}
\end{equation*}
By increasing $C_5$, the last term can be absorbed, giving $\|\chi f-f_{n,m}\|_{\mathcal L^2(\Omega)}\lesssim \|\mathcal P_{\mathfrak q}(\chi f-f_{n,m})\|_{\ell^2(X)}$.
Now by \eqref{eq:ls-minimizer-bounds}, \eqref{eq:ls-target-discrete-tail}, \eqref{eq:ls-network-discrete-tail}, and \eqref{eqn_alpha_norm_est}, 
\begin{equation*}
    \begin{split}
        \|\chi f-f_{n,m}\|_{\mathcal L^2(\Omega)}\lesssim&\|\mathcal P_{\mathfrak q}(\chi f-f_{n,m})\|_{\ell^2(X)}\leq\|\chi f-f_{n,m}\|_{\ell^2(X)}+\|\mathcal P_{\mathfrak q}^c(\chi f-f_{n,m})\|_{\ell^2(X)}\\
        \leq&\|\chi f-f_{n,m}\|_{\ell^2(X)}+\|\mathcal P_{\mathfrak q}^c\chi f\|_{\ell^2(X)}+\|\mathcal P_{\mathfrak q}^cf_{n,m}\|_{\ell^2(X)}\lesssim h^r\|f\|_{\mathcal W^{r,p}(\Omega)}.
    \end{split}
\end{equation*}

Finally, the standard low-frequency bound and Lemma \ref{lem:ls-full-sphere-tail} give
\begin{equation*}
\begin{aligned}
\|\chi f-f_{n,m}\|_{\mathcal H^s(\RR^d)}\leq&\|\mathcal{P}_\mq(\chi f-f_{n,m})\|_{\mathcal H^s(\RR^d)}+\|\pc_\mq(\chi f)\|_{\mathcal H^s(\RR^d)}+\|\pc_\mq f_{n,m}\|_{\mathcal H^s(\RR^d)}\\
\lesssim&\varepsilon^{-s}\|\chi f-f_{n,m}\|_{\mathcal L^2(\RR^d)}+\varepsilon^{r-s}\|f\|_{\mH^r(\Omega)}+\varepsilon^{-s}h^{k+\frac{1}{2}}\|\boldsymbol\alpha\|_2\\
\lesssim&h^r\varepsilon^{-s}\|f\|_{\mathcal W^{r,p}(\Omega)}=h^{r-s}\Big(\frac{h}{\underline h}\Big)^{\frac{ds}{d+2k+1}}\|f\|_{\mathcal W^{r,p}(\Omega)}.
\end{aligned}
\end{equation*}
Restricting to $\Omega'$ and using $\chi=1$ there proves the result.

\end{proof}

\subsection{Proof of Theorem~\ref{thm_inverse}}

\begin{proof}
We prove each item in turn.
\paragraph{Proof of item~\ref{item1}.}
   The low-frequency blocks are controlled by the $\mL$ norm. For large $q$, take $n=n_q:=\max\{n\in\NN:~2^q\geq C_3\underline h_n^{-1}\}$. Then $n\eqsim2^{qd}$, and \eqref{eqn_Q_norm} gives
   $$\|\Delta_{q,0}f_n\|_{\mL(\RR^d)}^2=a^\top Q_0(q)a\lesssim2^{-q(2k+1)}\|a\|_2^2\lesssim2^{-q(d+2k+1-2\alpha d)}.$$
   Together with \eqref{eq:euclidean-constrained-rate}, this gives
   \begin{equation*}
       \begin{split}
           \|\Delta_{q,0}f\|_{\mL(\RR^d)}\leq&\|\Delta_{q,0}f_n\|_{\mL(\RR^d)}+\|\Delta_{q,0}(f-f_n)\|_{\mL(\RR^d)}\leq\|\Delta_{q,0}f_n\|_{\mL(\RR^d)}+\|f-f_n\|_{\mL(\RR^d)}\\
           \lesssim&2^{-q\frac{d+2k+1-2\alpha d}{2}},
       \end{split}
   \end{equation*}
   and $\sup\limits_{q\geq0}2^{q(\frac{d+2k+1}{2}-\alpha d)}\|\Delta_{q,0}f\|_2<\infty$, i.e., $f\in\mathcal B_{2,\infty}^{\frac{d+2k+1}{2}-\alpha d}(\RR^d)$.

   \paragraph{Proof of item~\ref{item2}.}

Fix a cube $Q\Subset\Omega'$ and choose a compactly supported orthonormal wavelet system on $\RR^d$ with regularity greater than $\frac{d+2k+1}{2}$. There exists $q_0\geq0$ such that whenever $q\geq q_0$, there is at least one wavelet $\psi_q$ at scale $2^{-q}$ such that $\operatorname{supp}(\psi_q)\subset Q$. The collection $\{\psi_q\}_{q\geq q_0}$ is orthogonal and, by the standard wavelet characterizations of Sobolev and Besov spaces,
\begin{equation}\label{eqn:item2_wavelet_char}
\Big\|\sum_{q=q_0}^\infty \alpha_q\psi_q\Big\|_{\mathcal H^\gamma(\RR^d)}^2\eqsim\sum_{q=q_0}^\infty2^{2q\gamma}|\alpha_q|^2,
\qquad
\Big\|\sum_{q=q_0}^\infty\alpha_q\psi_q\Big\|_{\mathcal B_{2,\infty}^\gamma(\RR^d)}
\eqsim\sup_q2^{q\gamma}|\alpha_q|.
\end{equation}
Let
$$t=\frac{d+2k+1-2\alpha d}{4},\qquad f:=\sul_{q=q_0}^\infty2^{-qt}\psi_q,$$
so a direct computation using \eqref{eqn:item2_wavelet_char} yields $f\in\mL(\RR^d)$ and $f\notin\mathcal B_{2,\infty}^{\frac{d+2k+1}{2}-\alpha d}(\RR^d)$. Since $\operatorname{supp}(f)\subset Q\Subset\Omega'$, this also implies $f|_{\Omega'}\notin\mathcal B_{2,\infty}^{\frac{d+2k+1}{2}-\alpha d}(\Omega')$. For $n\geq2$, set
\[
q_n:=\left\lfloor
\frac{\alpha}{\frac{d+2k+1}{2}-t}\log_2n
\right\rfloor,\qquad g_n:=\sum_{q=q_0}^{q_n}2^{-qt}\psi_q.
\]
Then $q_n\to\infty$, and
\begin{equation}\label{eqn:item2_truncated_sobolev_euclidean}
\begin{split}
\|g_n\|_{\mathcal H^{\frac{d+2k+1}{2}}(\RR^d)}^2
\lesssim
\sum_{q=q_0}^{q_n}
2^{q(d+2k+1)}2^{-2qt}&\lesssim
2^{2q_n(\frac{d+2k+1}{2}-t)}
\lesssim
n^{2\alpha}.
\end{split}
\end{equation}
Moreover,
\begin{equation}\label{eqn:item2_tail_euclidean}
\|f-g_n\|_{\mL(\RR^d)}^2
=
\sum_{q>q_n}2^{-2qt}
\lesssim
2^{-2q_nt}
\longrightarrow0.
\end{equation}

By Theorem~\ref{thm_appr_rate}, for every sufficiently large $n$ there exists $f_n=\sul_{j=1}^na_{j,n}\varphi_{\theta_{j,n}^*}$ such that
\begin{equation}\label{eqn:item2_direct_euclidean}
\|g_n-f_n\|_{\mathcal{L}^2(\Omega')}
\lesssim
n^{-\frac{d+2k+1}{2d}}
\|g_n\|_{\mathcal H^{\frac{d+2k+1}{2}}(\Omega)}
\lesssim
n^{-\frac{d+2k+1}{2d}+\alpha},
\end{equation}
and
\begin{equation}\label{eqn:item2_coeff_euclidean}
\Big(n\sum_{j=1}^na_{j,n}^2\Big)^{\frac12}
\lesssim
\|g_n\|_{\mathcal H^{\frac{d+2k+1}{2}}(\Omega)}
\lesssim
n^\alpha.
\end{equation}
Now \eqref{eqn:item2_direct_euclidean} and \eqref{eqn:item2_tail_euclidean} imply
\[
\|f-f_n\|_{\mathcal{L}^2(\Omega')}\leq\|f-g_n\|_{\mathcal{L}^2(\Omega')}+\|g_n-f_n\|_{\mathcal{L}^2(\Omega')}
\longrightarrow0.
\]
If the constant in \eqref{eqn:item2_coeff_euclidean} exceeds $M$, rescale $f$ and its approximants by a sufficiently small positive constant. This proves item~\ref{item2}.

\paragraph{Proof of item~\ref{item3}.}
\paragraph{Case 1: $\alpha>\frac{1}{2}$.}
Let $f\in\mathcal L^2(\RR^d)$ satisfy \eqref{eqn:rate_alpha}. For $q\geq0$, let $n_q=2^{(q+1)d}$ and choose $v_q\in\chi L_{n_q}^k(\Theta_{n_q})$ such that
\begin{equation}\label{eqn:item3_dyadic_approximants}
    \|f-v_q\|_{\mathcal L^2(\RR^d)}\lesssim2^{-q(\frac{d+2k+1}{2}-\alpha d)}.
\end{equation}
Set $u_0:=v_0$ and $u_q:=v_q-v_{q-1}$ for $q\geq1$. By nestedness, $u_q\in\chi L_{n_q}^k(\Theta_{n_q})$, and \eqref{eqn:item3_dyadic_approximants} gives $\|u_q\|_{\mathcal L^2(\RR^d)}\lesssim2^{-q(\frac{d+2k+1}{2}-\alpha d)}$ after adjusting the constant to include the first term. Take $r\in(\frac{d+2k+1}{2}-\alpha d,k+\frac{1}{2})$. Antipodal quasi-uniformity and Theorem~\ref{thm_bernstein} yield
\begin{equation}\label{eqn:item3_difference_bernstein}
    \|u_q\|_{\mathcal H^r(\RR^d)}\lesssim n_q^{\frac{r}{d}}\|u_q\|_{\mathcal L^2(\RR^d)}\lesssim2^{q(r-\frac{d+2k+1}{2}+\alpha d)}.
\end{equation}
As $f=\sul_{q=0}^\infty u_q$ in $\mathcal L^2(\RR^d)$, equations \eqref{eqn:item3_dyadic_approximants}--\eqref{eqn:item3_difference_bernstein} give, for $Q\in\NN$,
\begin{equation}\label{eqn:item3_besov_block}
    \begin{split}
        \|\Delta_{Q,0}f\|_{\mathcal L^2(\RR^d)}&\leq\sum_{q<Q}\|\Delta_{Q,0}u_q\|_{\mathcal L^2(\RR^d)}+\sum_{q\geq Q}\|u_q\|_{\mathcal L^2(\RR^d)}\lesssim2^{-Qr}\sum_{q<Q}\|\Delta_{Q,0}u_q\|_{\mathcal H^r(\RR^d)}+\sum_{q\geq Q}\|u_q\|_{\mathcal L^2(\RR^d)}\\
        &\lesssim2^{-Qr}\sum_{q<Q}2^{q(r-\frac{d+2k+1}{2}+\alpha d)}+\sum_{q\geq Q}2^{-q(\frac{d+2k+1}{2}-\alpha d)}\lesssim2^{-Q(\frac{d+2k+1}{2}-\alpha d)}.
    \end{split}
\end{equation}
Consequently, $f\in\mathcal B_{2,\infty}^{\frac{d+2k+1}{2}-\alpha d}(\RR^d)$.

\paragraph{Case 2: $0<\alpha<\frac12$.}
By nestedness, fix $\theta_{1,1}^*\in\Theta_1$ and set $f=\varphi_{\theta_{1,1}^*}$. Then $f\in\chi L_n^k(\Theta_n)$ for every $n$, so the approximation error in \eqref{eqn:rate_alpha} is zero. On the other hand, Theorem~\ref{thm:diagonal-estimate} gives
\begin{equation*}\|\Delta_{q,0}f\|_2^2\eqsim2^{-q(2k+1)},\qquad q\geq q_0.\end{equation*}
Therefore,
\begin{equation*}2^{q(\frac{d+2k+1}{2}-\alpha d)}\|\Delta_{q,0}f\|_2\eqsim2^{qd(\frac12-\alpha)}\longrightarrow\infty,\end{equation*}
which establishes the counterexample.

\paragraph{Case 3: $\alpha=\frac12$.}
By nestedness, after relabeling the points if necessary, we may write
\[
\Theta_n=\{\theta_1^*,\ldots,\theta_n^*\},\qquad n=1,2\dots.
\]
Then, for each $\nu$, we have
\begin{equation*}
    \Theta_{2^{\nu+1}}\setminus\Theta_{2^{\nu}}=\{\theta_{2^{\nu}+1}^*,\theta_{2^{\nu}+2}^*,\dots,\theta_{2^{\nu+1}}^*\},\quad \nu=1,2,\dots
\end{equation*}
Set $a_1=a_2=0$. For each $\nu$, by \eqref{eqn_gram_norm_eq} and Theorem~\ref{thm_gram}, let $a(\nu)=(a_{2^{\nu}+1},\dots,a_{2^{\nu+1}})^\top$ be a unit vector in the eigenspace of the smallest eigenvalue of the matrix
$$K_0=K_0(\nu)=\left(\langle\varphi_{\theta_i^*},\varphi_{\theta_j^*}\rangle_{\mL(\RR^d)}\right)_{i,j=2^{\nu}+1}^{2^{\nu+1}},$$
so the corresponding function $\displaystyle g_\nu:=\sul_{j=2^{\nu}+1}^{2^{\nu+1}}a_j\varphi_{\theta_j^*}$ satisfies
\begin{equation}
    \|g_\nu\|_{\mL(\RR^d)}^2=a(\nu)^\top K_0(\nu)a(\nu)=\lambda_{\min}\big(K_0(\nu)\big)\lesssim 2^{-\nu\frac{2k+1}{d}}.
\end{equation}
Let
\begin{equation*}
f:=\sum_{\nu=1}^\infty g_\nu\in\mathcal L^2(\RR^d),\qquad F_\nu:=\sum_{\ell=1}^\nu g_\ell\in\chi L_{2^{\nu+1}}^k(\Theta_{2^{\nu+1}}),\qquad\nu=1,2,\dots.
\end{equation*}

For $2^{\nu+1}\leq n<2^{\nu+2}$, by nestedness, \eqref{eqn:rate_alpha} holds with $\alpha=\frac12$:
\begin{equation*}\inf_{f_n\in\chi L_n^k(\Theta_n)}\|f-f_n\|_{\mL(\RR^d)}\leq\|f-F_\nu\|_{\mL(\RR^d)}\leq\sum_{\ell>\nu}\|g_\ell\|_{\mL(\RR^d)}\lesssim2^{-\nu\frac{2k+1}{2d}}\lesssim n^{-\frac{2k+1}{2d}}.\end{equation*}

Next, we show that the stated Besov regularity fails. For each $\nu$, take $q_\nu=\lceil\log_2(C_3\underline h_{2^{\nu+1}}^{-1})\rceil$ with $C_3$ chosen so that \eqref{eqn_eigen_Q_lower} holds.

Let $a^{(\nu)}=(0,0,a_3,\dots,a_{2^{\nu+1}})^\top$, and form $Q_0(q_\nu)$ using $\Theta_{2^{\nu+1}}$. Taking $C_3$ sufficiently large for \eqref{eqn_eigen_Q_lower} gives
\begin{equation*}
\|\Delta_{q_\nu,0}F_\nu\|_{\mL(\RR^d)}^2=(a^{(\nu)})^\top Q_0(q_\nu)a^{(\nu)}\gtrsim2^{-q_\nu(2k+1)}\sul_{j=1}^{2^{\nu+1}}a_j^2\geq\nu2^{-q_\nu(2k+1)}.
\end{equation*}

Meanwhile,
\begin{equation*}\|\Delta_{q_\nu,0}(f-F_\nu)\|_{\mL(\RR^d)}\lesssim\|f-F_\nu\|_{\mL(\RR^d)}\lesssim2^{-\nu\frac{2k+1}{2d}}\lesssim2^{-q_\nu(k+\frac12)},
\end{equation*}
and hence, for sufficiently large $\nu$,
\begin{equation*}
    \|\Delta_{q_\nu,0}f\|_{\mL(\RR^d)}\geq\|\Delta_{q_\nu,0}F_\nu\|_{\mL(\RR^d)}-\|\Delta_{q_\nu,0}(f-F_\nu)\|_{\mL(\RR^d)}\gtrsim\sqrt{\nu}2^{-q_\nu(k+\frac12)}
\end{equation*}
Letting $\nu\to\infty$ gives $\|f\|_{\mathcal B_{2,\infty}^{k+\frac12}(\RR^d)}=\infty$.

\paragraph{Proof of item~\ref{item4}.}
For each $n$, choose
\begin{equation*}f_n=\sum_{j=1}^na_{j,n}\varphi_{\theta_{j,n}^*},\qquad \widetilde f_n=\sum_{j=1}^n\widetilde a_{j,n}\varphi_{\widetilde\theta_{j,n}^*}\end{equation*}
such that
\begin{equation*}\|f-f_n\|_{\mL(\RR^d)}+\|f-\widetilde f_n\|_{\mL(\RR^d)}\lesssim n^{-\frac{d+2k+1}{2d}+\alpha}.\end{equation*}
The union contains $2n$ distinct parameters with antipodal separation $\gtrsim n^{-1/d}$. Applying \eqref{eqn_gram_norm_eq} and Theorem~\ref{thm_gram} to the coefficients $a_{1,n},\ldots,a_{n,n},-\widetilde a_{1,n},\ldots,-\widetilde a_{n,n}$ gives
\begin{equation*}\|f_n-\widetilde f_n\|_{\mL(\RR^d)}^2\gtrsim n^{-\frac{2k+1}{d}}\Big(\sum_{j=1}^na_{j,n}^2+\sum_{j=1}^n\widetilde a_{j,n}^2\Big).\end{equation*}
On the other hand,
\begin{equation*}\|f_n-\widetilde f_n\|_{\mL(\RR^d)}\leq\|f-f_n\|_{\mL(\RR^d)}+\|f-\widetilde f_n\|_{\mL(\RR^d)}\lesssim n^{-\frac{d+2k+1}{2d}+\alpha}.\end{equation*}
Consequently,
\begin{equation*}\sum_{j=1}^na_{j,n}^2+\sum_{j=1}^n\widetilde a_{j,n}^2\lesssim n^{\frac{2k+1}{d}}n^{-\frac{d+2k+1}{d}+2\alpha}=n^{2\alpha-1}.\end{equation*}
In particular,
\begin{equation*}\Big(n\sum_{j=1}^na_{j,n}^2\Big)^{\frac12}\lesssim n^\alpha.\end{equation*}
Thus, for some $M$ independent of $n$, the approximants $f_n$ belong to $\chi L_{n,M}^{k,\alpha}(\Theta_n)$ and retain the rate \eqref{eq:euclidean-constrained-rate}. Item~\ref{item1} now implies
\begin{equation*}f\in\mathcal B_{2,\infty}^{\frac{d+2k+1}{2}-\alpha d}(\RR^d),\end{equation*}
which completes the proof.
\end{proof}

\bibliographystyle{abbrv}
\bibliography{ref}

\end{document}